\documentclass[11pt, a4paper,english]{article}

\usepackage{natbib}
\usepackage{url}

 \usepackage{amsmath,amssymb}
 \usepackage{bm}
 \usepackage{ascmac}
 \usepackage{amsthm}
 \usepackage{enumerate}
 \usepackage[dvips]{color}
 \usepackage{here}
\usepackage[pdftex]{graphicx}

\usepackage{mathrsfs} 

\theoremstyle{definition}
\newtheorem{theorem}{Theorem}[section]

\newtheorem{lemma}{Lemma}[section]

\newtheorem{definition}{Definition}[section]

\newtheorem{condition}{Condition}[section]

\makeatletter 
 \def\section{\@startsection {section}{1}{\z@}{3.5ex plus -1ex minus -.2ex}{2.3 ex plus .2ex}{\bf}}
 \def\@seccntformat#1{\csname the#1\endcsname.\ }
 \makeatother
 
  \makeatletter 
 \def\subsection{\@startsection {subsection}{1}{\z@}{3.5ex plus -1ex minus -.2ex}{2.3 ex plus .2ex}{\bf}}
 \def\@seccntformat#1{\csname the#1\endcsname.\ }
 \makeatother

\numberwithin{equation}{section} 

\newcommand{\argmin}{\operatornamewithlimits{argmin}}
\newcommand{\argmax}{\operatornamewithlimits{argmax}}

\usepackage{algorithm}
\usepackage{algcompatible}

\renewcommand{\algorithmicrequire}{\textbf{Input:}}
\renewcommand{\algorithmicensure}{\textbf{Output:}}
\newcommand{\1}{\mbox{1}\hspace{-0.25em}\mbox{l}}

\usepackage{slashbox}

\usepackage{geometry}
\begin{document}
\begin{center}
{\Large Bounding Box-based Multi-objective Bayesian Optimization of Risk Measures under Input Uncertainty} \vspace{5mm} 

Yu Inatsu$^{1,\ast}$ \  Shion Takeno$^{2}$ \  Hiroyuki Hanada$^{2}$ \  Kazuki Iwata$^{1}$ \      Ichiro Takeuchi$^{2,3}$ \vspace{3mm}   

$^1$ Department of Computer Science, Nagoya Institute of Technology    \\
$^2$ RIKEN Center for Advanced Intelligence Project \\
$^3$ Department of Mechanical Systems Engineering, Nagoya University \\
$^\ast$ E-mail: inatsu.yu@nitech.ac.jp
\end{center}

\vspace{5mm} 

\begin{abstract}
In this study, we propose a novel multi-objective Bayesian optimization (MOBO) method to efficiently identify the Pareto front (PF) defined by risk measures for black-box functions under the presence of input uncertainty (IU). Existing BO methods for Pareto optimization in the presence of IU are risk-specific or without theoretical guarantees, whereas our proposed method addresses general risk measures and has theoretical guarantees. The basic idea of the proposed method is to assume a Gaussian process (GP) model for the black-box function and to construct high-probability bounding boxes for the risk measures using the GP model. Furthermore, in order to reduce the uncertainty of non-dominated bounding boxes, we propose a method of selecting the next evaluation point using a maximin distance defined by the maximum value of a quasi distance based on bounding boxes. As theoretical analysis, we prove that the algorithm can return an arbitrary-accurate solution in a finite number of iterations with high probability, for various risk measures such as Bayes risk, worst-case risk, and value-at-risk. We also give a theoretical analysis that takes into account approximation errors because there exist non-negligible approximation errors (e.g., finite approximation of PFs and sampling-based approximation of bounding boxes) in practice. We confirm that the proposed method outperforms compared with existing methods not only in the setting with IU but also in the setting of ordinary MOBO through numerical experiments.
\end{abstract}

\section{Introduction}\label{sec:intro}
In this study, we treat a multi-objective Pareto optimization problem under input uncertainty (IU). 
In many real-world applications such as engineering, industry and computer simulations, it is often desired to simultaneously optimize an expensive-to-evaluate multi-objective black-box function. 
%
%
Because there is typically no point at which all functions are simultaneously optimal, the multi-objective optimization problem is often  formulated as a Pareto optimization problem to identify the Pareto front (PF). 
%
The black-box functions actually handled often have IU. 
%
Our motivating example in this study is an expensive-to-evaluate docking simulation for real-world chemical compounds. 
The purpose of this simulation is to evaluate the inhibitory performance of candidate compounds on specific sites of some target protein. 
Because each compound has uncertain isomers, this simulation is expressed as the Pareto optimization problem under IU. 

We consider a multi-objective black-box function optimization problem with $M$ objective functions under IU with $m \in \{1, 2, \ldots, M\}$. 
Let $f^{(m)}({\bm x}, {\bm w})$ be the $m$-th objective function, where ${\bm x} \in \mathcal{X}$ and ${\bm w} \in \Omega$ are called \textit{design variables} and \textit{environmental variables}, respectively.
%
%
The variable ${\bm x}$ is an input that can be completely controlled, whereas ${\bm w}$ is a random variable that cannot be controlled and follows some probability distribution. 
When considering a Pareto optimization problem in the presence of IU, it is necessary to consider optimization by taking into account the uncertainty of ${\bm w}$ that cannot be controlled. 
A \textit{risk measure} is the widely used function that is determined based on only ${\bm x}$ while considering the uncertainty of ${\bm w}$. 
Various risk measures, for example, Bayes risk, worst-case risk and value-at-risk, are used depending on the problem. 
%
%
Given a risk measure $F^{(m)} ({\bm x} ) \equiv \rho^{(m)}  (f^{(m)} ({\bm x},{\bm w})   )$, the problem that we treat in this study is formulated as  
\begin{align*}
{\rm optimize} \ (F^{(1)} ({\bm x} ), \ldots , F^{(M)} ({\bm x} )) \ {\rm s.t.} \ {\bm x} \in \mathcal{X}.
\end{align*}
%
%
%
%
%

Bayesian optimization (BO) \citep{shahriari2015taking} using Gaussian processes (GPs) \citep{GPML} is a powerful tool for optimizing black-box functions. 
Many BO methods have been proposed for both single-objective and multi-objective black-box functions without IU.
In contrast, designing BO methods for risk measures under the presence of IU is challenging. 
This is because risk measures cannot be observed directly  and do not generally follow GPs even if black-box functions follow GPs.
%
%
%
The main way to solve this problem is to design a predicted region that may contain a black-box function, then compute the risk measure on the region and use the lower and upper bounds of this to construct a predicted interval for the risk measure \citep{nguyen2021value,nguyen2021optimizing,pmlr-v108-kirschner20a}. 
%
%
As an exception, special risk measures such as Bayes risk  are known to follow GP in practice, allowing Bayesian quadrature (BQ)-based  inference \citep{beland2017bayesian}. 
Recently, multi-objective Bayesian optimization (MOBO) methods under IU have been proposed, which apply the BQ-based or predicted interval-based method 
  \citep{qing2023robust,iwazaki2020mean,rivier2022surrogate}. 
However, the BQ-based method proposed by \cite{qing2023robust} and Mean-variance-analysis (MVA)-based method proposed by \cite{iwazaki2020mean}  can only be applied to specific risk measures, and the surrogate-assisted bounding-box approach (SABBa) proposed by \cite{rivier2022surrogate} is a heuristic with no theoretical guarantee for the construction of the predicted region (interval) instead of being applicable to general risk measures. 

In this study, we propose a novel MOBO method based on \textit{high-probability bounding boxes (HPBBs)}  for risk measures using GP surrogate models, which solves the above problem. 
The basic idea of the proposed method is to design a high-probability credible region (HPCR) that contains a black-box function with high probability. 
We use the fact that many risk measures can be expressed as a composite of a tractable function and some monotonic function, and construct high-probability credible intervals (HPCIs) of risk measures as a transformation of the lower and upper bounds of the tractable function. 
We also propose a method for computing a sampling-based CI of risk measures on the HPCR. 
%
Furthermore, we provide theoretical guarantees for these two methods in the case with/without various approximation errors that may occur in the practical computation. 
Through these results, we can propose a theoretically guaranteed MOBO methods for general risk measures.
The characteristics of the proposed method and the representative existing methods are given in Table \ref{tab:proposed_and_previous}.  
Our contributions can be summarized as follows:
\begin{itemize}
\item We develop a general method for designing HPBB that can be applied to various risk measures. 
  \item We propose a novel acquisition function (AF) for MOBO under IU, which effectively incorporates the quantified uncertainty of Pareto optimal solutions using HPBB.
   \item We provide theoretical guarantees of accuracy and termination based on HPBB and the proposed AF, as well as a theoretical error analysis that accounts for various types of approximation errors that can occur in the practical computation.
\end{itemize}


\paragraph{Related Work}
In the optimization of expensive-to-evaluate black-box functions, BO has gained popularity and has been the subject of active research. 
 A variety of AFs for BO and MOBO settings have been introduced \citep{movckus1975bayesian,GPUCB,wang2017max,emmerich2008computation,svenson2010multiobjective,zuluaga2016varepsilon,knowles2006parego,suzuki2020multi}. 
%
%
%
Moreover, multi-objective optimization has also been extensively studied in the evolutionary computation community \citep{deb2002fast}. 
However, methodologies based on evolutionary computation often necessitate several thousand to tens of thousands of function evaluations \citep{deb2005searching,zhou2018multi}, which can be prohibitively costly. 

Studies on Pareto optimization under IU have also been gradually proposed in recent years, mainly in the development of BO methods to efficiently identify the PF defined by risk measures. 
Considered risk measures are, for example, Bayes risk  \citep{qing2023robust}, mean and negative standard deviation \citep{iwazaki2020mean}, and general risk measures \citep{rivier2022surrogate}. 
%
%
%
However, as mentioned earlier, these   are methods that risk-specific or  without theoretical guarantees. 
A BO method for a multivariate value-at-risk (MVaR) has also been proposed \citep{daulton2022robust}. 
This method is similar to other MOBO methods, but is very different in essence. 
In general, in Pareto optimization, the PF is defined as the boundary defined by the points satisfying Pareto optimality, i.e., the points define the PF. 
On the other hand, MVaR is itself a PF, and the PF considered in \cite{daulton2022robust} is defined as the boundary of the union of MVaR. 
Therefore, in \cite{daulton2022robust}, although the problem setup is Pareto optimization, the final PF is defined by PFs (MVaR). 
Thus, we only introduce it here because it differs from Pareto optimization in the essential point. 

\begin{table*}[tb]
  \begin{center}
    \caption{Characteristics of the proposed method and the representative existing methods}
    \begin{tabular}{c||c|c|c|c|c} \hline \hline
       & Proposed &  SABBa & BQ & MVA & MOBO without IU  \\ \hline 
  IU setting  & Yes  & Yes  &   Yes    & Yes & No    \\  
  General risk setting & Yes  & Yes  &   No    & No &No    \\                    
  Theoretical guarantees & Yes  & No  &   No    & Yes & Yes/No    \\  
    Approximation error setting  & Yes  & No  &   No    & No & No    \\                   \hline  \hline
    \end{tabular}
    \label{tab:proposed_and_previous}
  \end{center}
\end{table*}


\section{Preliminary}\label{sec:preliminary}
\paragraph{Problem Setup} 
Let $f^{(m)}: \mathcal{X} \times \Omega \to \mathbb{R}$ be an expensive-to-evaluate black-box function, where $ m \in \{1,2, \ldots , M \} \equiv [M]$. 
Assume that the set of design variables $\mathcal{X} $ and set of environmental variables $\Omega $ are compact and convex. 
For each $({\bm x},{\bm w} ) \in \mathcal{X} \times \Omega $, $f^{(m) } ({\bm x},{\bm w} ) $ is observed with noise as $y^{(m)} = f^{(m) } ({\bm x},{\bm w} ) + \varepsilon^{(m)} $, where  $\varepsilon^{(m)}$ follows normal distribution with mean 0 and variance $\varsigma^2 _m $, and  the sequence of noises  $( \varepsilon ^{(m)} _i )_{i \in \mathbb{N}, m \in [M] } $ is independent. 
In this study, ${\bm w} \in \Omega$ follows some distribution $P_w$, and   
 $( \varepsilon ^{(m)} _i )_{i \in \mathbb{N}, m \in [M] } $ and $({\bm w} _i ) _{ i \in \mathbb{N}} $ are mutually independent. 
In the BO framework including environment variables, two different settings for ${\bm w}$ exist called the \textit{simulator setting} and the \textit{uncontrollable setting} \citep{pmlr-v108-kirschner20a,iwazaki2020mean,pmlr-v162-inatsu22a}. 
In the simulator setting, ${\bm w}$ is fully controllable during optimization, whereas in the uncontrollable setting, ${\bm w}$ is not controllable even during optimization. 
In the main body, only the simulator setting is treated, and the uncontrollable setting is discussed in Appendix \ref{app:generalized}. 
Let $\rho^{(m)}( f^{(m)} ({\bm x} ,{\bm w} ) ) \equiv F^{(m)} ({\bm x}  ) $ be a risk measure. 
For example, the widely used Bayes and worst-case risks are given by $F^{(m)} ({\bm x}  ) = \mathbb{E}[f^{(m)} ({\bm x},{\bm w})]$ and $F^{(m)} ({\bm x} ) = \inf_{ {\bm w} \in \Omega} f^{(m)}  ({\bm x},{\bm w})  $, respectively, where the expectation is taken with respect to ${\bm w}$. 
The purpose of this study is to efficiently identify the PF  defined based on  $F^{(m)} ({\bm x} ) $. 
For any ${\bm x} \in \mathcal{X} $ and $E \subset \mathcal{X} $, let ${\bm F} ({\bm x} ) =  (F^{(1)} ({\bm x} ) , \ldots , F^{(M)} ({\bm x} ) )$ and ${\bm F} (E) =  \{  {\bm F}  ({\bm x} )  \mid {\bm x}  \in E \}$. 
Then, for any $ B \subset \mathbb{R}^M $,  the dominated region ${\rm Dom} (B)$ and PF  ${\rm Par} (B) $ of $B$ are  defined as $
{\rm Dom} (B) = \{  {\bm s} \in \mathbb{R}^M \mid ^\exists {\bm s}^\prime \in B \ {\rm s.t.} \ {\bm s} \leq {\bm s}^\prime \} $ and $
{\rm Par} (B) =  \partial ({\rm Dom} (B) ) 
$. 
Here, for any vector  
${\bm a} =(a_1 , \ldots, a_M), {\bm b} = (b_1,\ldots , b_M) \in \mathbb{R} ^M$ and set $C$,  ${\bm a} \leq {\bm b} $ represents 
$a_m \leq b_m $ for any $m \in [M]$, and  $\partial (C)$ represents the boundary of $C$. 
Let  $Z^\ast$ be our target PF.
 Then,   $Z^\ast$ can be expressed as 
$
Z^\ast =   {\rm Par }   ({\bm F}   (\mathcal{X}))
$.

\paragraph{Regularity Assumption}   
We introduce a \textit{regularity assumption} for $f^{(m)} $. 
For each $ m \in [M]$, 
let $k^{(m)} : ( \mathcal{X} \times \Omega )   \times  ( \mathcal{X} \times \Omega )   \to \mathbb{R}$ be a positive-definite kernel, where $ k^{(m)} (  ({\bm x},{\bm w} ) ,  ({\bm x},{\bm w} ) )  \leq 1 $ for any   
 $({\bm x} ,{\bm w} )  \in \mathcal{X} \times \Omega $. 
Also let  $ \mathcal{H}   (k^{(m)}) $ be a reproducing kernel Hilbert space corresponding to  $k^{(m)} $.  
We assume that $f^{(m)} $ is the element of $ \mathcal{H}   (k^{(m)}) $ and has the bounded Hilbert norm $\|  f^{(m)}  \|  _{ \mathcal{H}   (k^{(m)})  } \leq B_m    < \infty$. 

\paragraph{Gaussian Process Model} 
In this study, we use a GP   model for the black-box function $f^{(m)}$.
We assume the GP $\mathcal{G}\mathcal{P}(0, k^{(m)}(  ( {\bm x},{\bm w}  ),  ( {\bm x}^\prime,{\bm w}^\prime  )  )       )$ as the prior of $f^{(m) } $. 
For $m \in [M]$, given a dataset $\{ ({\bm x}_i,{\bm w}_i,y^{(m)}_i )\}_{i=1}^t$, where $t$ is the number of queried instances, the posterior  of $f^{(m)}$ is a GP. 
Then,  its posterior mean $ \mu^{(m)}_t ({\bm x},{\bm w}) $ and posterior variance $ \sigma^{(m) 2}_t ({\bm x},{\bm w}) $ can be calculated analytically \citep{GPML}. 

\section{Proposed Method}\label{sec:proposed_method}
In this section, we propose a BO method to efficiently identify $Z^\ast$. 
%
%
%
%
In Section \ref{sec:CI and BB}, we provide a method for computing the CI of $F^{(m)} ({\bm x} )$ using the CI of $f^{(m)} ({\bm x},{\bm w} )$. 
We also give  a bounding box for ${\bm F} ({\bm x} )$, which is the direct product of CIs.

\subsection{Credible Interval and Bounding Box}\label{sec:CI and BB}
For each input $({\bm x} ,{\bm w} ) \in \mathcal{X} \times \Omega $ and $t \geq 1$, the CI of $f^{(m)} ({\bm x},{\bm w} )$ is denoted by $Q^{(f^{(m)})}_{t-1} ({\bm {x}},{\bm w}) =[ l^{(f^{(m)})}_{t-1} ({\bm {x}},{\bm w}), u^{(f^{(m)})}_{t-1} ({\bm {x}},{\bm w})]$, where
     $l^{(f^{(m)})}_{t-1} ({\bm {x}},{\bm w})$ and  $u^{(f^{(m)})}_{t-1} ({\bm {x}},{\bm w})$ are given by 
\begin{align*}
l^{(f^{(m)})}_{t-1} ({\bm {x}},{\bm w}) = \mu^{(m)}_{t-1} ({\bm {x}},{\bm w}) - \beta^{1/2}_{m,t} \sigma^{(m)}_{t-1} ({\bm {x}},{\bm w}) ,\  
 u^{(f^{(m)})}_{t-1} ({\bm {x}},{\bm w}) = \mu^{(m)}_{t-1} ({\bm {x}},{\bm w}) + \beta^{1/2}_{m,t} \sigma^{(m)}_{t-1} ({\bm {x}},{\bm w})
.
\end{align*}
Here, $\beta^{1/2}_{m,t} \geq 0$ is a user-specified tradeoff parameter. 
If we set $\beta^{1/2}_{m,t}$ appropriately, $Q^{(f^{(m)})}_{t-1} ({\bm {x}},{\bm w}) $ becomes a HPCI which contains  
$f^{(m)} ({\bm x},{\bm w} )$ with high probability  (details are described in  Section \ref{sec:theoretical guarantees}). 
For ${\bm x} \in \mathcal{X}$, $t \geq 1$ and $m \in [M]$, we define the set of functions $G^{(m)} _{t-1} ({\bm x} ) $ as 
%
$$
G^{(m)} _{t-1} ({\bm x} ) = \{    g({\bm x},{\bm w}  )  \mid ^\forall {\bm w} \in \Omega , g({\bm x},{\bm w}  ) \in Q^{(f^{(m)})}_{t-1} ({\bm {x}},{\bm w}) \}. 
$$
Let $Q^{(F^{(m)})}_{t-1} ({\bm {x}}) = [{\rm lcb }^{(m)}_{t-1}  ({\bm x}  )  , {\rm ucb }^{(m)}_{t-1}  ({\bm x}  ) ]$ be a CI of  $F^{(m)}  ({\bm x}  ) $. 
Also let $
B_{t-1} ({\bm x} ) = Q^{(F^{(1)})}_{t-1} ({\bm {x}}) \times \cdots \times  Q^{(F^{(M)})}_{t-1} ({\bm {x}})
$ be a bounding box of 
${\bm F} ({\bm x} )$. 
Then, when $Q^{(f^{(m)})}_{t-1} ({\bm {x}},{\bm w}) $ is HPCI for all $m \in [M]$, $t \geq 1$, ${\bm x} \in \mathcal{X}$ and ${\bm w} \in \Omega$, a sufficient condition for $Q^{(F^{(m)})}_{t-1} ({\bm {x}})$ to also be HPCI is given as follows:
\begin{equation}
\begin{split}
&^\forall  g({\bm x}  ,{\bm w}  )   \in G^{(m)} _{t-1} ({\bm x} ), \\
&{\rm lcb }^{(m)}_{t-1}  ({\bm x}  )  \leq    \rho^{(m)}   (  g({\bm x}  ,{\bm w}  )   ) 
\leq {\rm ucb }^{(m)}_{t-1}  ({\bm x}  ).  
\end{split}
\label{eq:HPCI condition1} 
\end{equation}
If \eqref{eq:HPCI condition1}  holds, then   $B_{t-1} ({\bm x} ) $ is also a HPBB. 
Next, we provide computation methods for ${\rm lcb }^{(m)}_{t-1}  ({\bm x}  ) $ and ${\rm ucb }^{(m)}_{t-1}  ({\bm x}  ) $. 
First, we provide a \textit{generalized  method} for 
  ${\rm lcb }^{(m)}_{t-1}  ({\bm x}  ) $ and ${\rm ucb }^{(m)}_{t-1}  ({\bm x}  ) $ to satisfy \eqref{eq:HPCI condition1}. 
The  ${\rm lcb }^{(m)}_{t-1}  ({\bm x}  ) $ and ${\rm ucb }^{(m)}_{t-1}  ({\bm x}  ) $ by the generalized  method are calculated with 
\begin{align*}
{\rm lcb }^{(m)}_{t-1}  ({\bm x}  )  &= \inf _{  g({\bm x}  ,{\bm w}  )   \in G^{(m)} _{t-1} ({\bm x} ) }     \rho^{(m)}   (  g({\bm x}  ,{\bm w}  )   ) ,\\ 
{\rm ucb }^{(m)}_{t-1}  ({\bm x}  )  &= \sup _{  g({\bm x}  ,{\bm w}  )   \in G^{(m)} _{t-1} ({\bm x} ) }     \rho^{(m)}   (  g({\bm x}  ,{\bm w}  )   ).
\end{align*}
%
We emphasize that although the condition \eqref{eq:HPCI condition1} holds by using the generalized method, the inf and sup calculations in the generalized method are not always easy. 
%
%
Therefore, in this study, we give additional two computation methods for 
${\rm lcb }^{(m)}_{t-1}  ({\bm x}  ) $ and ${\rm ucb }^{(m)}_{t-1}  ({\bm x}  ) $, (i) \textit{decomposition method} and (ii) \textit{sampling method}. 
In the decomposition method, ${\rm lcb }^{(m)}_{t-1}  ({\bm x}  )$ and ${\rm ucb }^{(m)}_{t-1}  ({\bm x}  )$ in \eqref{eq:HPCI condition1} are calculated directly. 
Let $\rho (\cdot)$ be a risk measure. 
In many cases, $\rho (\cdot)$ can be decomposed as $\rho (\cdot)  = \tilde{\rho} \circ h (\cdot)$, where $ \tilde{\rho} (\cdot)$ and $h(\cdot)$ are respectively monotonic and tractable functions. 
The basic idea of the decomposition method is to compute the infimum and supremum of $h (g({\bm x}  ,{\bm w}  ))$ on $G^{(m)} _{t-1} ({\bm x} ) $, and then to compute  ${\rm lcb }^{(m)}_{t-1}  ({\bm x}  ) $ and ${\rm ucb }^{(m)}_{t-1}  ({\bm x}  ) $ by taking $\tilde{\rho} (\cdot)$ to these. 
Calculated values for several risk measures are given  in  Table \ref{tab:decomposition_result}. 
%
%
In the sampling method, we generate $S$ sample paths $f^{(m)} _1 ({\bm x} ,{\bm w} ) ,\ldots ,
f^{(m)} _S ({\bm x} ,{\bm w} ) $ of $f^{(m)} ({\bm x} ,{\bm w} )$ independently from the GP posterior  and compute 
\begin{align*}
{\rm lcb }^{(m)}_{t-1}  ({\bm x}  ) 
 &= \min_{  j \in [S], f^{(m)} _j ({\bm x} ,{\bm w} ) \in G^{(m)} _{t-1} ({\bm x} )    }      \rho ^{(m)}  (f^{(m)} _j ({\bm x} ,{\bm w} ))  ,\\
 {\rm ucb }^{(m)}_{t-1}  ({\bm x}  ) 
& = \max_{  j \in [S], f^{(m)} _j ({\bm x} ,{\bm w} ) \in G^{(m)} _{t-1} ({\bm x} )    }      \rho ^{(m)}  (f^{(m)} _j ({\bm x} ,{\bm w} ))  .
\end{align*}
However, in all cases of  generalized, decomposition and sampling methods, there is a case that \eqref{eq:HPCI condition1} is not satisfied due to approximation errors that may occur in practice, e.g., approximation errors in the expected value computation or insufficient approximation due to a small number of sample paths. 
These problems are discussed in Section \ref{sec:theoretical guarantees}. 

\begin{table}[tb]
  \centering
    \caption{The values of ${\rm lcb }^{(m)} _t ({\bm x}  ) $ and ${\rm ucb }^{(m)} _t ({\bm x}  ) $  for commonly used risk measures}
\scalebox{0.7}{
  \begin{tabular}{c||c|c|c} \hline 
Risk measure & Definition &  ${\rm lcb }^{(m)} _t ({\bm x}  )$  & ${\rm ucb }^{(m)} _t ({\bm x}  )$  \\ \hline \hline
Bayes risk & $\mathbb{E} [  f^{(m)}_{{\bm x},{\bm w} } ] $ & $\mathbb{E}[ l^{(m)}_{t,{\bm x},{\bm w} } ]$ & $\mathbb{E}[ u^{(m)}_{t,{\bm x},{\bm w} } ]$  \\ \hline 
Worst-case &  $\inf_{ {\bm w} \in \Omega  }   f^{(m)}_{{\bm x},{\bm w} }  $    & $ \inf_{ {\bm w} \in \Omega  }  l^{(m)}_{t,{\bm x},{\bm w} }  $& $   \inf_{ {\bm w} \in \Omega  }  u^{(m)}_{t,{\bm x},{\bm w} } $     \\ \hline 
Best-case &  $\sup_{ {\bm w} \in \Omega  }   f^{(m)}_{{\bm x},{\bm w} }    $   & $ \sup_{ {\bm w} \in \Omega  } l^{(m)}_{t,{\bm x},{\bm w} }  $&  $  \sup_{ {\bm w} \in \Omega  }  u^{(m)}_{t,{\bm x},{\bm w} } $       \\ \hline 
$\alpha$-value-at-risk  &        $\inf \{  b \in \mathbb{R} \mid  \alpha \leq \mathbb{P} ( f^{(m)}_{{\bm x},{\bm w} }  \leq b)    \}$               &  $\inf \{  b \in \mathbb{R} \mid \alpha \leq \mathbb{P} ( l^{(m)}_{t,{\bm x},{\bm w} } \leq b)      \}$  &$\inf \{  b \in \mathbb{R} \mid \alpha \leq \mathbb{P} ( u^{(m)}_{t,{\bm x},{\bm w} } \leq  b)      \}$    \\ \hline 
$\alpha$-conditional value-at-risk  & $ \mathbb{E}   [  f^{(m)}_{ {\bm x},{\bm w} } |   f^{(m)}_{ {\bm x},{\bm w} } \leq v_{f^{(m)}} ({\bm x};\alpha )    ]    $&  $\frac{1}{\alpha} \int_0 ^ \alpha  v_{l^{(m)}_{t} } ({\bm x};\alpha^\prime )  {\rm d} \alpha^\prime$        & $\frac{1}{\alpha} \int_0 ^ \alpha  v_{u^{(m)}_{t} } ({\bm x};\alpha^\prime )   {\rm d} \alpha^\prime$  \\ \hline 
Mean absolute deviation &  $  \mathbb{E}[ |  f^{(m)}_{{\bm x},{\bm w} }    -   \mathbb{E}[  f^{(m)}_{{\bm x},{\bm w} }    ]    |   ]    $& $  \mathbb{E}[  \min \{  | \check{l}^{(m)}_{t,{\bm x},{\bm w} }  | ,  | \check{u}^{(m)}_{t,{\bm x},{\bm w} }  |  \}   - {\rm STR}  (\check{l}^{(m)}_{t,{\bm x},{\bm w} } ,\check{u}^{(m)}_{t,{\bm x},{\bm w} } )   ]  $ &$  \mathbb{E}[ \max \{  | \check{l}^{(m)}_{t,{\bm x},{\bm w} }  | ,  | \check{u}^{(m)}_{t,{\bm x},{\bm w} }  |  \}  ] $    \\ \hline 
Standard deviation &  $  \sqrt{\mathbb{E}[ |  f^{(m)}_{{\bm x},{\bm w} }    -   \mathbb{E}[  f^{(m)}_{{\bm x},{\bm w} }    ]    |^2   ]   } $ & $ \sqrt{ \mathbb{E}[  \min \{  | \check{l}^{(m)}_{t,{\bm x},{\bm w} }  |^2 ,  | \check{u}^{(m)}_{t,{\bm x},{\bm w} }  |^2  \}   - {\rm STR} ^2 (\check{l}^{(m)}_{t,{\bm x},{\bm w} } ,\check{u}^{(m)}_{t,{\bm x},{\bm w} } )   ]  }$ &$  \sqrt{\mathbb{E}[ \max \{  | \check{l}^{(m)}_{t,{\bm x},{\bm w} }  |^2 ,  | \check{u}^{(m)}_{t,{\bm x},{\bm w} }  |^2  \}  ] }$  \\ \hline 
Variance & $  \mathbb{E}[ |  f^{(m)}_{{\bm x},{\bm w} }    -   \mathbb{E}[  f^{(m)}_{{\bm x},{\bm w} }    ]    |^2   ]    $ & $  \mathbb{E}[  \min \{  | \check{l}^{(m)}_{t,{\bm x},{\bm w} }  |^2 ,  | \check{u}^{(m)}_{t,{\bm x},{\bm w} }  |^2  \}   - {\rm STR} ^2 (\check{l}^{(m)}_{t,{\bm x},{\bm w} } ,\check{u}^{(m)}_{t,{\bm x},{\bm w} } )   ]  $ &$  \mathbb{E}[ \max \{  | \check{l}^{(m)}_{t,{\bm x},{\bm w} }  |^2 ,  | \check{u}^{(m)}_{t,{\bm x},{\bm w} }  |^2  \}  ] $  \\ \hline 
Distributionally robust&   $\inf _{  P \in \mathcal{A}  }   F^{(m)} ({\bm x}  ;P)$ & $\inf _{  P \in \mathcal{A}  }  {\rm lcb }^{(m)} _t ({\bm x} ;P )      $ &$\inf _{  P \in \mathcal{A}  }  {\rm ucb }^{(m)} _t ({\bm x} ;P )      $   \\ \hline 
Monotonic Lipschitz map & $\mathcal{M}   ( F^{(m)}  ({\bm x}  )  ) $ &         $\min \{  \mathcal{M} ({\rm lcb }^{(m)} _t ({\bm x}  )  )    , \mathcal{M} (     {\rm ucb }^{(m)} _t ({\bm x}  ) )  \}$                & $\max \{  \mathcal{M} ({\rm lcb }^{(m)} _t ({\bm x}  )  )    , \mathcal{M} (     {\rm ucb }^{(m)} _t ({\bm x}  ) )  \}$ \\ \hline 
Weighted sum & $\alpha_1 F^{(m_1 ) }  ({\bm x} ) + \alpha_2 F^{(m_2 ) }  ({\bm x} )$ &       $ \alpha_1 {\rm lcb }^{(m_1)} _t ({\bm x}  )  + \alpha_2 {\rm lcb }^{(m_2)} _t ({\bm x}  ) $     & $ \alpha_1 {\rm ucb }^{(m_1)} _t ({\bm x}  )  + \alpha_2 {\rm ucb }^{(m_2)} _t ({\bm x}  ) $   \\ \hline 
Probabilistic threshold &  $ \mathbb{P}  (    f^{(m)} _{ {\bm x},{\bm w} }  \geq \theta )$   &      $ \mathbb{P}  (    l^{(m)} _{t, {\bm x},{\bm w} }  \geq \theta )$     &  $ \mathbb{P}  (    u^{(m)} _{ t, {\bm x},{\bm w} }  \geq \theta )$    \\ \hline \hline
    \multicolumn{4}{c}{$ f^{(m)}_{{\bm x},{\bm w} } \equiv  f^{(m)} ({\bm x},{\bm w} )$,     $l^{(m)}_{t,{\bm x},{\bm w} } \equiv l^{(f^{(m)})}_t({\bm x},{\bm w} )$ , $u^{(m)}_{t,{\bm x},{\bm w} } \equiv u^{(f^{(m)})}_t({\bm x},{\bm w} )$, $v_{f^{(m)}} ({\bm x};\alpha ) \equiv   \inf \{  b \in \mathbb{R} \mid  \mathbb{P} ( f^{(m)}_{{\bm x},{\bm w} }  \leq b) \geq \alpha    \}$           } \\ 
  \multicolumn{4}{c}{    
 $v_{l^{(m)}_t} ({\bm x};\alpha ) \equiv   \inf \{  b \in \mathbb{R} \mid  \mathbb{P} ( l^{(m)}_{t,{\bm x},{\bm w} }  \leq b) \geq \alpha    \}$, 
 $v_{u^{(m)}_t} ({\bm x};\alpha ) \equiv   \inf \{  b \in \mathbb{R} \mid  \mathbb{P} ( u^{(m)}_{t,{\bm x},{\bm w} }  \leq b) \geq \alpha    \}$    , $ \alpha \in (0,1)   $    }  \\ 
  \multicolumn{4}{c}{    $\check{l}^{(m)}_{t,{\bm x},{\bm w} }  \equiv l^{(m)}_{t,{\bm x},{\bm w} }  - \mathbb{E}  [u^{(m)}_{t,{\bm x},{\bm w} }  ]   $ ,    $\check{u}^{(m)}_{t,{\bm x},{\bm w} }  \equiv u^{(m)}_{t,{\bm x},{\bm w} }  - \mathbb{E}  [l^{(m)}_{t,{\bm x},{\bm w} }  ]   $             , ${\rm STR}   (a,b)  \equiv \max \{ \min \{  -a,b \}, 0 \}  $ } \\  
  \multicolumn{4}{c}{     $ F^{(m)} ({\bm x}  ;P)$: Risk measure $F^{(m)} ({\bm x} )$ defined based on the distribution $P$          } \\  
  \multicolumn{4}{c}{     $  {\rm lcb }^{(m)} _t ({\bm x} ;P )  ,  {\rm ucb }^{(m)} _t ({\bm x} ;P )  $:       $  {\rm lcb }^{(m)} _t ({\bm x}  )  $ and $  {\rm ucb }^{(m)} _t ({\bm x}  )  $ for  $ F^{(m)} ({\bm x}  ;P)$   } \\  
  \multicolumn{4}{c}{     $ q^{(m)} (a ;F^{(m)} )$: a function $q^{(m)} (a) $ for $F^{(m)} ({\bm x} )$, does not depend on $P$   } \\  
  \multicolumn{4}{c}{     $\mathcal{M} (\cdot)$: Monotonic Lipschitz continuous map with a Lipschitz constant $K$          } \\  
  \multicolumn{4}{c}{     $\alpha_1,\alpha_2 \geq 0$          } \\  \hline 
  \multicolumn{4}{c}{    $\alpha$-value-at-risk is the same meaning as $\alpha$-quantile        } \\  \hline \hline
  \end{tabular}
}
\label{tab:decomposition_result}
\end{table}


\subsection{Pareto Front Estimation}
\begin{figure}[t]
\begin{center}
 \begin{tabular}{c}
 \includegraphics[width=1\textwidth]{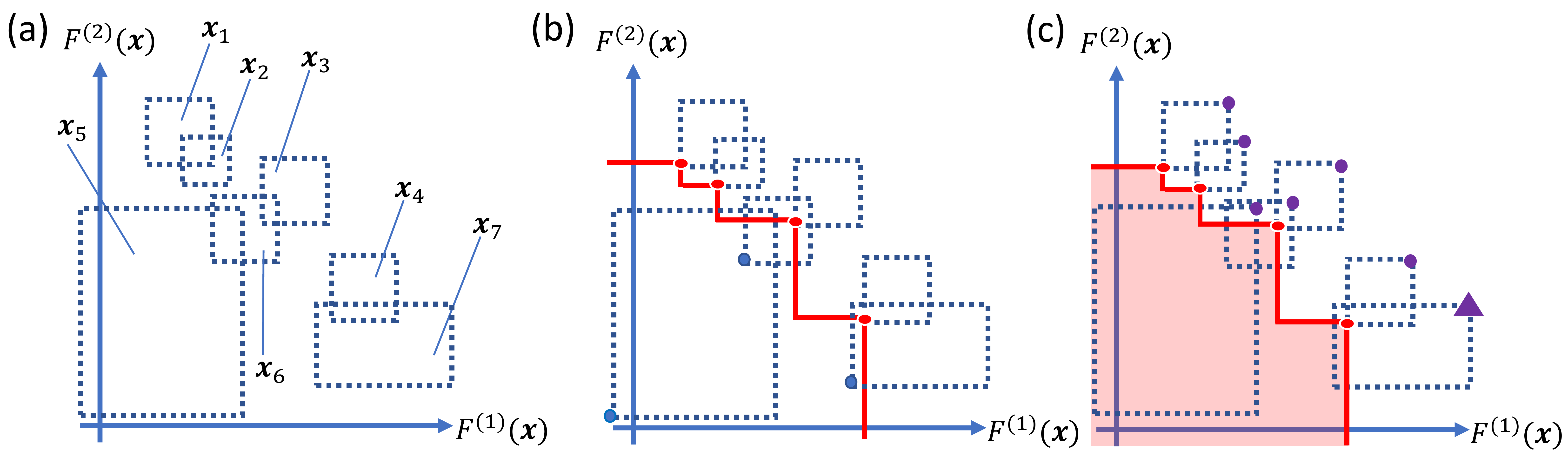} 
 \end{tabular}
\end{center}
 \caption{Conceptual diagrams of $ \text{\bf LCB}_t ({\bm x} )$, $ \text{\bf UCB}_t ({\bm x} )$, $\text{Par} (\text{\bf LCB}_t (\mathcal{X}))$, $\hat{\Pi}_t$ and AFs for seven input points ${\bm x}_1,\ldots, {\bm x}_7$.
At each point ${\bm x}$ in the left figure (a), $ \text{\bf LCB}_t ({\bm x} )$ and $ \text{\bf UCB}_t ({\bm x} )$ indicate the lower left point and the upper right point of the dashed rectangular region, respectively. 
In (b), the PF (red line) computed using each $ \text{\bf LCB}_t ({\bm x} )$ is $\text{Par} (\text{\bf LCB}_t (\mathcal{X}))$, and because  it is constructed by $ \text{\bf LCB}_t ({\bm x} _1) ,   \text{\bf LCB}_t ({\bm x} _2) , \text{\bf LCB}_t ({\bm x} _3) , \text{\bf LCB}_t ({\bm x} _4) $, $\hat{\Pi}_t$ is given by $\hat{\Pi}_t = \{  {\bm x}_1,{\bm x}_2,{\bm x}_3,{\bm x}_4 \}$.
In (c), the light red region indicates $\text{Dom} (\text{\bf LCB}_t (\hat{\Pi}_t )$), the region dominated by the red points ($\text{\bf LCB}_t (\hat{\Pi}_t )$), and $a^{(\mathcal{X})}_t ({\bm x} )$ is the closeness between the light red region and $\text{\bf UCB}_t ({\bm x} )$ (purple point). 
The furthest point is represented by the purple triangle, $\text{\bf UCB}_t ({\bm x}_7 )$.  
Thus, the next design variable to be evaluated is ${\bm x}_7$.
}
\label{fig:alg1}
\end{figure}

For any input ${\bm x} \in \mathcal{X}$ and  subset $E \subset \mathcal{X}$, 
we define $ \text{\bf LCB}_{t-1} ({\bm x} )$, 
$ \text{\bf UCB}_{t-1} ({\bm x} )$ and $ \text{\bf LCB}_{t-1} (E )$ as 
\begin{align*}
\text{\bf LCB}_{t-1} ({\bm x} )&= ( {\rm lcb}^{(1)}_{t-1}  ({\bm x} )  , \ldots ,   {\rm lcb}^{(M)}_{t-1}  ({\bm x} ) ), \ 
 \text{\bf UCB}_{t-1} ({\bm x} )=( {\rm ucb}^{(1)}_{t-1}  ({\bm x} )  , \ldots ,   {\rm ucb}^{(M)}_{t-1}  ({\bm x} ) ), \\
\text{\bf LCB}_{t-1} (E)&=  \{  \text{\bf LCB}_{t-1} ({\bm x} ) \mid {\bm x} \in E \} . 
\end{align*}
The estimated Pareto solution set $\hat{\Pi}_{t-1} \subset \mathcal{X}$ for the design variables is then defined as follows: 
$$
\hat{\Pi}_{t-1} = \{ {\bm x} \in \mathcal{X} \mid  \text{\bf LCB}_{t-1} ({\bm x} ) \in \text{Par} (\text{\bf LCB}_{t-1} (\mathcal{X})) \}
.
$$
Figure \ref{fig:alg1} (a) shows a conceptual diagram of $ \text{\bf LCB}_{t-1} ({\bm x} )$ and $ \text{\bf UCB}_{t-1} ({\bm x} )$, and (b) shows a conceptual diagram of $\text{Par} (\text{\bf LCB}_{t-1} (\mathcal{X}))$ and $\hat{\Pi}_{t-1}$. 
Here, in order to actually compute $\hat{\Pi}_{t-1} $, we need to compute the PF defined by $\text{\bf LCB}_{t-1} ({\bm x} )$. 
However, if $\mathcal{X}$ is an infinite set, then $\hat{\Pi}_{t-1} $ may also be an infinite set. 
In this case, since the exact calculation of $\hat{\Pi}_{t-1} $ is difficult,  it is necessary to make a finite approximation using an approximation solver such as NSGA-II \citep{deb2002fast}.  
The effects on this finite approximation are discussed in Section \ref{sec:theoretical guarantees}.

\subsection{Acquisition Function}
We propose an AF for determining the next point to be evaluated. 
First, for each point ${\bm a} \in \mathbb{R}^m $ and subset $B \subset \mathbb{R}^m$, we denote the quasi distance between them as 
$$
\text{dist} ({\bm a},B) = \min_ { {\bm b} \in B} d_\infty ({\bm a},{\bm b} ), 
$$
where $d_\infty ({\bm a},{\bm b} )$ denotes the metric function given by $d_\infty ({\bm a},{\bm b} ) = \max \{  |a_1-b_1|,\ldots , |a_m-b_m | \} . 
$
Using this, we define  AF $a^{(\mathcal{X})}_t ({\bm x})$ for ${\bm x} \in \mathcal{X}$ as 
$$
a^{(\mathcal{X})}_t ({\bm x} ) =  \text{dist}   (   \text{\bf UCB}_t ({\bm x} ), \text{Dom}  (\text{\bf LCB}_t (\hat{\Pi}_t )) )
.
$$
%
Then, the next design variable, ${\bm x}_{t+1}$, to be evaluated is selected by 
$
{\bm x}_{t+1} = \argmax_{ {\bm x} \in \mathcal{X} }   a^{(\mathcal{X})}_t ({\bm x} ). 
$ 
%
Hence, the value of $   a^{(\mathcal{X})}_t (  {\bm x}_{t+1} )   $ is equal to the following maximin distance:
$$
a^{(\mathcal{X})}_t (  {\bm x}_{t+1} )  
=\max_{ {\bm x} \in \mathcal{X} }   \min_{ {\bm b}  \in  \text{Dom}  (\text{\bf LCB}_t (\hat{\Pi}_t )) } d_\infty ( \text{\bf UCB}_t ({\bm x} ), {\bm b}).
$$
Figure \ref{fig:alg1} (c) shows a conceptual diagram of the AF $a^{(\mathcal{X})}_t ({\bm x} )$. 
The value of $a^{(\mathcal{X})}_t ({\bm x})$ can be computed analytically using the following lemma when $ \hat{\Pi}_t $ is finite:
\begin{lemma}\label{lem:AF_cal}
Let $ \text{\bf UCB}_t ({\bm x} ) = (u_1,\ldots , u_M)$ and $\text{\bf LCB}_t (\hat{\Pi}_t ) = \{ (l^{(i)}_1,\ldots , l^{(i)}_M ) \mid 1 \leq i \leq k \}$. 
Then, $a^{(\mathcal{X})}_t ({\bm x})$ can be computed by $a^{(\mathcal{X})}_t ({\bm x} ) = \max \{ \tilde{a}_t ({\bm x} ) , 0 \} $, where 
$$
\tilde{a}_t ({\bm x} ) = \min_{1 \leq i \leq k } \max \{ u_1 - l^{(i)}_1,\ldots ,  u_M - l^{(i)}_M \}.
$$
\end{lemma}
The proposed AF is based on the bounding box as well as existing bounding box-based AFs  \citep{iwazaki2020mean,zuluaga2016varepsilon,belakaria2020uncertainty}, but differs in the following points. 
Most of the existing methods focus only on reducing the size of the non-dominated bounding box \footnote{The bounding box $B_t ({\bm x} )$ with $    \text{\bf UCB}_t ({\bm x} ) \notin \text{Dom}  (\text{\bf LCB}_t (\hat{\Pi}_t )) $.}  (e.g., diagonal length and hypervolume),  and therefore do not aim to improve the estimated PF (size-based AFs choose $x_5$ in Fig. \ref{fig:alg1}, but the room for improvement is small). 
 Hence, these AFs focus on exploration. 
In contrast, the proposed AF focuses on the non-dominated bounding box with the largest maximin distance to the estimated PF.
 In this sense, the proposed AF focuses on  exploration, but also exploitation.

Next, we consider the choice of the environment variable ${\bm w}_{t+1} $. 
The variable ${\bm w}_{t+1}$ should be determined based on the uncertainty of the chosen bounding box $B_t ({\bm x}_{t+1} )$. 
 We define the uncertainty of $B_t ({\bm x}_{t+1} )$ by the maximum length of each edge $ \| \text{\bf UCB}_t ({\bm x}_{t+1} ) - \text{\bf LCB}_t ({\bm x}_{t+1} ) \|_\infty$. 
In many risk measures including Bayes risk, the following inequality holds:
\begin{equation}
\|  \text{\bf UCB}_t ({\bm x}_{t+1} ) -  \text{\bf LCB}_t ({\bm x}_{t+1} ) \|_\infty 
\leq q  ( \zeta_{t+1} ) ,
\label{eq:w_condition}
\end{equation}
where $q(\cdot ): [0,\infty ) \to [0, \infty )$ is a strictly increasing function defined by risk measures and satisfies $q(0) =0$, and $ \zeta_{t+1} =  \max_{{\bm w} \in \Omega} \sum_{m=1}^M  2 \beta^{1/2}_{m,t+1} \sigma^{(m)}_t ({\bm x}_{t+1},{\bm w}  )   $. 
Then, we choose ${\bm w}_{t+1}$ based on \eqref{eq:w_condition}. 
The next environmental variable, ${\bm w}_{t+1}$, to be evaluated is selected by ${\bm w}_{t+1} = \argmax _{ {\bm w} \in \Omega }   a^{(\Omega)}_t ({\bm w} )$, where 
$
   a^{(\Omega)}_t ({\bm w} )=\sum_{m=1}^M  2 \beta^{1/2}_{m,t+1} \sigma^{(m)}_t ({\bm x}_{t+1},{\bm w}  )$. 
%

\subsection{Stopping Condition}
We describe the stopping conditions of the proposed algorithm. 
From Fig. \ref{fig:alg1} (c), AF $a^{(\mathcal{X})}_t ({\bm x} )$ represents the closeness of the pessimistic PF and the optimistic predictive value of ${\bm F} ({\bm x} )$. 
That is, if this value is sufficiently small, there is little room for improvement in the PF; therefore, it is reasonable to use it as the stopping condition. 
Let $\epsilon >0$ be a predetermined desired accuracy parameter.
Then the algorithm is terminated if $ a^{(\mathcal{X})}_t ({\bm x}_{t+1} ) \leq \epsilon $ is satisfied. 
The pseudocode of the proposed algorithm is described in Algorithm \ref{alg:1}.

\begin{algorithm}[t]
    \caption{Bounding box-based MOBO of general risk measures}
    \label{alg:1}
    \begin{algorithmic}
        \REQUIRE GP priors $\mathcal{GP}(0,\ k^{(m)})$, tradeoff parameters $\{\beta_{m,t}\}_{t \geq 0}$, $m \in [M]$, accuracy parameter $\epsilon >0$ 
        \FOR { $ t= 0,1,2,\ldots $}
            \STATE Compute $Q^{(f^{(m)})}_{t} ({\bm x},{\bm w} )$  for all $m \in [M]$ and $(\bm{x}, {\bm w} )  \in \mathcal{X} \times \Omega$
            \STATE Compute $Q^{(F^{(m)})}_{t} ({\bm x} )$  for all $m \in [M]$ and $\bm{x}  \in \mathcal{X} $ by the generalized, decomposition or sampling method
		\STATE Compute $B_t ({\bm x} ) = Q^{(F^{(1)})}_{t} ({\bm x} ) \times \cdots \times Q^{(F^{(M)})}_{t} ({\bm x} ) $ for each ${\bm x} \in \mathcal{X}$ 
		\STATE Estimate $\hat{\Pi}_t $ by $B_t ({\bm x} )$
            \STATE Select the next evaluation point $\bm{x}_{t+1}$ by $a^{(\mathcal{X})}_{t}  ({\bm x} ) $  
		\IF {$  a^{(\mathcal{X})}_{t} ({\bm x}_{t+1})  \leq  \epsilon$}
		\STATE break
		\ENDIF
            \STATE Select the next evaluation point $\bm{w}_{t+1}$ by $a^{(\Omega)}_{t}  ({\bm w} ) $
            \STATE Observe $y^{(m)}_{t+1} = f^{(m)}(\bm{x}_{t+1}, \bm{w}_{t+1}) + \varepsilon^{(m)}_{t+1}$  at the point $(\bm{x}_{t+1}, \bm{w}_{t+1})$ for all $m \in [M]$
            \STATE Update GPs by adding observed points
        \ENDFOR
        \ENSURE Return $\hat{\Pi}_{t}$ as the estimated set of design variables
    \end{algorithmic}
\end{algorithm}

\section{Theoretical Analysis}\label{sec:theoretical guarantees}
In this section, we give the theorems for the accuracy and termination of the proposed algorithm. 
The details of the proofs are presented in Appendix \ref{app:proofs}. 
First, we quantify the goodness of the estimated $\hat{\Pi}_t$. 
If  $\hat{\Pi}_t$ is a good estimate, the following two indicators defined by $\hat{\Pi}_t$ should be small:
\begin{align*}
I^{(i)}_t &=  \max_{ {\bm y}  \in  Z^\ast   } {\rm dist}  ({\bm y} ,  {\rm Par} ({\bm F} (\hat{\Pi}_t)) ), \\ 
I^{(ii)}_t &=  \max_{ {\bm y}  \in  {\bm F} (\hat{\Pi}_t)     } {\rm dist}  ({\bm y} , Z^\ast )  
.
\end{align*}
Here, $I^{(i)}_t $ and $I^{(ii)}_t$ have similar meanings as recall and precision in the classification problem, respectively. 
For example, if $\hat{\Pi}_t$ is estimated as $\hat{\Pi}_t = \mathcal{X}$, $\hat{\Pi}_t$  contains all of true Pareto optimal design variables. 
In this case, since $ {\rm Par} ({\bm F} (\hat{\Pi}_t)) = {\rm Par} ({\bm F} (\mathcal{X})) = Z^\ast$, $I^{(i)}_t =0$. 
Similarly, when $\hat{\Pi}_t $ is estimated as $\hat{\Pi}_t = \{ {\bm x}^\ast _1 \} $, where ${\bm x}^\ast _1$ is one of true Pareto optimal design variables,  
$\hat{\Pi}_t $ does not have unnecessary points,  and $I^{(ii)}_t =0$.
As with recall and precision in ordinary classification problems, over (under)-estimation makes $I^{(ii)}_t $ ($I^{(i)}_t $) larger. 
For this reason, 
we define the \textit{inference discrepancy} $I_t = \max \{ I^{(i)}_t, I^{(ii)}_t \}$ for $\hat{\Pi}_t$ as  the goodness measure. 
Next, in order to show the theoretical validity of the proposed algorithm, we introduce the maximum information gain $\kappa^{(m)}_t $. 
This indicator is frequently used in theoretical analysis in the context of GP-based BOs and can be expressed as 
$$
\kappa ^{(m)}_t = 2^{-1}   \max_{  (\tilde{\bm x}_1, \tilde{\bm w}_1), \ldots ,  (\tilde{\bm x}_t, \tilde{\bm w}_t)   } \log {\rm det } ( {\bm I}_t + \varsigma^{-2}_m \tilde{\bm K}^{(m)}_t),
$$
where ${\bm I}_t$ is the $t \times t$ identity matrix, and $\tilde{\bm K}^{(m)}_t$ is the $t \times t $ matrix whose $(j,k)$-th element is $k^{(m)}   (    (\tilde{\bm x}_j, \tilde{\bm w}_j) , (\tilde{\bm x}_k, \tilde{\bm w}_k)          )$. 
It is known that the order of  $\kappa^{(m)} _t $ with respect to commonly used kernels such as linear, Gaussian and Mat\'{e}rn kernels is 
sublinear under mild conditions  (see, e.g., Theorem 5 in \cite{GPUCB}). 
Then, the following theorem holds:
\begin{lemma}[Theorem 3.11 in \cite{abbasi2013online}]\label{lem:HPCI}
Suppose that the regularity assumption holds. 
Let $\delta \in (0,1 )$, and define 
$$
\beta^{1/2}_{m,t} = B_m + \sqrt{ 2 \left (  \kappa^{(m)}_t + \log \frac{M}{\delta}  \right ) }. 
$$
Then, with probability at least $1-\delta$, the following inequality holds for any $t \geq 1 $, $m \in [M]$ and $({\bm x},{\bm w} ) \in \mathcal{X} \times \Omega $: 
$$
|f^{(m)} ({\bm x},{\bm w} )  - \mu^{(m)}_{t-1}   ({\bm x},{\bm w} )  | \leq \beta^{1/2}_{m,t} \sigma^{(m)}_{t-1}  ({\bm x},{\bm w} ).
$$
\end{lemma}
Using this, we give the following theorems for  the inference discrepancy, stopping condition and $q(a)$: 
\begin{theorem}\label{thm:inference discrepancy}
Suppose that the assumption of Lemma \ref{lem:HPCI} and the inequality \eqref{eq:HPCI condition1}  hold.
Let $t \geq 0$, $m \in [M]$, $\delta \in (0,1)$, and let $\beta^{1/2}_{m,t+1} $ be defined as in Lemma \ref{lem:HPCI}. 
In addition, let $\epsilon >0$ be a  predetermined desired accuracy parameter. 
Then, with probability at least $1-\delta$, the inequality $I_{t} \leq  a^{(\mathcal{X})}_{t} ({\bm x}_{t+1} ) $ holds for any $t \geq 0$ and ${\bm x}_{t+1}$. 
Therefore, if  the stopping condition 
  satisfies at $T$ iterations, the inference discrepancy $I_{T} $ satisfies $I_{T} \leq \epsilon $ with probability at least $1-\delta$.  
\end{theorem}
%
%
%
\begin{theorem}\label{thm:termination}
Suppose that the assumption in Theorem \ref{thm:inference discrepancy} holds. 
Let $q: [0,\infty) \to [0,\infty) $ be a strictly increasing function satisfying $q(0)=0$ and \eqref{eq:w_condition}. 
Also let 
$$
s_t = \sqrt{ \frac{\sum_{m=1} ^M    C_m \beta_{m,t+1} \kappa^{(m)}_{t+1} }{  t+1}  }  ,    
$$ where $C_m = \frac{8M} { \log (1+ \varsigma^{-2}_m) }$. 
Then, the inequality $  a^{(\mathcal{X})}_{\hat{t}} ({\bm x}_{{\hat{t}}+1} )  \leq q(s_t) $ holds for any $t \geq 0$ and some $ \hat{t} \leq t$. 
Therefore, the algorithm 
 terminates after at most $T$ iterations, where $T$ is the smallest positive integer satisfying $q(s_T) \leq \epsilon$. 
\end{theorem}
%
%
\begin{theorem}\label{thm:q}
Suppose that the assumption in Theorem \ref{thm:inference discrepancy} holds. 
Also assume that there exist strictly increasing functions $q^{(m)}: [0,\infty ) \to [0,\infty )$ 
satisfying $q^{(m)} (0) =0 $ and 
$$
| {\rm ucb}^{(m)}_{t} ({\bm x}_{t+1} ) -  {\rm lcb}^{(m)}_{t} ({\bm x}_{t+1} )  | 
 \leq q^{(m)}   ( \tilde{s}_t ) 
$$
for any $t \geq 0$, $m \in [M]$,  and $ {\bm x}_{t+1} \in \mathcal{X}$, where $\tilde{s}_t =  
\max_{ {\bm w} \in \Omega} 2 \beta^{1/2}_{m,t+1} \sigma^{(m)}_{t} ({\bm x}_{t+1} , {\bm w}  )
$. 
Then,  $q (a) \equiv \max_{m \in [M] }  q^{(m)} (a) $ is the strictly increasing function and satisfies $q(0) =0$ and \eqref{eq:w_condition}. 
\end{theorem}
Specific forms of $q^{(m)} (a )$ for commonly used risk measures are described in Table \ref{tab:q}.

\begin{table}[tb]
  \centering
    \caption{Specific forms of $q^{(m)} (a)$  for commonly used risk measures}
  \begin{tabular}{c||c|c} \hline 
Risk measure & Definition  & $q^{(m)} (a) $ \\ \hline \hline
Bayes risk & $\mathbb{E} [  f^{(m)}_{{\bm x},{\bm w} } ] $  & $a$ \\ \hline 
Worst-case &  $\inf_{ {\bm w} \in \Omega  }   f^{(m)}_{{\bm x},{\bm w} }  $     &   $  a  $  \\ \hline 
Best-case &  $\sup_{ {\bm w} \in \Omega  }   f^{(m)}_{{\bm x},{\bm w} }    $   & $  a $   \\ \hline 
$\alpha$-value-at-risk  &        $\inf \{  b \in \mathbb{R} \mid  \alpha \leq \mathbb{P} ( f^{(m)}_{{\bm x},{\bm w} }  \leq b)    \}$               & $a$  \\ \hline 
$\alpha$-conditional value-at-risk  & $ \mathbb{E}   [  f^{(m)}_{ {\bm x},{\bm w} } |   f^{(m)}_{ {\bm x},{\bm w} } \leq v_{f^{(m)}} ({\bm x};\alpha )    ]    $&  $a$ \\ \hline 
Mean absolute deviation &  $  \mathbb{E}[ |  f^{(m)}_{{\bm x},{\bm w} }    -   \mathbb{E}[  f^{(m)}_{{\bm x},{\bm w} }    ]    |   ]    $&  $2a$  \\ \hline 
Standard deviation &  $  \sqrt{\mathbb{E}[ |  f^{(m)}_{{\bm x},{\bm w} }    -   \mathbb{E}[  f^{(m)}_{{\bm x},{\bm w} }    ]    |^2   ]   } $ &  $\sqrt{8B_m a + 5 a^2 }$ \\ \hline 
Variance & $  \mathbb{E}[ |  f^{(m)}_{{\bm x},{\bm w} }    -   \mathbb{E}[  f^{(m)}_{{\bm x},{\bm w} }    ]    |^2   ]    $ &   $8B_m a + 5 a^2 $\\ \hline 
Distributionally robust&   $\inf _{  P \in \mathcal{A}  }   F^{(m)} ({\bm x}  ;P)$ & $q^{(m)}(a;F^{(m)})$ \\ \hline 
Monotonic Lipschitz map & $\mathcal{M}   ( F^{(m)}  ({\bm x}  )  ) $ &    $K   q^{(m)} (a) $\\ \hline 
Weighted sum & $\alpha_1 F^{(m_1 ) }  ({\bm x} ) + \alpha_2 F^{(m_2 ) }  ({\bm x} )$ & $\alpha_1 q^{(m_1) } (a) + \alpha_2 q^{(m_2 ) } (a)$  \\ \hline 
Probabilistic threshold &  $ \mathbb{P}  (    f^{(m)} _{ {\bm x},{\bm w} }  \geq \theta )$   &    - \\ \hline \hline
    \multicolumn{3}{c}{$ f^{(m)}_{{\bm x},{\bm w} } \equiv  f^{(m)} ({\bm x},{\bm w} )$,     $v_{f^{(m)}} ({\bm x};\alpha ) \equiv   \inf \{  b \in \mathbb{R} \mid  \mathbb{P} ( f^{(m)}_{{\bm x},{\bm w} }  \leq b) \geq \alpha    \}$       , $\alpha \in (0,1)$    } \\ 
  \multicolumn{3}{c}{     $ F^{(m)} ({\bm x}  ;P)$: Risk measure $F^{(m)} ({\bm x} )$ defined based on the distribution $P$          } \\  
  \multicolumn{3}{c}{     $ q^{(m)} (a ;F^{(m)} )$: a function $q^{(m)} (a) $ for $F^{(m)} ({\bm x} )$, does not depend on $P$   } \\  
  \multicolumn{3}{c}{     $\mathcal{M} (\cdot)$: Monotonic Lipschitz continuous map with a Lipschitz constant $K$          } \\  
  \multicolumn{3}{c}{     $\alpha_1,\alpha_2 \geq 0$          } \\  \hline 
  \multicolumn{3}{c}{    $\alpha$-value-at-risk is the same meaning as $\alpha$-quantile        } \\  \hline \hline
  \end{tabular}
\label{tab:q}
\end{table}

\subsection{Theoretical Error Analysis}
In this subsection, we give an extension of Theorem \ref{thm:inference discrepancy} and \ref{thm:termination} when approximation errors  are included in the algorithm. 
In practice, the algorithm includes the following approximation errors:  
(i) Errors in the computation of ${\rm lcb}^{(m)}_{t-1} ({\bm x} ), {\rm ucb}^{(m)}_{t-1} ({\bm x} ) $,  
(ii) errors in computing $\hat{\Pi}_{t-1} $ due to the finite approximation of the estimated PF, and 
(iii) computational errors in maximizing the AFs $a^{(\mathcal{X})}_{t-1} ({\bm x} )$ and $a^{(\Omega)}_{t-1} ({\bm w} )$. 
Let $\epsilon_{{\rm lcb}}, \epsilon_{{\rm ucb}}, \epsilon_{{\rm PF}}, \epsilon_{\mathcal{X}}, \epsilon_{\Omega}$ be non-negative error parameters that represent the errors in these approximations, respectively. 
We consider the case that the following four error inequalities hold for any 
$t \geq 0$, $m \in [M] $, ${\bm x}, {\bm x}_{t+1} \in \mathcal{X}$,  $ {\bm w}_{t+1} \in \Omega$ and $g({\bm x}  ,{\bm w}  )   \in G^{(m)} _{t} ({\bm x} )$: 
\begin{align*}
{\rm lcb }^{(m)}_{t}  ({\bm x}  ) - \epsilon_{{\rm lcb}}  \leq    \rho^{(m)}   (  g({\bm x}  ,{\bm w}  )   ) &\leq 
{\rm ucb }^{(m)}_{t}  ({\bm x}  ) + \epsilon_{{\rm ucb}}, \\
\max_{   {\bm y}  \in {\rm Par}  ( \text{\bf LCB}_{t }(\hat{\Pi}_{t}  ) )   } {\rm dist} ({\bm y} ,  {\rm Par}  ( \text{\bf LCB}_{t }(\mathcal{X}  ) )  ) & \leq \epsilon_{{\rm PF}} , \\
 \max_{ {\bm x}  \in \mathcal{X} }  a^{(\mathcal{X})} _{t}  ({\bm x} )    -   a^{(\mathcal{X})} _{t}  ({\bm x}_{t+1} ) &\leq \epsilon_{\mathcal{X}} , \\
 \max_{ {\bm w}  \in \Omega }  a^{(\Omega)} _{t}  ({\bm w} )    -   a^{(\Omega)} _{t}  ({\bm w}_{t+1} )   &\leq \epsilon_{\Omega} . 
\end{align*}
These inequalities imply that the difference between the desired and actual calculated values is less than the error parameter. 
In this case, a desirable property is that these error parameters simply add to the inequalities in Theorem \ref{thm:inference discrepancy} and \ref{thm:termination}. 
Here, we must emphasize that it is not obvious whether the above is true or not.
This is because the inference discrepancy is defined by the  combination of operations such as the computation of bounding box and the estimation of $\hat{\Pi}_{t-1}$, and it is not obvious how the approximation error affects the inequality.  
The next theorem shows how these approximation errors affect the inequalities:
\begin{theorem}\label{thm:inference discrepancy with errors}
Suppose that the assumption in Lemma \ref{lem:HPCI} holds. 
Let $t \geq 0$, $m \in [M]$, $\delta \in (0,1)$, and let $\beta^{1/2}_{m,t+1} $ be defined as in Lemma \ref{lem:HPCI}. 
In addition, let $\epsilon >0$ be a  predetermined desired accuracy parameter. 
Moreover, let $\epsilon_{{\rm lcb}}, \epsilon_{{\rm ucb}}, \epsilon_{{\rm PF}}, \epsilon_{\mathcal{X}}, \epsilon_{\Omega}$ be non-negative error parameters satisfying the error inequalities.  
Then, with probability at least $1-\delta$, the inequality $I_{t} \leq  a^{(\mathcal{X})}_{t} ({\bm x}_{t+1} )  +  \epsilon_{{\rm lcb}} + \epsilon_{{\rm ucb}}  + \epsilon_{\mathcal{X}}$ holds for any $t \geq 0$ and ${\bm x}_{t+1}$. 
Therefore, if  the stopping condition 
  satisfies at $T$ iterations, the inference discrepancy $I_{T} $ satisfies  $I_{T} \leq \epsilon +  \epsilon_{{\rm lcb}} + \epsilon_{{\rm ucb}}  + \epsilon_{\mathcal{X}}$ with probability at least $1-\delta$.  
\end{theorem}

\begin{theorem}\label{thm:termination with errors}
Suppose that the assumption in Theorem \ref{thm:inference discrepancy with errors} holds. 
Let $q: [0,\infty) \to [0,\infty) $ be a strictly increasing function satisfying $q(0)=0$ and \eqref{eq:w_condition}. 
Then, the inequality $  a^{(\mathcal{X})}_{\hat{t}} ({\bm x}_{{\hat{t}}+1} )  \leq \epsilon_{{\rm PF}} + q(\epsilon_{\Omega} + 
s_t
) $ holds for any $t \geq 0$ and some $ {\hat{t}} \leq t$, where $C_m$ and $s_t$ are given by Theorem \ref{thm:termination}. 
Therefore, the algorithm 
 terminates after at most $T$ iterations, where $T$ is the smallest positive integer satisfying $ \epsilon_{{\rm PF}} + q(\epsilon_{\Omega} + 
s_T 
) \leq \epsilon$. 
%
%
%
\end{theorem}
  
%
Note that for Theorem \ref{thm:termination with errors}, the integer $T$ satisfying the theorem's last inequality does not always exist. 
However, the left hand side in  this inequality is merely an upper bound of $a^{(\mathcal{X} ) } _{\hat{t}} ({\bm x}_{{\hat{t}}+1} ) $. 
Thus, in some cases the actual value of  $a^{(\mathcal{X} ) } _{\hat{t}} ({\bm x}_{{\hat{t}}+1} ) $   satisfies $a^{(\mathcal{X} ) } _{\hat{t}} ({\bm x}_{{\hat{t}}+1} )  \leq \epsilon$ and the stopping condition is satisfied.
%

\section{Numerical Experiments}\label{sec:exp}
In this section, we confirm the performance of the proposed method using synthetic functions and real-world docking simulations. 
For all experiments, we used Gaussian kernels and GP models. 
Experimental details and additional experiments 
 are described in Appendix \ref{app:exp_details}.

\subsection{Synthetic Function}
We confirm the performance of the proposed method through synthetic  functions. 
Although the proposed method is constructed under the presence of IU, the algorithm itself can be applied even when there is no IU. 
Therefore, in the synthetic function experiments, we compared the proposed method with existing MOBO methods without (with) IU. 

\begin{figure*}[tb]
\begin{center}
 \begin{tabular}{cccc}
 \includegraphics[width=0.225\textwidth]{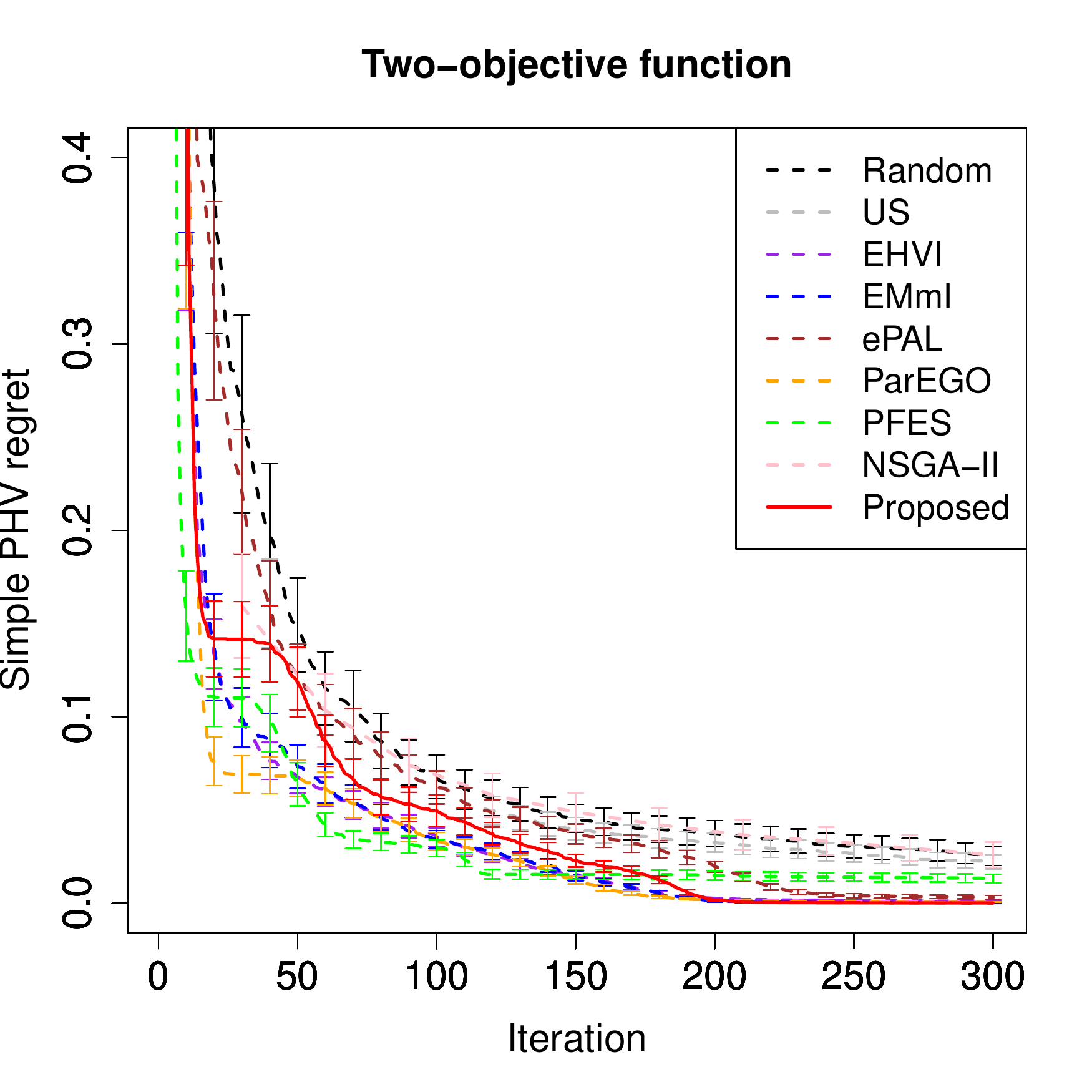} &
 \includegraphics[width=0.225\textwidth]{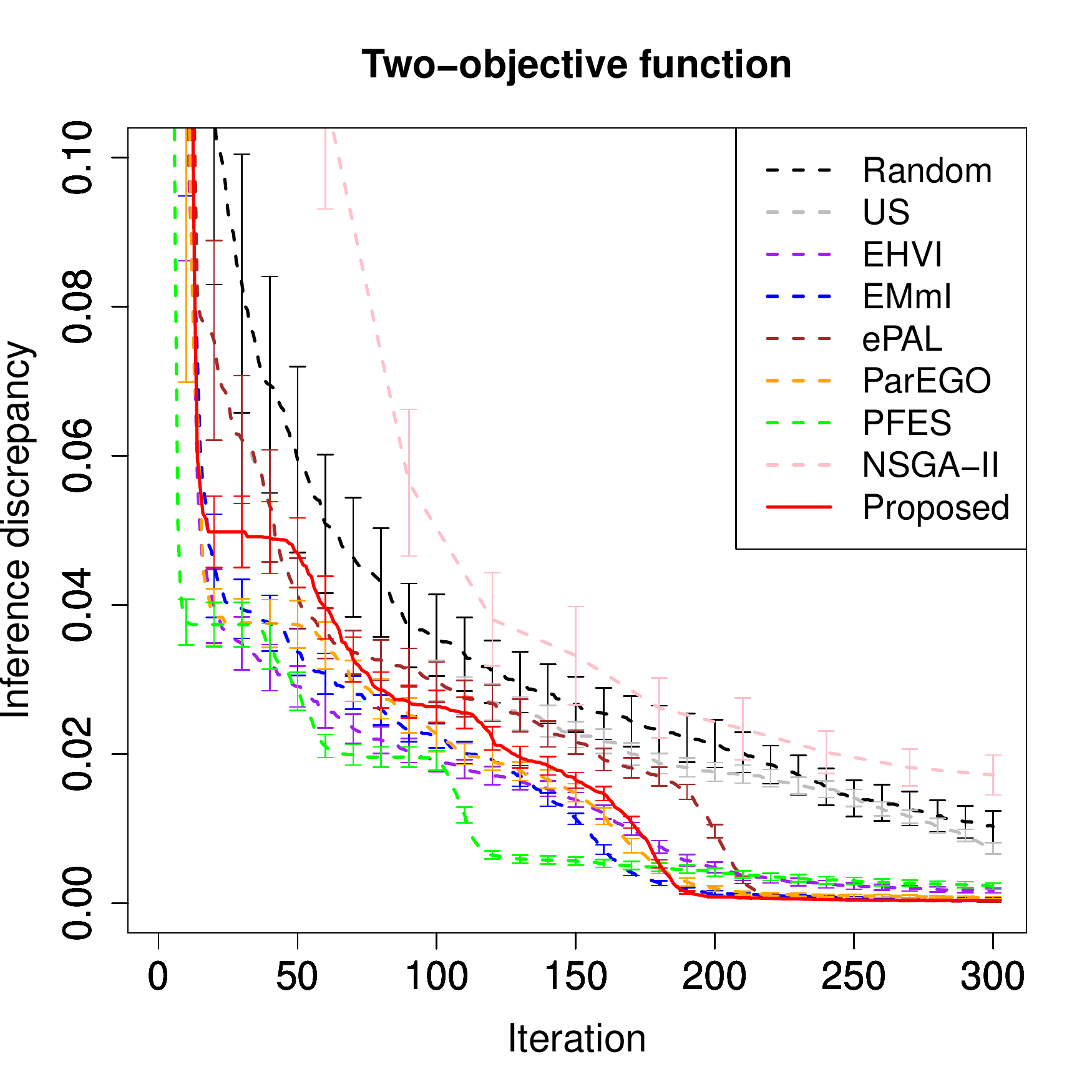} &
 \includegraphics[width=0.225\textwidth]{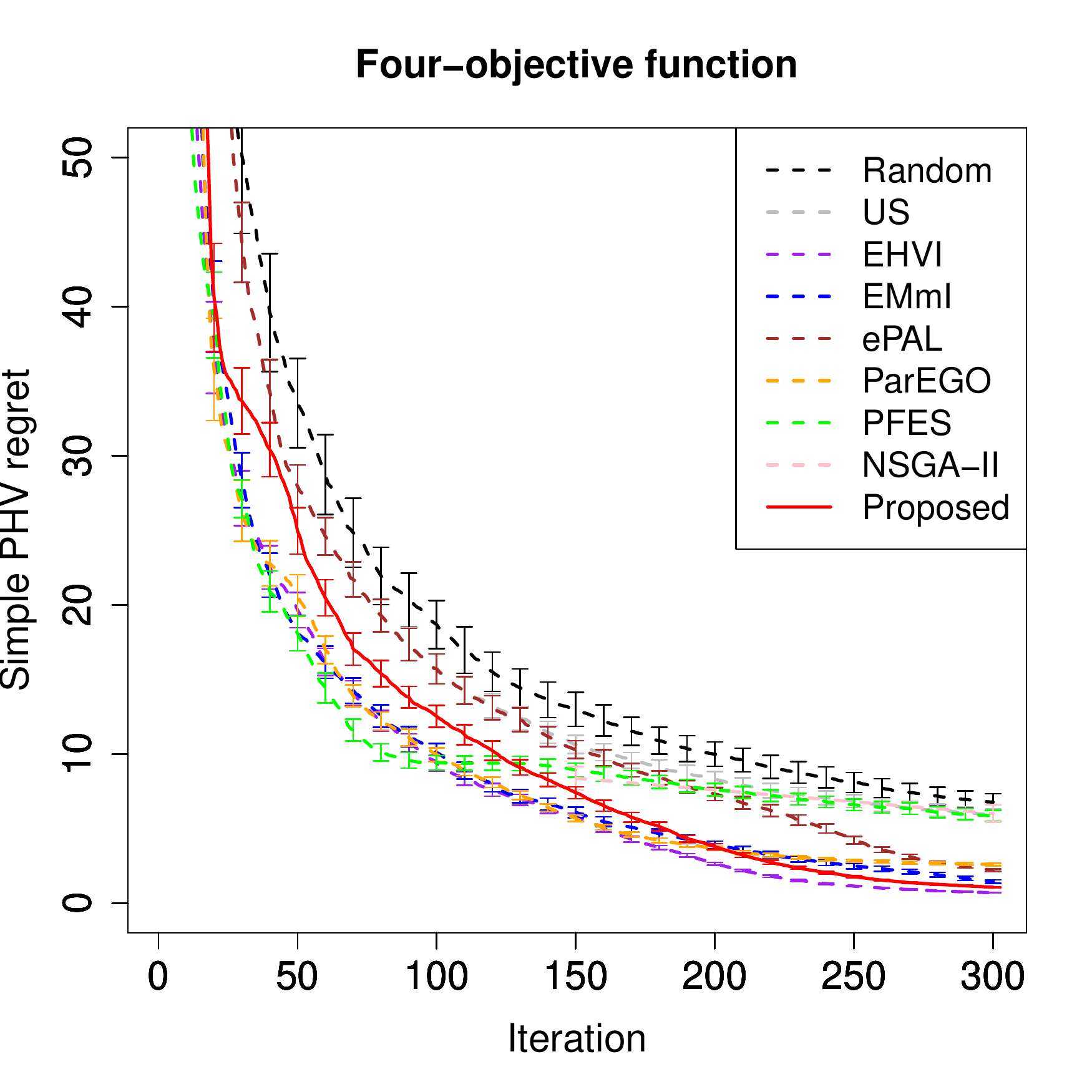} &
 \includegraphics[width=0.225\textwidth]{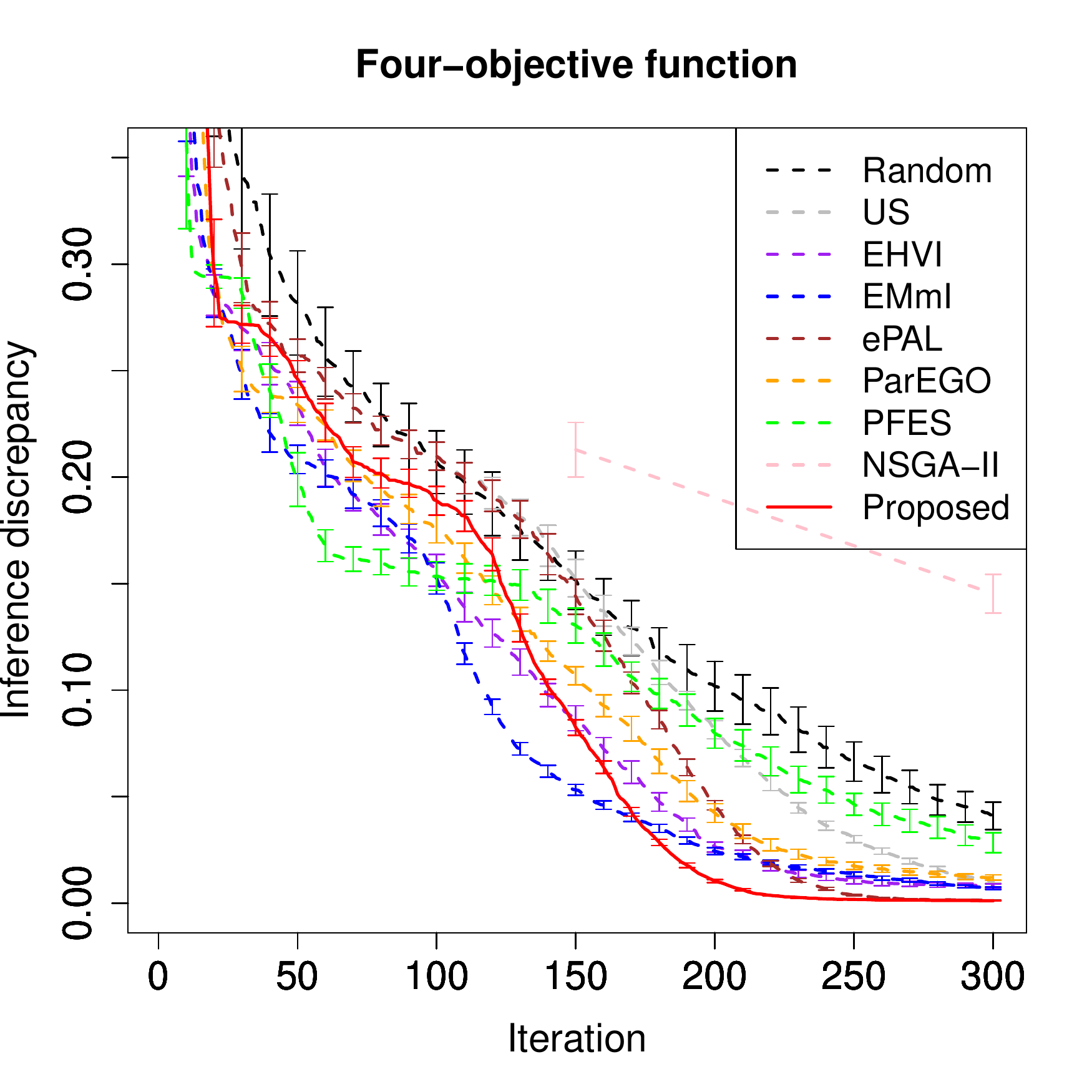} \\
 \includegraphics[width=0.225\textwidth]{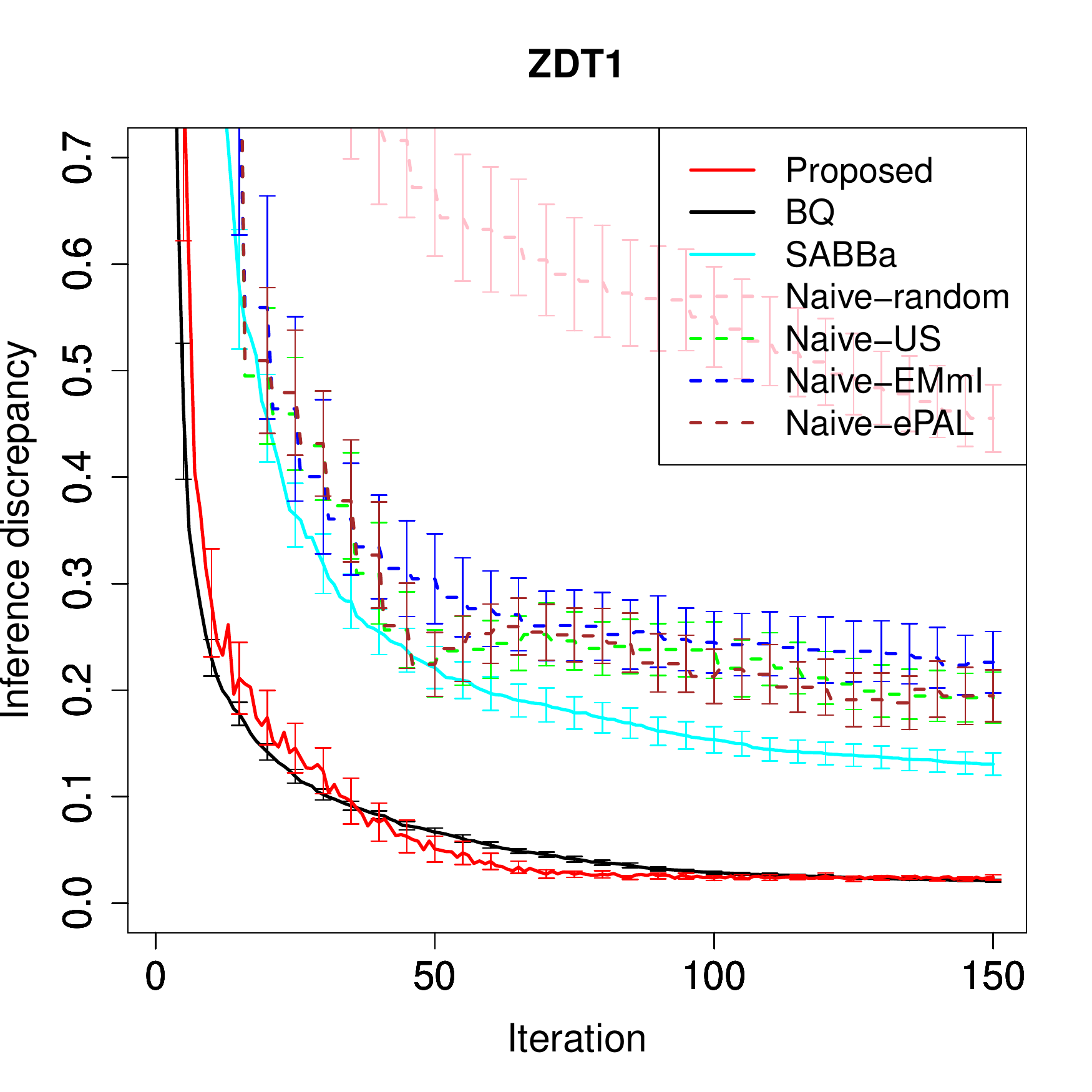} &
 \includegraphics[width=0.225\textwidth]{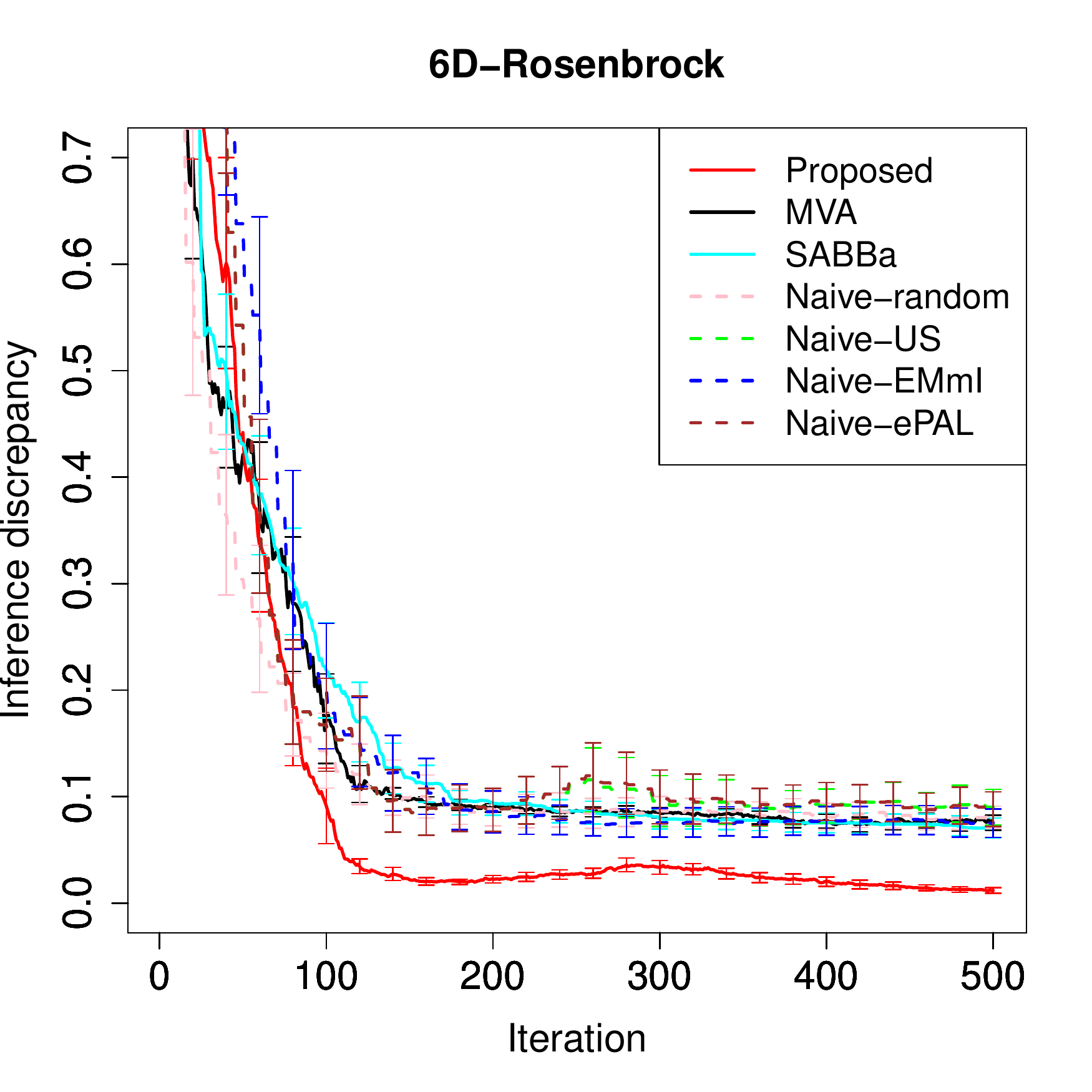} &
 \includegraphics[width=0.225\textwidth]{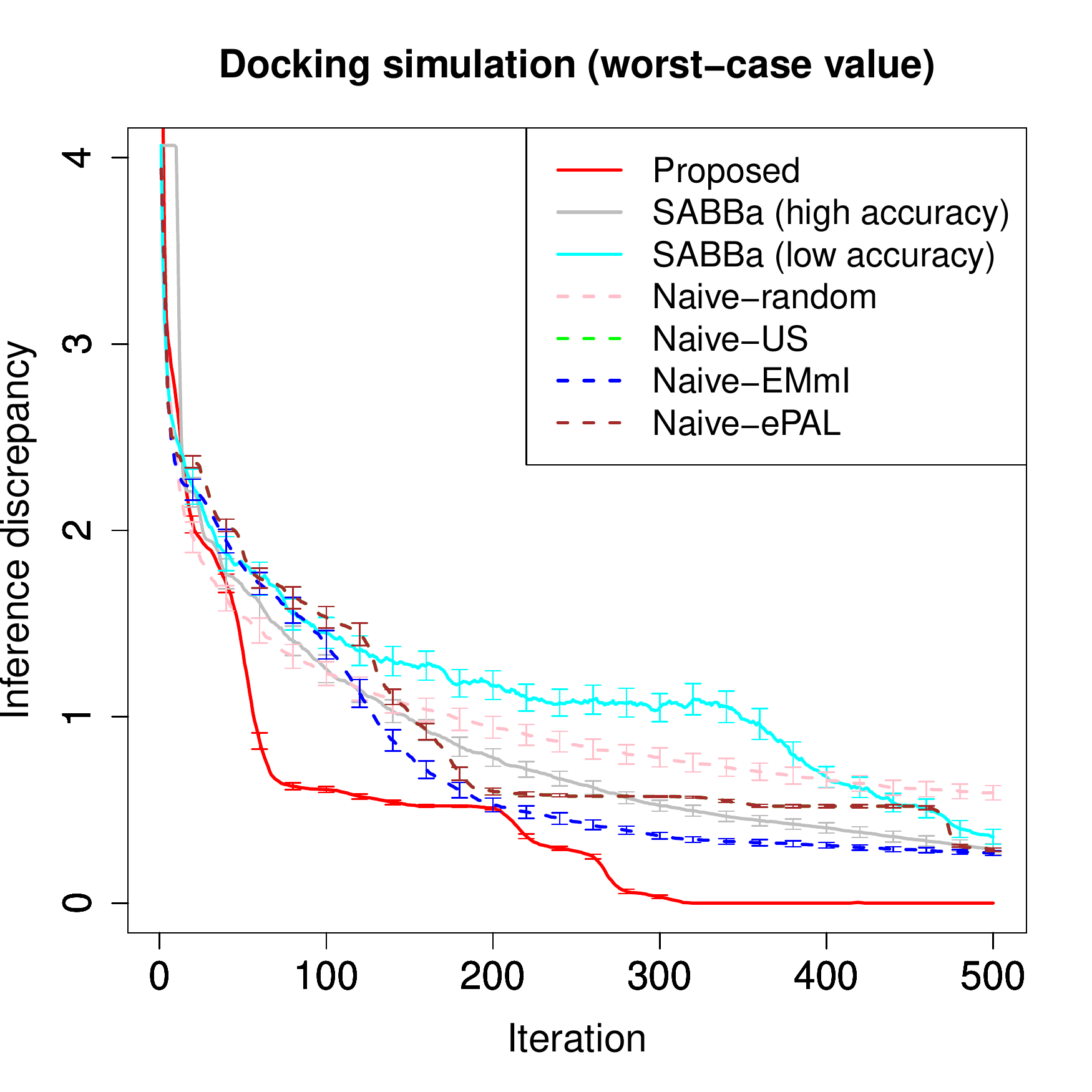} &
 \includegraphics[width=0.225\textwidth]{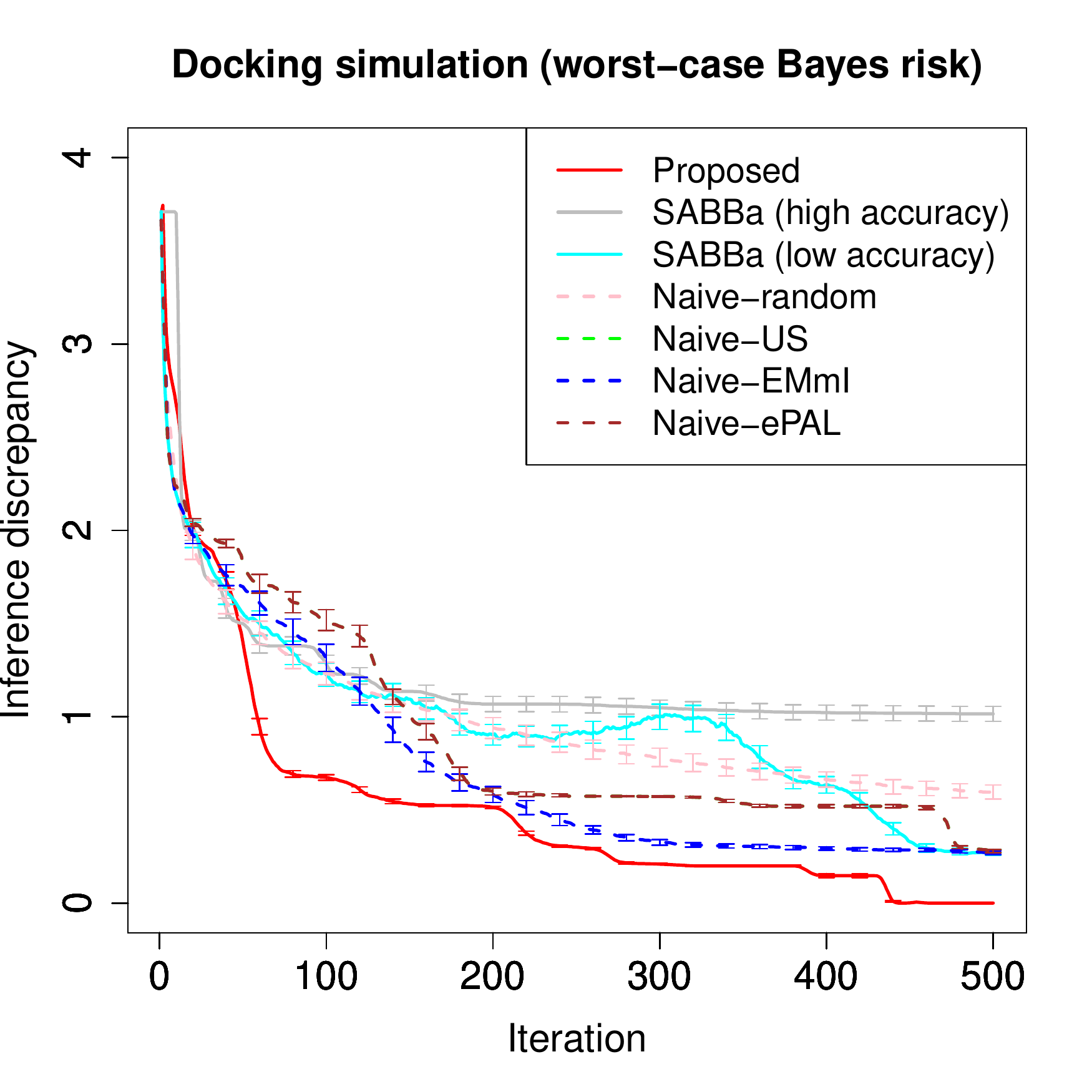} 
 \end{tabular}
\end{center}
 \caption{Comparison with MOBO methods. 
Solid (and dashed) lines are averages of the evaluation measures (simple PHV regret and inference discrepancy) for each iteration in  100, 920 or 429 trials. 
Each error bar length represents the six times the standard error. 
In the top row, the left two columns represent the two-objective setup and the right two columns represent the four-objective setup.
In the bottom row, the left two columns respectively represent the ZDT1 and six-dimensional Rosenbrock setups in the synthetic experiment, and  the right two columns respectively represent WC and WCBR setups in  the real-world docking simulation. 
}
\label{fig:exp_all}
\end{figure*}

In the experiments under no IU, the input space $\mathcal{X} $ was a set of grid points divided into $[-5,5] \times [-5,5]$ equally spaced at $50 \times 50$. 
For black-box functions, we used Booth, Matyas, Himmelblau's and McCormic benchmark functions. 
We performed a two-objective optimization using the first two and a four-objective optimization using all four.  
As evaluation indicators, we used the simple Pareto hypervolume (PHV) regret, which is a commonly used indicator in the context of MOBOs, and inference discrepancy. 
As AFs, we  considered the random sampling (Random), uncertainty sampling (US), EHVI \citep{emmerich2008computation}, EMmI \citep{svenson2010multiobjective}, ePAL \citep{zuluaga2016varepsilon}, ParEGO \citep{knowles2006parego}, PFES \citep{suzuki2020multi} and proposed AF (Proposed). 
We also compared the commonly used evolutionary computation-based method NSGA-II \citep{deb2002fast}. 
Under this setup, one initial point was taken at random and the algorithm was run until the number of iterations reached 300. 
This simulation repeated 100 times and the average simple PHV regret and inference discrepancy at each iteration were calculated.
From the top of Fig. \ref{fig:exp_all}, it can be confirmed that the performance at the end of 300 iterations is comparable or better than the existing methods except for the simple PHV regret in the four-objective setting. 
In particular, the proposed method significantly outperforms other methods for inference discrepancy in the four-objective setting after about 180 iterations.

In the experiment under IU, the input space $\mathcal{X} \times \Omega $ was a compact subset, and we considered  infinite and finite set settings.
We set $\mathcal{X} \times \Omega =[0.25,0.75]^2 \times [-0.25,0.25]^2$ in the infinite set setting. 
In the finite setting,  $\mathcal{X} \times \Omega$ was a set of grid points divided into $[-1,1]^3 \times [-1,1]^3$ equally spaced at $7^3 \times 7^3=117649$. 
The black-box function in the infinite setting was used the ZDT1 benchmark function $\text{\bf ZDT1} ({\bm a} ) \in \mathbb{R}^2$ 
 with a two-dimensional input ${\bm a} $, and the environmental variable ${\bm w}$ was used as the input noise for ${\bm x}$. 
Thus, our considered black-box function was defined by $\text{\bf ZDT1} ({\bm x} +{\bm w}) $. 
We assumed ${\bm w}$ was the uniform distribution on $\Omega$ and used the Bayes risk $\mathbb{E} [\text{\bf ZDT1}  ({\bm x} +{\bm w})] $. 
On the other hand, the black-box function in the finite setting was used the six-dimensional Rosenbrock function $f(w_1,w_2,x_1,x_2,x_3,w_3) \in \mathbb{R}$. 
We assume that  ${\bm w}$ was a discretized normal distribution on $\Omega$. 
As risk measures, we used the expectation 
 and negative standard deviation. 
As comparison methods, we considered the BQ-based method  \citep{qing2023robust}, MVA-based method  \citep{iwazaki2020mean} and SABBa-based method  \citep{rivier2022surrogate}. 
%
Furthermore, four naive methods, Naive-random, Naive-US, Naive-EMmI and Naive-ePAL, were used for comparison.
In the naive methods, ${\bm w}$ was generated five times from the same ${\bm x}$ in one iteration $t$, and the sample mean and the negative square root of the sample variance of the black-box function values were calculated. 
By using ${\bm x}$ and these values, the experiments in naive four methods were  performed as a usual MOBO. 
The name after ``Naive-'' means the name of the used AF.
We used the inference discrepancy as the evaluation indicator. 
Under this setup, one initial point was taken at random and the algorithm was run until the number of iterations reached 150 and 500. 
This simulation repeated 100 times and the average inference discrepancy at each iteration were calculated. 
From the  bottom  of Fig. \ref{fig:exp_all}, it can be confirmed that the proposed method achieves the same or better performance as the existing methods. 
In particular, the results are comparable to those of BQ, which is a limited method applicable only to the Bayes risk case.

\subsection{Real-world Docking Simulation}\label{sec:exp_docking}
In this subsection, we applied the proposed method to docking simulations for real-world chemical compounds. 
The purpose of this simulation is to evaluate the inhibitory performance of candidate compounds on two specific sites of the target protein ``KAT1'', the structure of this protein is available at \url{https://pdbj.org/mine/summary/6v1x}, and to enumerate the Pareto optimal compounds in the presence of structural uncertainty (isomers). 
%
We used the software suite {\em Schr\"{o}dinger} \citep{schrodinger2021} to calculate docking scores and explanatory variables in the compounds. 
%
%
%
%
%
%
Each  compound $C_i$ may have an  isomer $S_{ij}$, and in this simulation the maximum number of  isomers  was limited to 10.
 For each $i$, we computed a 51-dimensional isomer-independent design variables ${\bm x}_i $ and a 51-dimensional environment variable ${\bm w}_{ij} $ that can vary with isomers, using explanatory  variables of $(C_i,S_{ij})$. 
Thus, the black-box functions, the docking scores in two sites, can be expressed as  $f^{(1)} ({\bm x}_i,{\bm w}_{ij} )$ and  $f^{(2)}  ({\bm x}_i,{\bm w}_{ij} )$, respectively. 
We emphasize that the number of isomers $N_i$ was not same for all $i$. 
As risk measures for $C_i$, we considered the worst-case (WC) 
 and worst-case Bayes risk (WCBR). 
For each compound, WC is defined as the minimum docking scores, 
and WCBR is defined as the minimum weighted average of docking scores in predefined candidate weights. 
%
%
The total number of compounds was 429, and the total number of data including isomers was 920. 
We compared the 
SABBa, Proposed and naive four methods. 
In the SABBa method, we considered two different accuracy parameter settings, a high accuracy model and a low accuracy model.
In addition, in the naive four  methods, we calculated docking scores for all isomers in the compound $C_i$ at iteration $t$ and determined the exact risk values.  
In this experiment, the observation noise was zero. 
Under this setup, one initial point was taken from the data and the algorithm was run until the number of iterations reached 500. 
In 
SABBa and Proposed, by changing the initial point, this simulation repeated 920 times. 
Similarly, in naive methods, by changing the initial compound, this simulation repeated 429 times. 
We calculated the average inference discrepancy at each iteration. 
From the bottom in Fig.\ref{fig:exp_all}, we can confirm that the proposed method is superior to other methods.  
In addition, only the proposed method correctly identifies the true PF at the end of 500 iterations for all  risk measures at all 920 different initial points.
Specifically, after 425 iterations for WC 
 and 465 iterations for WCBR, the true PF is identified for all 920 different initial points. 
Therefore, compared to the exhaustive search, the number of iterations required to find the true PF can be reduced to about half. 
Thus, the sample efficient decision making was achieved 
 in our motivating example.

%
%
%

\section{Conclusion}\label{sec:conclusion}
In this study, we proposed the efficient MOBO method for identifying the PF defined by general risk measures. 
The proposed method can work with (and without) IU and has theoretical guarantees. 
In various risk measures, we proved that the algorithm can return an arbitrary-accurate 
solution with high probability in a finite number of iterations. 
Through numerical experiments, we confirmed that the proposed method outperforms existing methods. 
Moreover, from the real-world docking simulation that is our motivating example, we confirmed that by using the proposed method, the number of function evaluations required to identify the true PF has been successfully reduced to about half that of the exhaustive search. 

The proposed method has two limitations.
First, although we have given a theoretical analysis of how approximation errors in the proposed method affect the final results, we have not mentioned an estimate of the degree of approximation errors in the first place.
Thus, as a practical matter, it is difficult to estimate the final accuracy of the proposed method considering the approximation error in advance.
Second, the proposed method does not consider constraint conditions. 
In actual applications, Pareto optimization under some constraints is often considered.
We can  apply the proposed method to this setting directly by designing a HPBB for the constraint function. 
However, it is not obvious whether theoretical results derived in this study can be derived in the same way in such a case. 
The above problems are left for future work.

\section*{Acknowledgement}
This work was partially supported by JSPS KAKENHI (JP20H00601,JP23K16943,JP23K19967), JST ACT-X (JPMJAX23CD), JST CREST (JPMJCR21D3, JPMJCR22N2), JST Moonshot R\&D (JPMJMS2033-05), JST AIP Acceleration Research (JPMJCR21U2), NEDO (JPNP18002, JPNP20006) and RIKEN Center for Advanced Intelligence Project.

\bibliography{myref}

\begin{thebibliography}{}

\bibitem[Abbasi-Yadkori, 2013]{abbasi2013online}
Abbasi-Yadkori, Y. (2013).
\newblock Online learning for linearly parametrized control problems.

\bibitem[Belakaria et~al., 2020]{belakaria2020uncertainty}
Belakaria, S., Deshwal, A., Jayakodi, N.~K., and Doppa, J.~R. (2020).
\newblock Uncertainty-aware search framework for multi-objective bayesian
  optimization.
\newblock In {\em Proceedings of the AAAI Conference on Artificial
  Intelligence}, volume~34, pages 10044--10052.

\bibitem[Beland and Nair, 2017]{beland2017bayesian}
Beland, J.~J. and Nair, P.~B. (2017).
\newblock Bayesian optimization under uncertainty.
\newblock In {\em NIPS BayesOpt 2017 workshop}, volume~2.

\bibitem[Daulton et~al., 2022]{daulton2022robust}
Daulton, S., Cakmak, S., Balandat, M., Osborne, M.~A., Zhou, E., and Bakshy, E.
  (2022).
\newblock Robust multi-objective bayesian optimization under input noise.
\newblock In {\em International Conference on Machine Learning}, pages
  4831--4866. PMLR.

\bibitem[Deb and Gupta, 2005]{deb2005searching}
Deb, K. and Gupta, H. (2005).
\newblock Searching for robust pareto-optimal solutions in multi-objective
  optimization.
\newblock In {\em International conference on evolutionary multi-criterion
  optimization}, pages 150--164. Springer.

\bibitem[Deb et~al., 2002]{deb2002fast}
Deb, K., Pratap, A., Agarwal, S., and Meyarivan, T. (2002).
\newblock A fast and elitist multiobjective genetic algorithm: Nsga-ii.
\newblock {\em IEEE transactions on evolutionary computation}, 6(2):182--197.

\bibitem[Emmerich and Klinkenberg, 2008]{emmerich2008computation}
Emmerich, M. and Klinkenberg, J.-w. (2008).
\newblock The computation of the expected improvement in dominated hypervolume
  of pareto front approximations.
\newblock {\em Rapport technique, Leiden University}, 34:7--3.

\bibitem[Inatsu et~al., 2021]{inatsu2021active}
Inatsu, Y., Iwazaki, S., and Takeuchi, I. (2021).
\newblock Active learning for distributionally robust level-set estimation.
\newblock In {\em International Conference on Machine Learning}, pages
  4574--4584. PMLR.

\bibitem[Inatsu et~al., 2022]{pmlr-v162-inatsu22a}
Inatsu, Y., Takeno, S., Karasuyama, M., and Takeuchi, I. (2022).
\newblock {B}ayesian optimization for distributionally robust
  chance-constrained problem.
\newblock In Chaudhuri, K., Jegelka, S., Song, L., Szepesvari, C., Niu, G., and
  Sabato, S., editors, {\em Proceedings of the 39th International Conference on
  Machine Learning}, volume 162 of {\em Proceedings of Machine Learning
  Research}, pages 9602--9621. PMLR.

\bibitem[Iwazaki et~al., 2021a]{iwazaki2021bayesian}
Iwazaki, S., Inatsu, Y., and Takeuchi, I. (2021a).
\newblock Bayesian quadrature optimization for probability threshold robustness
  measure.
\newblock {\em Neural Computation}, 33(12):3413--3466.

\bibitem[Iwazaki et~al., 2021b]{iwazaki2020mean}
Iwazaki, S., Inatsu, Y., and Takeuchi, I. (2021b).
\newblock Mean-variance analysis in bayesian optimization under uncertainty.
\newblock In Banerjee, A. and Fukumizu, K., editors, {\em Proceedings of The
  24th International Conference on Artificial Intelligence and Statistics},
  volume 130 of {\em Proceedings of Machine Learning Research}, pages 973--981.
  PMLR.

\bibitem[Kirschner et~al., 2020]{pmlr-v108-kirschner20a}
Kirschner, J., Bogunovic, I., Jegelka, S., and Krause, A. (2020).
\newblock Distributionally robust bayesian optimization.
\newblock In Chiappa, S. and Calandra, R., editors, {\em Proceedings of the
  Twenty Third International Conference on Artificial Intelligence and
  Statistics}, volume 108 of {\em Proceedings of Machine Learning Research},
  pages 2174--2184. PMLR.

\bibitem[Kirschner and Krause, 2018]{pmlr-v75-kirschner18a}
Kirschner, J. and Krause, A. (2018).
\newblock Information directed sampling and bandits with heteroscedastic noise.
\newblock In Bubeck, S., Perchet, V., and Rigollet, P., editors, {\em
  Proceedings of the 31st Conference On Learning Theory}, volume~75 of {\em
  Proceedings of Machine Learning Research}, pages 358--384. PMLR.

\bibitem[Knowles, 2006]{knowles2006parego}
Knowles, J. (2006).
\newblock Parego: A hybrid algorithm with on-line landscape approximation for
  expensive multiobjective optimization problems.
\newblock {\em IEEE transactions on evolutionary computation}, 10(1):50--66.

\bibitem[Kusakawa et~al., 2022]{kusakawa2022bayesian}
Kusakawa, S., Takeno, S., Inatsu, Y., Kutsukake, K., Iwazaki, S., Nakano, T.,
  Ujihara, T., Karasuyama, M., and Takeuchi, I. (2022).
\newblock Bayesian optimization for cascade-type multistage processes.
\newblock {\em Neural Computation}, 34(12):2408--2431.

\bibitem[Makarova et~al., 2021]{NEURIPS2021_8f97d1d7}
Makarova, A., Usmanova, I., Bogunovic, I., and Krause, A. (2021).
\newblock Risk-averse heteroscedastic bayesian optimization.
\newblock In Ranzato, M., Beygelzimer, A., Dauphin, Y., Liang, P., and Vaughan,
  J.~W., editors, {\em Advances in Neural Information Processing Systems},
  volume~34, pages 17235--17245. Curran Associates, Inc.

\bibitem[Mo{\v{c}}kus, 1975]{movckus1975bayesian}
Mo{\v{c}}kus, J. (1975).
\newblock On bayesian methods for seeking the extremum.
\newblock In {\em Optimization Techniques IFIP Technical Conference:
  Novosibirsk, July 1--7, 1974}, pages 400--404. Springer.

\bibitem[Nguyen et~al., 2021a]{nguyen2021optimizing}
Nguyen, Q.~P., Dai, Z., Low, B. K.~H., and Jaillet, P. (2021a).
\newblock Optimizing conditional value-at-risk of black-box functions.
\newblock {\em Advances in Neural Information Processing Systems},
  34:4170--4180.

\bibitem[Nguyen et~al., 2021b]{nguyen2021value}
Nguyen, Q.~P., Dai, Z., Low, B. K.~H., and Jaillet, P. (2021b).
\newblock Value-at-risk optimization with gaussian processes.
\newblock In {\em International Conference on Machine Learning}, pages
  8063--8072. PMLR.

\bibitem[Qing et~al., 2023]{qing2023robust}
Qing, J., Couckuyt, I., and Dhaene, T. (2023).
\newblock A robust multi-objective bayesian optimization framework considering
  input uncertainty.
\newblock {\em Journal of Global Optimization}, 86(3):693--711.

\bibitem[Rahimi and Recht, 2007]{rahimi2007random}
Rahimi, A. and Recht, B. (2007).
\newblock Random features for large-scale kernel machines.
\newblock {\em Advances in neural information processing systems}, 20.

\bibitem[Rasmussen and Williams, 2005]{GPML}
Rasmussen, C.~E. and Williams, C. K.~I. (2005).
\newblock {\em Gaussian Processes for Machine Learning (Adaptive Computation
  and Machine Learning)}.
\newblock The MIT Press.

\bibitem[Rivier and Congedo, 2022]{rivier2022surrogate}
Rivier, M. and Congedo, P.~M. (2022).
\newblock Surrogate-assisted bounding-box approach applied to constrained
  multi-objective optimisation under uncertainty.
\newblock {\em Reliability Engineering \& System Safety}, 217:108039.

\bibitem[{Schr\"{o}dinger LLC}, 2021]{schrodinger2021}
{Schr\"{o}dinger LLC} (2021).
\newblock Schr\"{o}dinger release 2021-2.

\bibitem[Shahriari et~al., 2015]{shahriari2015taking}
Shahriari, B., Swersky, K., Wang, Z., Adams, R.~P., and De~Freitas, N. (2015).
\newblock Taking the human out of the loop: A review of bayesian optimization.
\newblock {\em Proceedings of the IEEE}, 104(1):148--175.

\bibitem[Srinivas et~al., 2010]{GPUCB}
Srinivas, N., Krause, A., Kakade, S.~M., and Seeger, M.~W. (2010).
\newblock Gaussian process optimization in the bandit setting: No regret and
  experimental design.
\newblock In F{\"{u}}rnkranz, J. and Joachims, T., editors, {\em Proceedings of
  the 27th International Conference on Machine Learning (ICML-10), June 21-24,
  2010, Haifa, Israel}, pages 1015--1022. Omnipress.

\bibitem[Suzuki et~al., 2020]{suzuki2020multi}
Suzuki, S., Takeno, S., Tamura, T., Shitara, K., and Karasuyama, M. (2020).
\newblock Multi-objective bayesian optimization using pareto-frontier entropy.
\newblock In {\em International Conference on Machine Learning}, pages
  9279--9288. PMLR.

\bibitem[Svenson and Santner, 2010]{svenson2010multiobjective}
Svenson, J.~D. and Santner, T.~J. (2010).
\newblock Multiobjective optimization of expensive black-box functions via
  expected maximin improvement.
\newblock {\em The Ohio State University, Columbus, Ohio}, 32.

\bibitem[Takeno et~al., 2023]{pmlr-v202-takeno23a}
Takeno, S., Inatsu, Y., and Karasuyama, M. (2023).
\newblock Randomized {G}aussian process upper confidence bound with tighter
  {B}ayesian regret bounds.
\newblock In Krause, A., Brunskill, E., Cho, K., Engelhardt, B., Sabato, S.,
  and Scarlett, J., editors, {\em Proceedings of the 40th International
  Conference on Machine Learning}, volume 202 of {\em Proceedings of Machine
  Learning Research}, pages 33490--33515. PMLR.

\bibitem[Vershynin, 2018]{vershynin2018high}
Vershynin, R. (2018).
\newblock {\em High-dimensional probability: An introduction with applications
  in data science}, volume~47.
\newblock Cambridge university press.

\bibitem[Wang and Jegelka, 2017]{wang2017max}
Wang, Z. and Jegelka, S. (2017).
\newblock Max-value entropy search for efficient bayesian optimization.
\newblock In {\em International Conference on Machine Learning}, pages
  3627--3635. PMLR.

\bibitem[Zhou et~al., 2018]{zhou2018multi}
Zhou, Q., Jiang, P., Huang, X., Zhang, F., and Zhou, T. (2018).
\newblock A multi-objective robust optimization approach based on gaussian
  process model.
\newblock {\em Structural and Multidisciplinary Optimization}, 57:213--233.

\bibitem[Zuluaga et~al., 2016]{zuluaga2016varepsilon}
Zuluaga, M., Krause, A., and P{\"u}schel, M. (2016).
\newblock $\varepsilon$-pal: an active learning approach to the multi-objective
  optimization problem.
\newblock {\em The Journal of Machine Learning Research}, 17(1):3619--3650.

\end{thebibliography}
\bibliographystyle{apalike}

\newpage
\section*{Appendix}

\setcounter{section}{0}
\renewcommand{\thesection}{\Alph{section}}
\renewcommand{\thesubsection}{\thesection.\arabic{subsection}}

\section{Extension}\label{app:generalized}
In this section, we extend the proposed method. 
We consider the following four extensions:
\begin{itemize}
\item The number of black-box functions and the number of risk measures are different. 
\item The true noise distribution follows some heteroscedastic sub-Gaussian distribution. 
\item The distribution of ${\bm w} $ depends on the design variable ${\bm x} $. 
\item We consider the uncontrollable setting, that is,   ${\bm w} $ cannot be controlled even during optimization.  
\end{itemize}

\subsection{Extension of Problem Setup}
\paragraph{Preliminary} 
Let $f^{(m)}: \mathcal{X} \times \Omega \to \mathbb{R}$ be an expensive-to-evaluate black-box function, where $ m \in [M_f]$ and $M_f \geq 1$. 
Assume that the set of design variables $\mathcal{X} $ and set of environmental variables $\Omega $ are compact and convex. 
For each design variable ${\bm x} \in \mathcal{X}$, the environmental variable ${\bm w} $ follows some probability distribution $P_{ {\bm w} } ({\bm x} )$, which depends on ${\bm x}$, and takes values in a compact and convex subset $\Omega_{ {\bm x} }  \subset \Omega $.
For each iteration $t$, input $({\bm x}_t,{\bm w}_t ) \in \mathcal{X} \times \Omega $, and $m \in [M_f]$, the value of the black-box function $f^{(m) }$ is observed with noise as $y^{(m)} _t= f^{(m) } ({\bm x}_t,{\bm w}_t ) + \eta^{(m)} ({\bm x}_t,{\bm w}_t) $, where 
$\eta ^{(m)} ({\bm x}_t,{\bm w}_t) $ is zero-mean noise independent across different iteration $t$, $m \in [M_f]$ and ${\bm w}_t$. 
In this section, we assume that $\eta^{(m)} ({\bm x}_t,{\bm w}_t) $ is a sub-Gaussian heteroscedastic noise that depends on $({\bm x},{\bm w} ,m)$. 
\begin{definition}
Let  $\eta$ be a zero-mean real-valued random variable. 
 Then, $\eta$  is  $\tau$-sub-Gaussian if there exists  a positive constant $\tau^2 $ such that 
$$
^\forall a \in \mathbb{R}, \quad \mathbb{E} [ e^{ a \eta} ] \leq \exp \left (    
\frac{a^2 \tau^2  }{2}
\right ).
$$
\end{definition}
Commonly used distributions such as Gaussian, Bernoulli and uniform are sub-Gaussian \citep{vershynin2018high}. 
We assume that the random variables $\{ {\bm w}_t ,  \eta^{(m)} ({\bm x}_t,{\bm w}_t) \}_{ t \geq 1, m \in [M_f] } $ are mutually independent. 
For ${\bm w} $, we consider the both  simulator and uncontrollable settings.  
 Let $\rho^{(m,l)}( f^{(m)} ({\bm x} ,{\bm w} ) ) \equiv F^{(m,l)} ({\bm x}  ) $ be a risk measure, where $ l \in \{1,\ldots , L_m \} $ and $L_1+\cdots + L_{M_f} \equiv L \geq 2 $. 
The purpose of this study is to efficiently identify the PF  defined based on  $F^{(m,l)} ({\bm x} ) $. 
For any ${\bm x} \in \mathcal{X} $ and $E \subset \mathcal{X} $, let 
$$
{\bm F} ({\bm x} ) =  (F^{(1,1)} ({\bm x} ) , \ldots , F^{(1,L_1)} ({\bm x} ),\ldots ,F^{(M_f,1)} ({\bm x} ) , \ldots , F^{(M_f,L_{M_f})} ({\bm x} )  )
$$
 and ${\bm F} (E) =  \{  {\bm F}  ({\bm x} )  \mid {\bm x}  \in E \}$. 
Then, for any $ B \subset \mathbb{R}^L $,  the dominated region ${\rm Dom} (B)$ and PF  ${\rm Par} (B) $ of $B$ are  defined as 
\begin{align*}
{\rm Dom} (B) = \{  {\bm s} \in \mathbb{R}^L \mid ^\exists {\bm s}^\prime \in B \ {\rm s.t.} \ {\bm s} \leq {\bm s}^\prime \} , \ 
{\rm Par} (B) =  \partial ({\rm Dom} (B) ) .
\end{align*}
Let  $Z^\ast$ be our target PF. 
 Then,   $Z^\ast$ can be expressed as 
$$
Z^\ast =   {\rm Par }   ({\bm F}   (\mathcal{X})).
$$

\paragraph{Regularity Assumption}   
We introduce a regularity assumption for $f^{(m)} $. 
For each $ m \in [M_f]$, 
let $k^{(m)} : ( \mathcal{X} \times \Omega )   \times  ( \mathcal{X} \times \Omega )   \to \mathbb{R}$ be a positive-definite kernel, where $ k^{(m)} (  ({\bm x},{\bm w} ) ,  ({\bm x},{\bm w} ) )  \leq 1 $ for any   
 $({\bm x} ,{\bm w} )  \in \mathcal{X} \times \Omega $. 
Also let  $ \mathcal{H}   (k^{(m)}) $ be a reproducing kernel Hilbert space (RKHS) corresponding  to  $k^{(m)} $.  
We assume that $f^{(m)} $ is the element of $ \mathcal{H}   (k^{(m)}) $ and has the bounded Hilbert norm $\|  f^{(m)}  \|  _{ \mathcal{H}   (k^{(m)})  } \leq B_m    < \infty$. 
Moreover, we assume that the noise $\eta^{(m)} ({\bm x},{\bm w} ) $ is $\tau ({\bm x},{\bm w} ,m)$-sub-Gaussian, where  $\tau ({\bm x},{\bm w} ,m) \equiv \tau_{{\bm x},{\bm w},m}  $ satisfies $ \tau_{{\bm x},{\bm w},m}  \in [\underline{\tau},\bar{\tau}] $ for some $\bar{\tau} \geq \underline{\tau} >0$. 

\paragraph{Gaussian Process Model} 
We use a GP   model for the black-box function $f^{(m)}$.
Let $\lambda _1 , \ldots , \lambda_{M_f} $ be positive numbers. 
We assume the GP $\mathcal{G}\mathcal{P}(0, \tilde{k}  (  ( {\bm x},{\bm w}  ),  ( {\bm x}^\prime,{\bm w}^\prime  )  )  )$ as the prior of $f^{(m) } $, where $ \tilde{k}  (  ( {\bm x},{\bm w}  ),  ( {\bm x}^\prime,{\bm w}^\prime  )  )      $ is given by 
$$
 \tilde{k}  (  ( {\bm x},{\bm w}  ),  ( {\bm x}^\prime,{\bm w}^\prime  )  )  = \frac{1}{ \lambda_m}  k^{(m)} (  ( {\bm x},{\bm w}  ),  ( {\bm x}^\prime,{\bm w}^\prime  )  )   .   
$$ 
Furthermore, we consider the zero-mean normal distribution with variance $\tau^2_{ {\bm x},{\bm w},m  }$,  as the error distribution in the GP model. 
For $m \in [M_f]$, given a dataset $\{ ({\bm x}_i,{\bm w}_i,y^{(m)}_i )\}_{i=1}^t$, where $t$ is the number of queried instances, the posterior  of $f^{(m)}$ is a GP. 
Then,  its posterior mean $ \tilde{\mu}^{(m)}_t ({\bm x},{\bm w}) $ and posterior variance $ \tilde{\sigma}^{(m) 2}_t ({\bm x},{\bm w}) $ can be calculated as follows:
\begin{align*}
&\tilde{\mu}^{(m)}_t ({\bm x},{\bm w}) = \tilde{\bm k}^{(m)} _t ({\bm x},{\bm w} )^\top  (\tilde{{\bm K}}^{(m)}_t + {\bm \Sigma}^{(m)}_t )^{-1} {\bm y}^{(m)}_t , \\  
&\tilde{\sigma}^{(m) 2}_t ({\bm x},{\bm w} )= \tilde{k}^{(m)}(  ({\bm x},{\bm w}),  ({\bm x},{\bm w})) 
   -\tilde{\bm k}^{(m)}_t ({\bm x},{\bm w})^\top 
(\tilde{\bm K}^{(m)}_t +  {\bm \Sigma}^{(m)}_t )^{-1} \tilde{\bm k}^{(m)} _t ({\bm x},{\bm w}),
\end{align*} 
where $\tilde{\bm k}^{(m)}_t ({\bm x},{\bm w} ) $ is the $t$-dimensional vector, whose $j$-th element is $ \tilde{k}^{(m)}(({\bm x},{\bm w} ) ,({\bm x}_j,{\bm w}_j)) $, ${\bm y}^{(m)}_t = (y^{(m)}_1,\ldots , y^{(m)}_t )^\top $, $ {\bm \Sigma}^{(m)}_t $ is the $t \times t$ diagonal matrix whose $(j,j)$-th element is $\tau^2_{{\bm x}_t,{\bm w}_t,m   }$, $\tilde{\bm K}^{(m)}_t $ is the 
$t \times t$ matrix whose 
$(j,k)$-th element is $ \tilde{k}^{(m)}(({\bm x}_j,{\bm w}_j ) ,({\bm x}_k,{\bm w}_k))$, with a superscript $\top$ indicating the transpose of vectors or matrices.

\subsection{Extension of Proposed Method}
\paragraph{Credible Interval and Bounding Box} 
For each input $({\bm x} ,{\bm w} ) \in \mathcal{X} \times \Omega $ and $t \geq 1$, the CI of $f^{(m)} ({\bm x},{\bm w} )$ is denoted by $\tilde{Q}^{(f^{(m)})}_{t-1} ({\bm {x}},{\bm w}) =[ \tilde{l}^{(f^{(m)})}_{t-1} ({\bm {x}},{\bm w}), \tilde{u}^{(f^{(m)})}_{t-1} ({\bm {x}},{\bm w})]$, where
     $\tilde{l}^{(f^{(m)})}_{t-1} ({\bm {x}},{\bm w})$ and  $\tilde{u}^{(f^{(m)})}_{t-1} ({\bm {x}},{\bm w})$ are given by 
\begin{align*}
\tilde{l}^{(f^{(m)})}_{t-1} ({\bm {x}},{\bm w})& = \tilde{\mu}^{(m)}_{t-1} ({\bm {x}},{\bm w}) - \tilde{\beta}^{1/2}_{m,t} \tilde{\sigma}^{(m)}_{t-1} ({\bm {x}},{\bm w}) , \\
 \tilde{u}^{(f^{(m)})}_{t-1} ({\bm {x}},{\bm w}) &= \tilde{\mu}^{(m)}_{t-1} ({\bm {x}},{\bm w}) + \tilde{\beta}^{1/2}_{m,t} \tilde{\sigma}^{(m)}_{t-1} ({\bm {x}},{\bm w}). 
\end{align*}
For ${\bm x} \in \mathcal{X}$, $t \geq 1$ and $m \in [M_f]$, we define the set of functions $\tilde{G}^{(m)} _{t-1} ({\bm x} ) $ as 
%
$$
\tilde{G}^{(m)} _{t-1} ({\bm x} ) = \{    g({\bm x},{\bm w}  )  \mid ^\forall {\bm w} \in \Omega , g({\bm x},{\bm w}  ) \in \tilde{Q}^{(f^{(m)})}_{t-1} ({\bm {x}},{\bm w}) \}. 
$$
Let $\tilde{Q}^{(F^{(m,l)})}_{t-1} ({\bm {x}}) = [{\rm lcb }^{(m,l)}_{t-1}  ({\bm x}  )  , {\rm ucb }^{(m,l)}_{t-1}  ({\bm x}  ) ]$ be a CI of  $F^{(m,l)}  ({\bm x}  ) $. 
Also let $$
\tilde{B}_{t-1} ({\bm x} ) = \prod_{m=1}^{M_f}  \prod_{l=1}^{L_m} \tilde{Q}^{(F^{(m,l)})}_{t-1} ({\bm {x}}) 
$$
 be a bounding box of 
${\bm F} ({\bm x} )$. 
Then, when $\tilde{Q}^{(f^{(m)})}_{t-1} ({\bm {x}},{\bm w}) $ is HPCI, a sufficient condition for $Q^{(F^{(m,l)})}_{t-1} ({\bm {x}})$ to also be HPCI is given as follows:
\begin{equation}
\begin{split}
^\forall  g({\bm x}  ,{\bm w}  )   \in \tilde{G}^{(m)} _{t-1} ({\bm x} ), \ 
{\rm lcb }^{(m,l)}_{t-1}  ({\bm x}  )  \leq    \rho^{(m,l)}   (  g({\bm x}  ,{\bm w}  )   ) 
\leq {\rm ucb }^{(m,l)}_{t-1}  ({\bm x}  ).  
\end{split} \label{eq:HPCI condition1_app}
\end{equation}
If \eqref{eq:HPCI condition1_app}  holds, then   $\tilde{B}_{t-1} ({\bm x} ) $ is also a HPBB. 
Next, we provide computation methods for ${\rm lcb }^{(m,l)}_{t-1}  ({\bm x}  ) $ and ${\rm ucb }^{(m,l)}_{t-1}  ({\bm x}  ) $. 
First, we provide a generalized  method for 
  ${\rm lcb }^{(m,l)}_{t-1}  ({\bm x}  ) $ and ${\rm ucb }^{(m,l)}_{t-1}  ({\bm x}  ) $ to satisfy \eqref{eq:HPCI condition1_app}. 
The  ${\rm lcb }^{(m,l)}_{t-1}  ({\bm x}  ) $ and ${\rm ucb }^{(m,l)}_{t-1}  ({\bm x}  ) $ by the generalized  method are calculated with 
\begin{align*}
{\rm lcb }^{(m,l)}_{t-1}  ({\bm x}  )  &= \inf _{  g({\bm x}  ,{\bm w}  )   \in \tilde{G}^{(m)} _{t-1} ({\bm x} ) }     \rho^{(m,l)}   (  g({\bm x}  ,{\bm w}  )   ) , \\
{\rm ucb }^{(m)}_{t-1}  ({\bm x}  )  &= \sup _{  g({\bm x}  ,{\bm w}  )   \in \tilde{G}^{(m)} _{t-1} ({\bm x} ) }     \rho^{(m,l)}   (  g({\bm x}  ,{\bm w}  )   ).
\end{align*}
%
The condition \eqref{eq:HPCI condition1_app} holds by using the generalized method, the inf and sup calculations in the generalized method are not always easy. 
%
%
Therefore,  we give additional two computation methods for 
${\rm lcb }^{(m,l)}_{t-1}  ({\bm x}  ) $ and ${\rm ucb }^{(m,l)}_{t-1}  ({\bm x}  ) $, the decomposition method and  sampling method. 
Let $\rho (\cdot)$ be a risk measure. 
In many cases, $\rho (\cdot)$ can be decomposed as $\rho (\cdot)  = \tilde{\rho} \circ h (\cdot)$, where $ \tilde{\rho} (\cdot)$ and $h(\cdot)$ are respectively monotonic and tractable functions. 
The basic idea of the decomposition method is to compute the infimum and supremum of $h (g({\bm x}  ,{\bm w}  ))$ on $\tilde{G}^{(m)} _{t-1} ({\bm x} ) $, and then to compute  ${\rm lcb }^{(m,l)}_{t-1}  ({\bm x}  ) $ and ${\rm ucb }^{(m,l)}_{t-1}  ({\bm x}  ) $ by taking $\tilde{\rho} (\cdot)$ to these. 
Calculated values for several risk measures are given  in Table \ref{tab:decomposition_result}, where  we omit the notation \textasciitilde and $l$ in the table for simplicity. 
Note that by combining several risk measures such as the Bayes risk, standard deviation, monotonic Lipschitz map and weighted sum, we can obtain the result for mixed risk measures such as $0.7 F^{(m_1) } ({\bm x} ) - 0.3 F^{(m_2) } ({\bm x} )$, where 
$ F^{(m_1) } ({\bm x} )$ and $ F^{(m_2) } ({\bm x} )$ are the Bayes risk and standard deviation, respectively. 
In the sampling method, we generate $S$ sample paths $f^{(m)} _1 ({\bm x} ,{\bm w} ) ,\ldots ,
f^{(m)} _S ({\bm x} ,{\bm w} ) $ of $f^{(m)} ({\bm x} ,{\bm w} )$ independently from the GP posterior  and compute 
\begin{align*}
{\rm lcb }^{(m,l)}_{t-1}  ({\bm x}  ) 
 &= \min_{  j \in [S], f^{(m)} _j ({\bm x} ,{\bm w} ) \in \tilde{G}^{(m)} _{t-1} ({\bm x} )    }      \rho ^{(m,l)}  (f^{(m)} _j ({\bm x} ,{\bm w} ))  , \\ 
{\rm ucb }^{(m,l)}_{t-1}  ({\bm x}  ) 
 &= \max_{  j \in [S], f^{(m)} _j ({\bm x} ,{\bm w} ) \in \tilde{G}^{(m)} _{t-1} ({\bm x} )    }      \rho ^{(m,l)}  (f^{(m)} _j ({\bm x} ,{\bm w} ))  .
\end{align*}

\paragraph{Pareto Front Estimation} 
For any input ${\bm x} \in \mathcal{X}$ and  subset $E \subset \mathcal{X}$, 
we define $ \text{\bf LCB}_{t-1} ({\bm x} )$, 
$ \text{\bf UCB}_{t-1} ({\bm x} )$ and $ \text{\bf LCB}_{t-1} (E )$ as 
\begin{align*}
\text{\bf LCB}_{t-1} ({\bm x} )&= ( {\rm lcb}^{(1,1)}_{t-1}  ({\bm x} )  , \ldots ,   {\rm lcb}^{(M_f,L_{M_f})}_{t-1}  ({\bm x} ) ), 
 \text{\bf UCB}_{t-1} ({\bm x} )=( {\rm ucb}^{(1,1)}_{t-1}  ({\bm x} )  , \ldots ,   {\rm ucb}^{(M_f,L_{M_f})}_{t-1}  ({\bm x} ) ), \\
\text{\bf LCB}_{t-1} (E)&=  \{  \text{\bf LCB}_{t-1} ({\bm x} ) \mid {\bm x} \in E \} . 
\end{align*}
The estimated Pareto solution set $\hat{\Pi}_{t-1} \subset \mathcal{X}$ for the design variables is then defined as follows: 
$$
\hat{\Pi}_{t-1} = \{ {\bm x} \in \mathcal{X} \mid  \text{\bf LCB}_{t-1} ({\bm x} ) \in \text{Par} (\text{\bf LCB}_{t-1} (\mathcal{X})) \}.
$$
Here, in order to actually compute $\hat{\Pi}_{t-1} $, we need to compute the PF defined by $\text{\bf LCB}_{t-1} ({\bm x} )$. 
However, if $\mathcal{X}$ is an infinite set, then $\hat{\Pi}_{t-1} $ may also be an infinite set. 
In this case, since the exact calculation of $\hat{\Pi}_{t-1} $ is difficult,  it is necessary to make a finite approximation using an approximation solver such as NSGA-II \citep{deb2002fast}.  

\paragraph{Acquisition Function} 
We propose an AF for determining the next point to be evaluated. 
We define  AF $a^{(\mathcal{X})}_t ({\bm x})$ for ${\bm x} \in \mathcal{X}$ as 
$$
a^{(\mathcal{X})}_t ({\bm x} ) =  \text{dist}   (   \text{\bf UCB}_t ({\bm x} ), \text{Dom}  (\text{\bf LCB}_t (\hat{\Pi}_t )) )
$$
%
Then, the next design variable, ${\bm x}_{t+1}$, to be evaluated is selected by 
$$
{\bm x}_{t+1} = \argmax_{ {\bm x} \in \mathcal{X} }   a^{(\mathcal{X})}_t ({\bm x} ). 
$$
%
Hence, the value of $   a^{(\mathcal{X})}_t (  {\bm x}_{t+1} )   $ is equal to the following maximin distance:
$$
a^{(\mathcal{X})}_t (  {\bm x}_{t+1} )  
=\max_{ {\bm x} \in \mathcal{X} }   \min_{ {\bm b}  \in  \text{Dom}  (\text{\bf LCB}_t (\hat{\Pi}_t )) } d_\infty ( \text{\bf UCB}_t ({\bm x} ), {\bm b}).
$$  
The value of $a^{(\mathcal{X})}_t ({\bm x})$ can be computed analytically using the following lemma when $ \hat{\Pi}_t $ is finite:
\begin{lemma}\label{lem:AF_cal_app}
Let $ \text{\bf UCB}_t ({\bm x} ) = (u_1,\ldots , u_L)$ and $\text{\bf LCB}_t (\hat{\Pi}_t ) = \{ (l^{(i)}_1,\ldots , l^{(i)}_L ) \mid 1 \leq i \leq k \}$. 
Then, $a^{(\mathcal{X})}_t ({\bm x})$ can be computed by 
\begin{align*}
a^{(\mathcal{X})}_t ({\bm x} ) = \max \{ \tilde{a}_t ({\bm x} ) , 0 \} , \ 
\tilde{a}_t ({\bm x} ) = \min_{1 \leq i \leq k } \max \{ u_1 - l^{(i)}_1,\ldots ,  u_L - l^{(i)}_L \}. 
\end{align*}
\end{lemma}
Next, we consider the simulator setting. 
In this case, we have to select the environment variable ${\bm w}_{t+1} $. 
Based on the fact that many risk measures including Bayes risk satisfy 
\begin{equation}
\|  \text{\bf UCB}_t ({\bm x}_{t+1} ) -  \text{\bf LCB}_t ({\bm x}_{t+1} ) \|_\infty 
\leq q  \left (  \max_{{\bm w} \in \Omega_{{\bm x}_{t+1}}} \sum_{m=1}^{M_f}   2 \beta^{1/2}_{m,t+1} \tilde{\sigma}^{(m)}_t ({\bm x}_{t+1},{\bm w}  ) \right ) ,\label{eq:w_condition_app}
\end{equation}
where $q(\cdot ): [0,\infty ) \to [0, \infty )$ is a strictly increasing function defined by risk measures and satisfies $q(0) =0$,   we choose ${\bm w}_{t+1}$ as follows:
\begin{align*}
 {\bm w}_{t+1} = \argmax _{ {\bm w} \in \Omega_{{\bm x}_{t+1} } }   a^{(\Omega_{{\bm x}_{t+1}})}_t ({\bm w} ), \ 
   a^{(\Omega_{{\bm x}_{t+1}})}_t ({\bm w} ) =\sum_{m=1}^{M_f}   2 \beta^{1/2}_{m,t+1} \tilde{\sigma}^{(m)}_t ({\bm x}_{t+1},{\bm w}  ). 
\end{align*}

On the other hand, in the uncontrollable setting, since we cannot control ${\bm w}$, ${\bm w} _{t+1} $ is defined as the sample from $\Omega_{  {\bm x} }  $.

\subsection{Stopping Condition}
We describe the stopping conditions of the proposed algorithm. 
Let $\epsilon >0$ be a predetermined desired accuracy parameter.
Then the algorithm is terminated if $ a^{(\mathcal{X})}_t ({\bm x}_{t+1} ) \leq \epsilon $ is satisfied. 
The pseudocode of the proposed algorithm is described in Algorithm \ref{alg:2}.

\begin{algorithm}[t]
    \caption{Bounding box-based MOBO of general risk measures under extended problem setup}
    \label{alg:2}
    \begin{algorithmic}
        \REQUIRE GP priors $\mathcal{GP}(0,\ k^{(m)})$, tradeoff parameters $\{\beta_{m,t}\}_{t \geq 0}$, accuracy parameter $\epsilon >0$, $m \in [M_f]$, $M_f \geq 1$, $L_m \geq 1 $, $L \geq 2$ 
        \FOR { $ t= 0,1,2,\ldots $}
            \STATE Compute $\tilde{Q}^{(f^{(m)})}_{t} ({\bm x},{\bm w} )$  for all $m \in [M_f]$ and $(\bm{x}, {\bm w} )  \in \mathcal{X} \times \Omega$
            \STATE Compute $\tilde{Q}^{(F^{(m,l)})}_{t} ({\bm x} )$  for all $m \in [M_f]$, $l \in [L_m]$ and $\bm{x}  \in \mathcal{X} $ by the generalized, decomposition or sampling method
		\STATE Compute $\tilde{B}_t ({\bm x} ) = \prod_{m=1}^{M_f} \prod_{l=1}^{L_m} Q^{(F^{(m.l)})}_{t} ({\bm x} )  $ for each ${\bm x} \in \mathcal{X}$ 
		\STATE Estimate $\hat{\Pi}_t $ by $\tilde{B}_t ({\bm x} )$
            \STATE Select the next evaluation point $\bm{x}_{t+1}$ by $a^{(\mathcal{X})}_{t}  ({\bm x} ) $   
		\IF {$  a^{(\mathcal{X})}_{t} ({\bm x}_{t+1})  \leq  \epsilon$}
		\STATE break
		\ENDIF
		\IF {simulator setting}
		\STATE Select the next evaluation point $\bm{w}_{t+1}$ by $a^{(\Omega_{{\bm x}_{t+1}})}_{t}  ({\bm w} ) $
		\ELSE[uncontrollable setting]
		\STATE ${\bm w}_{t+1} $ is generated from $P_{{\bm w} }  ({\bm x}_{t+1}  ) $
		\ENDIF
            \STATE Observe $y^{(m)}_{t+1} = f^{(m)}(\bm{x}_{t+1}, \bm{w}_{t+1}) + \eta^{(m)} ({\bm x}_{t+1},{\bm w}_{t+1})$  at the point $(\bm{x}_{t+1}, \bm{w}_{t+1})$ for all $m \in [M_f]$
            \STATE Update GPs by adding observed points
        \ENDFOR
        \ENSURE Return $\hat{\Pi}_{t}$ as the estimated set of design variables
    \end{algorithmic}
\end{algorithm}

\subsection{Theoretical Analysis} 
In this subsection, we give the theorems for the accuracy and termination of the proposed algorithm. 
First, we quantify the goodness of the estimated $\hat{\Pi}_t$. 
If  $\hat{\Pi}_t$ is a good estimate, the following two indicators defined by $\hat{\Pi}_t$ should be small:
\begin{align*}
I^{(i)}_t &=  \max_{ {\bm y}  \in  Z^\ast   } {\rm dist}  ({\bm y} ,  {\rm Par} ({\bm F} (\hat{\Pi}_t)) ), \\ 
I^{(ii)}_t &=  \max_{ {\bm y}  \in  {\bm F} (\hat{\Pi}_t)     } {\rm dist}  ({\bm y} , Z^\ast ).  
\end{align*}
Using these, we define the inference discrepancy $I_t = \max \{ I^{(i)}_t, I^{(ii)}_t \}$ for $\hat{\Pi}_t$ as  the goodness measure. 
Next, in order to show the theoretical validity of the proposed algorithm, we introduce the maximum information gain $\tilde{\kappa}^{(m)}_t $. 
The maximum information gain $\tilde{\kappa}^{(m)} _T $ under the heteroscedastic sub-Gaussian setting can be expressed as follows \citep{NEURIPS2021_8f97d1d7}:
\begin{align*}
\tilde{\kappa} ^{(m)}_T =  \max_{ ({\bm x}_1,{\bm w}_1), \ldots,  ({\bm x}_T,{\bm w}_T)   }         \frac{1}{2}   \sum_{t=1}^T   \log \left (
1+ \frac{ \tilde{\sigma}^{(m)2}_{t-1} ({\bm x}_t,{\bm w}_t)   }{  \tau^2_{{\bm x}_t,{\bm w}_t,m}  }
\right ).
\end{align*}
The order of  $\tilde{\kappa}^{(m)} _T $ with respect to widely used kernels such as linear and squared exponential kernels is 
derived by \cite{NEURIPS2021_8f97d1d7}.
Then, the following theorem holds: 
\begin{lemma}[Lemma 7 in \cite{pmlr-v75-kirschner18a}]\label{lem:HPCI_app}
Suppose that the regularity assumption holds. 
Let $\delta \in (0,1 )$, $\lambda_1 ,\ldots , \lambda_{M_f} >0$ and define 
$$
\tilde{\beta}^{1/2}_{m,t} = B_m \sqrt{\lambda_m}  +  \sqrt{
2 \log  \left (  
\frac{     {\rm det}  (  \lambda_m {\bm \Sigma}^{(m)}_t +\tilde{ {\bm K} }^{(m)}_t)^{1/2}        }{ M^{-1}_f  \delta  {\rm det} (\lambda_m {\bm \Sigma}^{(m)}_t)^{1/2}  }
\right )
}.
$$
Then, with probability at least $1-\delta$, the following inequality holds for any $t \geq 1 $, $m \in [M_f]$ and $({\bm x},{\bm w} ) \in \mathcal{X} \times \Omega $: 
$$
|f^{(m)} ({\bm x},{\bm w} )  - \tilde{\mu}^{(m)}_{t-1}   ({\bm x},{\bm w} )  | \leq \tilde{\beta}^{1/2}_{m,t} \tilde{\sigma}^{(m)}_{t-1}  ({\bm x},{\bm w} ).
$$
\end{lemma}
Note that from the definition of the maximum information gain, when $\lambda_m =1$, $M_f =M \geq2 $, $L_m =1$ and $\tau^2_{{\bm x},{\bm w},m} = \varsigma^2_m $ and the true noise distribution is Gaussian, the inequality $\tilde{\beta}^{1/2}_{m,t}  \leq \beta^{1/2}_{m,t}$ holds, where $\beta^{1/2}_{m,t} $ is given by Lemma \ref{lem:HPCI}. 
Using this, we give the theorems for the accuracy, termination, $q(a)$ and approximation errors under both the simulator and uncontrollable settings. 
\begin{theorem}[Simulator and uncontrollable settings]\label{thm:inference discrepancy_app}
Suppose that the assumption of Lemma \ref{lem:HPCI_app} and the inequality \eqref{eq:HPCI condition1_app}  hold.
Let $t \geq 0$, $m \in [M_f]$, $\delta \in (0,1)$, and let $\tilde{\beta}^{1/2}_{m,t+1} $ be defined as in Lemma \ref{lem:HPCI_app}. 
In addition, let $\epsilon >0$ be a  predetermined desired accuracy parameter. 
Then, with probability at least $1-\delta$, the inequality $I_{t} \leq  a^{(\mathcal{X})}_{t} ({\bm x}_{t+1} ) $ holds for any $t \geq 0$ and ${\bm x}_{t+1}$. 
Therefore, if  the stopping condition 
  satisfies at $T$ iterations, the inference discrepancy $I_{T} $ satisfies $I_{T} \leq \epsilon $ with probability at least $1-\delta$.  
\end{theorem}

\begin{theorem}[Simulator setting]\label{thm:termination_app}
Suppose that the assumption in Theorem \ref{thm:inference discrepancy_app} holds. 
Let $q: [0,\infty) \to [0,\infty) $ be a strictly increasing function satisfying $q(0)=0$ and \eqref{eq:w_condition_app}. 
Also let 
$$
s_t =  \sqrt{\frac{\sum_{m=1} ^{M_f}    \tilde{C}_m \tilde{\beta}_{m,t+1} \tilde{\kappa}^{(m)}_{t+1} }{   t+1}}, 
$$
 where $\tilde{C}_m = \frac{8M_f}{  \log (1+ \lambda^{-1}_m \underline{\tau}^{-2}) }$. 
Then, the inequality $  a^{(\mathcal{X})}_{\hat{t}} ({\bm x}_{{\hat{t}}+1} )  \leq q(s_t) $ holds for any $t \geq 0$ and some $ \hat{t} \leq t$. 
Therefore, Algorithm \ref{alg:2}
 terminates after at most $T$ iterations, where $T$ is the smallest positive integer satisfying $q(s_T) \leq \epsilon$. 
\end{theorem}

\begin{theorem}[Simulator and uncontrollable settings]\label{thm:q_app}
Suppose that the assumption in Theorem \ref{thm:inference discrepancy_app} holds. 
Also assume that there exist strictly increasing functions $q^{(m,l)}: [0,\infty ) \to [0,\infty )$ 
satisfying $q^{(m,l)} (0) =0 $ and 
$$
| {\rm ucb}^{(m,l)}_{t} ({\bm x}_{t+1} ) -  {\rm lcb}^{(m,l)}_{t} ({\bm x}_{t+1} )  | 
 \leq q^{(m,l)}   ( \tilde{s}_t ) 
$$
for any $t \geq 0$, $m \in [M_f]$, $l \in [L_m]$ and $ {\bm x}_{t+1} \in \mathcal{X}$, where 
$$
\tilde{s}_t =  
\max_{ {\bm w} \in \Omega_{{\bm x}_{t+1}}} 2 \tilde{\beta}^{1/2}_{m,t+1} \tilde{\sigma}^{(m)}_{t} ({\bm x}_{t+1} , {\bm w}  ).
$$
Then,  $q (a) \equiv \max_{m \in [M_f], l \in [L_m] }  q^{(m,l)} (a) $ is the strictly increasing function and satisfies $q(0) =0$ and \eqref{eq:w_condition_app}. 
\end{theorem}
Specific forms of $q^{(m,l)} (a )$ for commonly used risk measures are described in Table \ref{tab:q}. 
For simplicity, we omitted $l$ in the table. 
From Table \ref{tab:q}, the probabilistic threshold measure does not satisfy the inequality in Theorem \ref{thm:q_app}. 
For example, if $f^{(m)} ({\bm x},{\bm w} ) = \theta$, then with high probability ${\rm ucb}^{(m)}_t ({\bm x}_{t+1} ) $ and    ${\rm lcb}^{(m)}_t ({\bm x}_{t+1} ) $ are respectively close to one and zero even when $\tilde{s}_t $ is close to zero.
  \cite{iwazaki2021bayesian,inatsu2021active} proposed BO methods for the (distributionally robust) probabilistic threshold measure and 
 confronted the same problem. 
They solved this problem by assuming the condition that the probability of a black-box function accumulating in the neighborhood of the threshold is small, and derived   ${\rm ucb}^{(m)}_t ({\bm x}_{t+1} ) -{\rm lcb}^{(m)}_t ({\bm x}_{t+1} )  \leq \tilde{q}(\tilde{s}_t) + \xi$, where $\tilde{q} (a) =0$ if $a \leq c$ and otherwise $\tilde{q} (a) =1$, and $c$ is some positive constant. 

Next, we consider the approximation error setting. 
Let $\epsilon_{{\rm lcb}}, \epsilon_{{\rm ucb}}, \epsilon_{{\rm PF}}, \epsilon_{\mathcal{X}}, \epsilon_{\Omega}$ be non-negative error parameters that represent the errors in these approximations, respectively. 
We consider the case that the following four error inequalities hold for any 
$t \geq 0$, $m \in [M_f] $, $ l \in [L_m]$, ${\bm x}, {\bm x}_{t+1} \in \mathcal{X}$,  $ {\bm w}_{t+1} \in \Omega_{{\bm x}_{t+1}}$ and $g({\bm x}  ,{\bm w}  )   \in \tilde{G}^{(m)} _{t} ({\bm x} )$: 
\begin{align}
{\rm lcb }^{(m,l)}_{t}  ({\bm x}  ) - \epsilon_{{\rm lcb}}  \leq    \rho^{(m,l)}   (  g({\bm x}  ,{\bm w}  )   ) & \leq 
{\rm ucb }^{(m,l)}_{t}  ({\bm x}  ) + \epsilon_{{\rm ucb}}, \label{eq:approx1} \\
\max_{   {\bm y}  \in {\rm Par}  ( \text{\bf LCB}_{t }(\hat{\Pi}_{t}  ) )   } {\rm dist} ({\bm y} ,  {\rm Par}  ( \text{\bf LCB}_{t }(\mathcal{X}  ) )  ) & \leq \epsilon_{{\rm PF}} , \label{eq:approx2}  \\ 
 \max_{ {\bm x}  \in \mathcal{X} }  a^{(\mathcal{X})} _{t}  ({\bm x} )    -   a^{(\mathcal{X})} _{t}  ({\bm x}_{t+1} ) & \leq \epsilon_{\mathcal{X}} , \label{eq:approx3}  \\ 
 \max_{ {\bm w}  \in \Omega_{{\bm x}_{t+1}} }  a^{(\Omega_{{\bm x}_{t+1}})} _{t}  ({\bm w} )    -   a^{(\Omega_{{\bm x}_{t+1}})} _{t}  ({\bm w}_{t+1} )   & \leq \epsilon_{\Omega} .  \label{eq:approx4} 
\end{align}

\begin{theorem}[Simulator setting]\label{thm:inference discrepancy with errors_app}
Suppose that the assumption in Lemma \ref{lem:HPCI_app} holds. 
Let $t \geq 0$, $m \in [M_f]$, $l \in [L_m]$, $\delta \in (0,1)$, and let $\tilde{\beta}^{1/2}_{m,t+1} $ be defined as in Lemma \ref{lem:HPCI_app}. 
In addition, let $\epsilon >0$ be a  predetermined desired accuracy parameter. 
Moreover, let $\epsilon_{{\rm lcb}}, \epsilon_{{\rm ucb}}, \epsilon_{{\rm PF}}, \epsilon_{\mathcal{X}}, \epsilon_{\Omega}$ be non-negative error parameters satisfying  \eqref{eq:approx1}--\eqref{eq:approx4}. 
Then, with probability at least $1-\delta$, the inequality $I_{t} \leq  a^{(\mathcal{X})}_{t} ({\bm x}_{t+1} )  +  \epsilon_{{\rm lcb}} + \epsilon_{{\rm ucb}}  + \epsilon_{\mathcal{X}}$ holds for any $t \geq 0$ and ${\bm x}_{t+1}$. 
Therefore, if  the stopping condition 
  satisfies at $T$ iterations, the inference discrepancy $I_{T} $ satisfies  $I_{T} \leq \epsilon +  \epsilon_{{\rm lcb}} + \epsilon_{{\rm ucb}}  + \epsilon_{\mathcal{X}}$ with probability at least $1-\delta$.  
\end{theorem}

\begin{theorem}[Simulator setting]\label{thm:termination with errors_app}
Suppose that the assumption in Theorem \ref{thm:inference discrepancy with errors_app} holds. 
Let $q: [0,\infty) \to [0,\infty) $ be a strictly increasing function satisfying $q(0)=0$ and \eqref{eq:w_condition_app}. 
Then, the inequality $  a^{(\mathcal{X})}_{\hat{t}} ({\bm x}_{{\hat{t}}+1} )  \leq \epsilon_{{\rm PF}} + q(\epsilon_{\Omega} + 
s_t
) $ holds for any $t \geq 0$ and  some $ {\hat{t}} \leq t$, where  $s_t$ is given by Theorem \ref{thm:termination_app}. 
Therefore, Algorithm \ref{alg:2}
 terminates after at most $T$ iterations, where $T$ is the smallest positive integer satisfying $ \epsilon_{{\rm PF}} + q(\epsilon_{\Omega} + 
s_T 
) \leq \epsilon$. 
%
%
%
\end{theorem}
Note that for Theorem \ref{thm:termination with errors_app}, the integer $T$ satisfying the theorem's last inequality does not always exist. 
However, the left hand side in  this inequality is merely an upper bound of $a^{(\mathcal{X} ) } _{\hat{t}} ({\bm x}_{{\hat{t}}+1} ) $. 
Thus, in some cases the actual value of  $a^{(\mathcal{X} ) } _{\hat{t}} ({\bm x}_{{\hat{t}}+1} ) $   satisfies $a^{(\mathcal{X} ) } _{\hat{t}} ({\bm x}_{{\hat{t}}+1} )  \leq \epsilon$ and the stopping condition is satisfied.

\paragraph{Uncontrollable Setting} 
We provide theoretical results for the uncontrollable setting. 
First, we define the following two additional conditions:
\begin{condition}\label{cond:ub}
Let ${\rm Nei} ({\bm a} ;r ) $ be an open ball with center ${\bm a} $ and radius $r >0$, where the distance is taken with respect to $L_1$-distance. 
Then, for any  ${\bm x}  \in \mathcal{X}$, $\hat{\bm w} \in \Omega_{ {\bm x}  }   $ and $\zeta >0$, 
 $P_{ {\bm w} } ({\bm x} ) $ satisfies
$$
  \mathbb{P}_{ P_{{\bm w}} ({{\bm x}})    }  [ {\bm w} \in  {\rm Nei} ( \hat{\bm w} ;  \zeta )  ]     >0, 
$$
where $ \mathbb{P}_{ P_{{\bm w}} ({{\bm x}})    } [\cdot ] $ is the probability measure with respect to $P_{{\bm w}} ({{\bm x}})$. 
\end{condition}
\begin{condition}\label{cond:lc_sd}
Let $L_{\sigma} $ be a positive number.  
Then, $\tilde{\sigma}^{(m)}_t ({\bm x},{\bm w} ) $ is an $L_{\sigma} $-data-independent-Lipschitz continuous, that is,  the following inequality holds for any $t \geq 1$, $ m \in [M_f]$ and $ \{ ({\bm x}_i,{\bm w}_i ) \}_{i=1}^t $:
$$
^\forall ({\bm x},{\bm w} ), (\tilde{\bm x},\tilde{\bm w} ) \in \mathcal{X} \times \Omega,     |   \tilde{\sigma}^{(m)}_t ({\bm x},{\bm w} ) - \tilde{\sigma}^{(m)}_t (\tilde{\bm x},\tilde{\bm w} )        | \leq L_{\sigma}   \| ({\bm x}^\top,{\bm w}^\top )^\top - (\tilde{\bm x}^\top,\tilde{\bm w}^\top )  \|_1  
$$
\end{condition}
Condition \ref{cond:ub} implies that the support of $P_{\bm w} ({\bm x} ) $ is equal to $\Omega_{\bm w} $. 
The assumption that the support of the distribution of ${\bm w}$ and the the set of ${\bm w}$ are the same is also used in existing studies that conduct theoretical analysis of BOs for risk measures under IU \citep{nguyen2021value,pmlr-v162-inatsu22a}.
Similarly, Condition \ref{cond:lc_sd} is introduced by \cite{kusakawa2022bayesian}, and they proved that Condition \ref{cond:lc_sd} holds if the linear, Gaussian or Mat\'{e}rn (with parameter $\nu >1$) is used. 
Their proof is given under constant variance of the normal error distribution for GP models, but similar arguments can be derived in the  setting considered in this section.  
We also define a maximal $\zeta$-separated subset of $\Omega_{{\bm x}}$:
\begin{definition}
Let $\zeta $ be a positive number. Then, a subset $S \subset \Omega_{{\bm x}}$ is called the maximal $\zeta$-separated subset of $\Omega_{\bm x}$, if the following holds:
\begin{enumerate}
\item For any ${\bm w},{\bm w}^\prime \in S$, ${\bm w} \neq {\bm w}^\prime   \Rightarrow \|  {\bm w} - {\bm w}^\prime \|_1 > \zeta $. 
\item For any ${\bm w} \in \Omega_{{\bm x} }$, there exists ${\bm w}^\prime \in S$ such that $\| {\bm w}-{\bm w}^\prime \|_1 \leq \zeta $. 
\end{enumerate}
\end{definition}
Note that a compact set $A$ has a maximal $\zeta$-separated  subset. 
Let $\mathcal{S} (\Omega_{{\bm x} } ; \zeta ) $ be a maximal $\zeta$-separated subset of $ \Omega_{{\bm x} }$. 
From Condition \ref{cond:ub} and compactness of $\Omega_{ {\bm x} }$, for any $\zeta >0 $ and ${\bm x} \in \mathcal{X}$, the following holds: 
\begin{equation}
\min_{\hat{{\bm w}}   \in  \mathcal{S} (\Omega_{{\bm x} } ; \zeta )   } \mathbb{P} _{ P_{{\bm w}} ({\bm x} )  }     [     {\bm w} \in  {\rm Nei} ( \hat{\bm w}_i ;  \zeta/2 )     ]  \equiv \underline{p_{{\bm x},\zeta}} >0. \label{eq:x_zeta}
\end{equation}
In contrast, \eqref{eq:x_zeta} does not necessarily guarantee $\inf_{{\bm x} \in \mathcal{X}} \underline{p_{{\bm x},\zeta}} >0$. 
However, $\inf_{{\bm x} \in \mathcal{X}} \underline{p_{{\bm x},\zeta}} =0$ implies that given $\zeta >0$ and for any $\nu >0$, there exist an open ball ${\rm Nei } (\hat{\bm w};\zeta/2 ) $ and $\hat{\bm x} \in \mathcal{X}$ such that $ \mathcal{P}_{P_{{\bm w}} (\hat{\bm x})} [{\bm w} \in {\rm Nei } (\hat{\bm w};\zeta/2 ) ]   < \nu$. 
This means that the probability of ${\bm w}$ realizes to the open ball with   radius $\zeta $ can be as small as desired.
Thus, to avoid this extreme case, we assume 
\begin{equation}
\inf_{{\bm x} \in \mathcal{X}} \underline{p_{{\bm x},\zeta}} \equiv \underline{p_{\zeta}} >0. \label{eq:p_under}
\end{equation}
Then, the following theorems hold:
\begin{theorem}[Uncontrollable setting]\label{thm:termination_app_uncontrollable}
Suppose that the assumption in Theorem \ref{thm:inference discrepancy_app} holds. 
Let $q: [0,\infty) \to [0,\infty) $ be a strictly increasing function satisfying $q(0)=0$ and \eqref{eq:w_condition_app}. 
Assume that Condition  \ref{cond:ub} and \ref{cond:lc_sd} hold. 
Let $\zeta _1, \ldots , \zeta_t $ be  positive numbers and  $\underline{p_{\zeta_1}} , \ldots , \underline{p_{\zeta_t}}$ be  numbers defined  by \eqref{eq:p_under}. 
Let $\underline{\tilde{p}_{\zeta_i}} =   \min_{1 \leq j \leq i } \underline{p_{\zeta_j}} $, $\tilde{\beta}^{1/2}_t = \max_{1 \leq m \leq M_f } \tilde{\beta}^{1/2}_{m,t} $, $    \tilde{\kappa}_t = \max_{1 \leq m \leq M_f}  \tilde{\kappa}^{(m)}_t  $ and 
define
\begin{align*}
\hat{s}_t =  \frac{2 M_f  L_{\sigma} \tilde{\beta}^{1/2}_{t+1} (1+    \underline{\tilde{p}_{\zeta_{t+1}}}^{-1}   )}{t+1} \sum_{i=1}^{t+1}   \zeta_i +   \frac{16J \log (8J/\delta)  \tilde{\beta}^{1/2}_{t+1} \underline{\tilde{p}_{\zeta_{t+1}}}^{-1}  }{t+1} +
\sqrt{\frac{   \hat{C}  \underline{\tilde{p}_{\zeta_{t+1}}}^{-2}   \tilde{\beta}_{t+1} \tilde{\kappa}_{t+1} }{   t+1}}, 
\end{align*}
 where $J = M_f \max \{1, \lambda^{-1}_1, \ldots , \lambda^{-1}_{M_f} \}$,  $\hat{C} = M_f \max_{1 \leq m \leq M_f} \hat{C}_m $ and $\hat{C}_m = \frac{32M_f}{  \log (1+ \lambda^{-1}_m \underline{\tau}^{-2}) }$. 
Then, with probability at least $1-\delta$, the inequality $  a^{(\mathcal{X})}_{\hat{t}} ({\bm x}_{{\hat{t}}+1} )  \leq q(\hat{s}_t) $ holds for any $t \geq 0$ and some $ \hat{t} \leq t$. 
Therefore,  with probability at least $1-\delta$, Algorithm \ref{alg:2}
 terminates after at most $T$ iterations, where $T$ is the smallest positive integer satisfying $q(\hat{s}_T) \leq \epsilon$. 
\end{theorem}
In Theorem \ref{thm:termination_app_uncontrollable}, the choice of $\zeta_1,\ldots , \zeta_t$ is important and must be chosen that $\hat{s}_t$ converges to 0. 
The simplest example is the case where $\Omega_{\bm x} $ is a finite set and equal to $\Omega$ for all ${\bm x} \in \mathcal{X}$.  
In this case, noting that $\lim_{ \zeta \to 0 }  \underline{p_\zeta}  >0$ and  $\sum_{t=1}^{\infty} \zeta_t = t^{-2} = \pi^2/6$, $\hat{s}_t $ converges to 0  when $\tilde{\beta}^{1/2}_t $ and $ \tilde{\beta}_t \tilde{\kappa}_t $ are sublinear.    
\cite{pmlr-v162-inatsu22a} used the finiteness assumption for  set of the environmental variables in theoretical analysis for uncontrollable settings under  IU. 
On the other hand, \cite{iwazaki2020mean}  considered the Bayes risk and standard deviation risk under the uncontrollable setting, and they derived the similar theoretical result without the finiteness assumption.  
Their approach can be used for moment-based risk measures such as Bayes risk, but not for quantile-based methods such as the worst-case risk.
As another example, when $\Omega_{ \bm x} = \Omega =[0,1]$ and $P_{\bm w} $ follows the uniform distribution on $\Omega$, the orders of $\underline{\tilde{p}_{\zeta_t}}^{-1} $ and $\sum_{i=1}^t \zeta_i $ are respectively $ \log t $ and  $t /\log t $ if $\zeta_i = 1/(\log i) $. 
Then, the dominant term of $\hat{s}_t $ is the first term and its order is $\tilde{\beta}^{1/2}_t$. 
Recently, \cite{pmlr-v202-takeno23a} has proposed a method in which $\tilde{\beta}_t$ does not diverge to infinity   by stochastically sampling $\tilde{\beta}_t$ under the assumption that the true black-box function follows GP.
Since their method is not an RKHS setting, nor is it a multi-objective optimization setting, it is not clear whether it is applicable to our setting, but it is one direction to consider.

\begin{theorem}[Uncontrollable setting]\label{thm:inference discrepancy with errors_app_uncontrollable}
Suppose that the assumption in Lemma \ref{lem:HPCI_app} holds. 
Let $t \geq 0$, $m \in [M_f]$, $l \in [L_m]$, $\delta \in (0,1)$, and let $\tilde{\beta}^{1/2}_{m,t+1} $ be defined as in Lemma \ref{lem:HPCI_app}. 
In addition, let $\epsilon >0$ be a  predetermined desired accuracy parameter. 
Moreover, let $\epsilon_{{\rm lcb}}, \epsilon_{{\rm ucb}}, \epsilon_{{\rm PF}}, \epsilon_{\mathcal{X}}$ be non-negative error parameters satisfying  \eqref{eq:approx1}--\eqref{eq:approx3}. 
Then, with probability at least $1-\delta$, the inequality $I_{t} \leq  a^{(\mathcal{X})}_{t} ({\bm x}_{t+1} )  +  \epsilon_{{\rm lcb}} + \epsilon_{{\rm ucb}}  + \epsilon_{\mathcal{X}}$ holds for any $t \geq 0$ and ${\bm x}_{t+1}$. 
Therefore, if  the stopping condition 
  satisfies at $T$ iterations, the inference discrepancy $I_{T} $ satisfies  $I_{T} \leq \epsilon +  \epsilon_{{\rm lcb}} + \epsilon_{{\rm ucb}}  + \epsilon_{\mathcal{X}}$ with probability at least $1-\delta$.  
\end{theorem}

\begin{theorem}[Uncontrollable setting]\label{thm:termination with errors_app_uncontrollable}
Suppose that the assumptions in Theorem \ref{thm:termination_app_uncontrollable} and Theorem \ref{thm:inference discrepancy with errors_app_uncontrollable} holds. 
Let $q: [0,\infty) \to [0,\infty) $ be a strictly increasing function satisfying $q(0)=0$ and \eqref{eq:w_condition_app}. 
Then, with probability at least $1-\delta$, the inequality $  a^{(\mathcal{X})}_{\hat{t}} ({\bm x}_{{\hat{t}}+1} )  \leq \epsilon_{{\rm PF}} + q(
\hat{s}_t
) $ holds for any $t \geq 0$ and  some $ {\hat{t}} \leq t$, where  $\hat{s}_t$ is given by Theorem \ref{thm:termination_app_uncontrollable}. 
Therefore, with probability at least $1-\delta$, Algorithm \ref{alg:2}
 terminates after at most $T$ iterations, where $T$ is the smallest positive integer satisfying $ \epsilon_{{\rm PF}} + q( 
\hat{s}_T 
) \leq \epsilon$. 
%
%
%
\end{theorem}

\section{Proofs}\label{app:proofs}
In this section, we prove all theorems, lemmas and the results in Table \ref{tab:decomposition_result} and \ref{tab:q}. 

\subsection{Proof of Table \ref{tab:decomposition_result} and \ref{tab:q}}
In this proof, we omit the notation  \textasciitilde  and $(m)$ for simplicity. 
Let ${\bm x} \in \mathcal{X}$, ${\bm w} \in \Omega_{{\bm x}} $, $ t \geq 0$ and $\beta^{1/2}_{t+1}  \geq 0$. 
Assume that $l_{t,{\bm x},{\bm w} } \leq f({\bm x},{\bm w} ) \leq u_{t,{\bm x},{\bm w} } $, where 
$ l_{t,{\bm x},{\bm w} }  = \mu_t ({\bm x},{\bm w} ) - \beta^{1/2}_{t+1} \sigma_t ({\bm x},{\bm w} ) $ and 
$ u_{t,{\bm x},{\bm w} }  = \mu_t ({\bm x},{\bm w} ) + \beta^{1/2}_{t+1} \sigma_t ({\bm x},{\bm w} ) $. 
\paragraph{Bayes Risk} 
Since ${\bm w} $ is a random variable, $l_{t,{\bm x},{\bm w} }$, $u_{t,{\bm x},{\bm w} }$ and $f({\bm x},{\bm w} )$ are also random variables. 
Hence, from the monotonicity of expectation and $l_{t,{\bm x},{\bm w} } \leq f({\bm x},{\bm w} ) \leq u_{t,{\bm x},{\bm w} } $, we have 
$$
{\rm lcb}_t ({\bm x} ) \equiv \mathbb{E} [ l_{t,{\bm x},{\bm w} } ] \leq \mathbb{E}[ f({\bm x},{\bm w} ) ] \leq 
 \mathbb{E} [ u_{t,{\bm x},{\bm w} } ]  \equiv {\rm ucb}_t ({\bm x} ) .
$$
In addition, from the definition of $l_{t,{\bm x},{\bm w} }$ and $u_{t,{\bm x},{\bm w} }$, we get 
$$
0 \leq  {\rm ucb}_t ({\bm x} ) -  {\rm lcb}_t ({\bm x} ) = \mathbb{E}[  u_{t,{\bm x},{\bm w} } - l_{t,{\bm x},{\bm w} } ] =
\mathbb{E}[2 \beta^{1/2}_{t+1} \sigma_t ({\bm x},{\bm w} ) ] \leq \max_{{\bm w} \in \Omega_{{\bm x}}}  2 \beta^{1/2}_{t+1} \sigma_t ({\bm x},{\bm w} ) .
$$
\paragraph{Worst-case} 
From the definition of infimum, noting that  $l_{t,{\bm x},{\bm w} } \leq f({\bm x},{\bm w} ) \leq u_{t,{\bm x},{\bm w} } $, we obtain 
$$
{\rm lcb}_t ({\bm x} ) \equiv  \inf_{{\bm w} \in \Omega_{{\bm x}}} l_{t,{\bm x},{\bm w} }  \leq 
  \inf_{{\bm w} \in \Omega_{{\bm x}}} f({\bm x},{\bm w} ) 
\leq 
 \inf_{{\bm w} \in \Omega_{{\bm x}}} u_{t,{\bm x},{\bm w} }  \equiv  {\rm ucb}_t ({\bm x} ). 
$$
Moreover, from the property of infimum, for any $\epsilon >0$, there exists $\hat{\bm w} \in \Omega_{{\bm x}} $ such that 
$ l_{t,{\bm x},\hat{\bm w} } \leq {\rm lcb} _t  ({\bm x} ) + \epsilon$. 
Therefore, noting that ${\rm ucb}_t ({\bm x} ) \leq u_{t,{\bm x},  \hat{\bm w}}$, we get 
$$
{\rm ucb}_t ({\bm x} ) - {\rm lcb}_t ({\bm x} ) \leq u_{t,{\bm x},  \hat{\bm w}} - l_{t,{\bm x},  \hat{\bm w}}  +\epsilon
 = 2 \beta^{1/2}_{t+1} \sigma_t ({\bm x},\hat{\bm w} ) + \epsilon \leq \max_{{\bm w} \in \Omega_{{\bm x}}}  2 \beta^{1/2}_{t+1} \sigma_t ({\bm x},{\bm w} ) + \epsilon.
$$
Since $\epsilon$ is an arbitrary positive number, we have 
$$
0 \leq {\rm ucb}_t ({\bm x} ) - {\rm lcb}_t ({\bm x} ) \leq  \max_{{\bm w} \in \Omega_{{\bm x}}}  2 \beta^{1/2}_{t+1} \sigma_t ({\bm x},{\bm w} ) .
$$

\paragraph{Best-case} 
From the definition of supremum, noting that  $l_{t,{\bm x},{\bm w} } \leq f({\bm x},{\bm w} ) \leq u_{t,{\bm x},{\bm w} } $, we obtain 
$$
{\rm lcb}_t ({\bm x} ) \equiv  \sup_{{\bm w} \in \Omega_{{\bm x}}} l_{t,{\bm x},{\bm w} }  \leq 
  \sup_{{\bm w} \in \Omega_{{\bm x}}} f({\bm x},{\bm w} ) 
\leq 
 \sup_{{\bm w} \in \Omega_{{\bm x}}} u_{t,{\bm x},{\bm w} }  \equiv  {\rm ucb}_t ({\bm x} ). 
$$
Moreover, from the property of supremum, for any $\epsilon >0$, there exists $\hat{\bm w} \in \Omega_{{\bm x}} $ such that 
$    {\rm ucb} _t  ({\bm x} ) -\epsilon \leq  u_{t,{\bm x},\hat{\bm w} } $. 
Therefore, noting that ${\rm lcb}_t ({\bm x} ) \geq l_{t,{\bm x},  \hat{\bm w}}$, we get 
$$
{\rm ucb}_t ({\bm x} ) - {\rm lcb}_t ({\bm x} ) \leq u_{t,{\bm x},  \hat{\bm w}} - l_{t,{\bm x},  \hat{\bm w}}  +\epsilon
 = 2 \beta^{1/2}_{t+1} \sigma_t ({\bm x},\hat{\bm w} ) + \epsilon \leq \max_{{\bm w} \in \Omega_{{\bm x}}}  2 \beta^{1/2}_{t+1} \sigma_t ({\bm x},{\bm w} ) + \epsilon.
$$
Since $\epsilon$ is an arbitrary positive number, we have 
$$
0 \leq {\rm ucb}_t ({\bm x} ) - {\rm lcb}_t ({\bm x} ) \leq  \max_{{\bm w} \in \Omega_{{\bm x}}}  2 \beta^{1/2}_{t+1} \sigma_t ({\bm x},{\bm w} ) .
$$

\paragraph{$\alpha$-value-at-risk} 
Let $\alpha \in (0,1)$. 
For any $b \in \mathbb{R} $, $f({\bm x},{\bm w} ) \leq u_{t,{\bm x},{\bm w} } $ implies that $  \mathbb{P} (u_{t,{\bm x},{\bm w} } \leq b) \leq \mathbb{P} (f({\bm x},{\bm w}) \leq b) $. 
Thus, letting ${\rm ucb}_t ({\bm x} ) \equiv  \inf \{  b \in \mathbb{R} \mid \alpha \leq \mathbb{P} (u_{t,{\bm x},{\bm w} } \leq b) \}$, 
we obtain 
$$
\alpha \leq \mathbb{P} (u_{t,{\bm x},{\bm w} } \leq {\rm ucb}_t ({\bm x} ) ) \leq \mathbb{P} (f({\bm x},{\bm w}) \leq {\rm ucb}_t ({\bm x} ) ) .  
$$
This implies that 
$$
\inf \{  b \in \mathbb{R} \mid \alpha \leq \mathbb{P} (f({\bm x},{\bm w}) \leq b ) \} \leq  {\rm ucb}_t ({\bm x} ) .
$$
By using the same argument, we get 
$$
{\rm lcb}_t ({\bm x} ) \equiv  \inf \{  b \in \mathbb{R} \mid \alpha \leq \mathbb{P} (l_{t,{\bm x},{\bm w} } \leq b) \}
\leq 
\inf \{  b \in \mathbb{R} \mid \alpha \leq \mathbb{P} (f({\bm x},{\bm w}) \leq b ) \} .
$$
Furthermore, noting that the definition of  $l_{t,{\bm x},{\bm w} } $ and  $u_{t,{\bm x},{\bm w} } $, we get 
$$
u_{t,{\bm x},{\bm w} } \leq l_{t,{\bm x},{\bm w} } +  \max_{{\bm w} \in \Omega_{{\bm x}}} 2 \beta^{1/2}_{t+1} \sigma_t ({\bm x},{\bm w} ) . 
$$
Therefore, we have 
\begin{align*}
0 \leq {\rm ucb}_t ({\bm x} ) - {\rm lcb}_t ({\bm x} ) & =  
\inf \{  b \in \mathbb{R} \mid \alpha \leq \mathbb{P}  ( u_{t,{\bm x},{\bm w} } \leq b ) \} -
\inf \{  b \in \mathbb{R} \mid \alpha \leq \mathbb{P}  ( l_{t,{\bm x},{\bm w} } \leq b ) \}  \\
&\leq \inf \{  b \in \mathbb{R} \mid \alpha \leq \mathbb{P}  ( l_{t,{\bm x},{\bm w} } + \max_{{\bm w} \in \Omega_{{\bm x}}} 2 \beta^{1/2}_{t+1} \sigma_t ({\bm x},{\bm w} ) \leq b ) \} -
\inf \{  b \in \mathbb{R} \mid \alpha \leq \mathbb{P}  ( l_{t,{\bm x},{\bm w} } \leq b ) \}  \\
&=  {\rm lcb}_t ({\bm x} )  +  \max_{{\bm w} \in \Omega_{{\bm x}}} 2 \beta^{1/2}_{t+1} \sigma_t ({\bm x},{\bm w} ) - {\rm lcb}_t ({\bm x} ) = \max_{{\bm w} \in \Omega_{{\bm x}}} 2 \beta^{1/2}_{t+1} \sigma_t ({\bm x},{\bm w} ). 
\end{align*}

\paragraph{$\alpha$-conditional value-at-risk} 
Let $\alpha \in (0,1)$. 
From \cite{nguyen2021optimizing}, $\alpha $-conditional value-at-risk for $f({\bm x},{\bm w} )$ can be written as follows:
$$
\frac{1}{\alpha}  \int_0 ^\alpha v_f ({\bm x}; \alpha^\prime )   {\rm d} \alpha^\prime . 
$$
Thus, we have 
$$
  {\rm lcb}_t ({\bm x} ) \equiv 
\frac{1}{\alpha}  \int_0 ^\alpha v_{l_t} ({\bm x}; \alpha^\prime )   {\rm d} \alpha^\prime 
\leq
\frac{1}{\alpha}  \int_0 ^\alpha v_f ({\bm x}; \alpha^\prime )   {\rm d} \alpha^\prime 
\leq 
\frac{1}{\alpha}  \int_0 ^\alpha v_{u_t} ({\bm x}; \alpha^\prime )   {\rm d} \alpha^\prime  \equiv {\rm ucb}_t ({\bm x} ). 
$$
In addition, noting that $ v_{u_t} ({\bm x}; \alpha^\prime ) -  v_{l_t} ({\bm x}; \alpha^\prime ) \leq \max_{{\bm w} \in \Omega_{{\bm x}}} 2 \beta^{1/2}_{t+1} \sigma_t ({\bm x},{\bm w} )
$, we get 
$$
0 \leq  {\rm ucb}_t ({\bm x} ) - {\rm lcb}_t ({\bm x} ) 
=
\frac{1}{\alpha} \int_0 ^\alpha (    v_{u_t} ({\bm x}; \alpha^\prime ) -  v_{l_t} ({\bm x}; \alpha^\prime )     ) {\rm d} \alpha^\prime 
\leq \frac{1}{\alpha} \int_0 ^\alpha \max_{{\bm w} \in \Omega_{{\bm x}}} 2 \beta^{1/2}_{t+1} \sigma_t ({\bm x},{\bm w} ) {\rm d} \alpha^\prime = \max_{{\bm w} \in \Omega_{{\bm x}}} 2 \beta^{1/2}_{t+1} \sigma_t ({\bm x},{\bm w} ).
$$

\paragraph{Mean Absolute Deviation, Standard Deviation and Variance} 
From $l_{t,{\bm x},{\bm w} } \leq f({\bm x},{\bm w} ) \leq u_{t,{\bm x},{\bm w} } $, we get 
$$
-\mathbb{E}[ u_{t,{\bm x},{\bm w} } ] \leq -\mathbb{E}[ f({\bm x},{\bm w} ) ]   \leq  -\mathbb{E}[ l_{t,{\bm x},{\bm w} } ]. 
$$
Hence, we have 
$$
\check{l}_{t,{\bm x},{\bm w} } \equiv  l_{t,{\bm x},{\bm w}}  -\mathbb{E}[ u_{t,{\bm x},{\bm w} } ] 
\leq 
 f({\bm x},{\bm w})  -\mathbb{E}[ f({\bm x},{\bm w} ) ]  
\leq 
u_{t,{\bm x},{\bm w}}  -\mathbb{E}[ l_{t,{\bm x},{\bm w} } ]  \equiv \check{u}_{t,{\bm x},{\bm w} }   . 
$$
Therefore, we obtain 
$$
| f({\bm x},{\bm w})  -\mathbb{E}[ f({\bm x},{\bm w} ) ]  |
\leq 
\max \{   |\check{l}_{t,{\bm x},{\bm w} } | , |\check{u}_{t,{\bm x},{\bm w} }  | \} .
$$
Similarly, if $ \check{l}_{t,{\bm x},{\bm w} } <0 $ and $ \check{u}_{t,{\bm x},{\bm w} } >0 $, then we  have 
$$
| f({\bm x},{\bm w})  -\mathbb{E}[ f({\bm x},{\bm w} ) ]  | \geq 0.
$$
On the other hand, if $ \check{l}_{t,{\bm x},{\bm w} } \geq 0 $ or $ \check{u}_{t,{\bm x},{\bm w} } \leq 0$, then we get 
$$
| f({\bm x},{\bm w})  -\mathbb{E}[ f({\bm x},{\bm w} ) ]  | \geq \min \{   |\check{l}_{t,{\bm x},{\bm w} } | , |\check{u}_{t,{\bm x},{\bm w} }  | \} .
$$
Thus, by combining these, for any $\check{l}_{t,{\bm x},{\bm w} }  $ and $ \check{u}_{t,{\bm x},{\bm w} }  $, we obtain 
\begin{align*}
| f({\bm x},{\bm w})  -\mathbb{E}[ f({\bm x},{\bm w} ) ]  | & \geq \min \{   |\check{l}_{t,{\bm x},{\bm w} } | , |\check{u}_{t,{\bm x},{\bm w} }  | \}    -        \max \{ \min \{  - \check{l}_{t,{\bm x},{\bm w} } ,   \check{u}_{t,{\bm x},{\bm w} } \} , 0 \}  \\ 
& \equiv 
\min \{   |\check{l}_{t,{\bm x},{\bm w} } | , |\check{u}_{t,{\bm x},{\bm w} }  | \}    -        {\rm STR} (\check{l}_{t,{\bm x},{\bm w} },\check{u}_{t,{\bm x},{\bm w} }   )     .
\end{align*}
Hence, we have 
\begin{align*}
{\rm lcb}_t ({\bm x} ) \equiv \mathbb{E} [  \min \{   |\check{l}_{t,{\bm x},{\bm w} } | , |\check{u}_{t,{\bm x},{\bm w} }  | \}    -        {\rm STR} (\check{l}_{t,{\bm x},{\bm w} },\check{u}_{t,{\bm x},{\bm w} }   )    ] 
&\leq 
\mathbb{E}[  | f({\bm x},{\bm w})  -\mathbb{E}[ f({\bm x},{\bm w} ) ]  | ] \\
&\leq 
\mathbb{E} [ \max \{   |\check{l}_{t,{\bm x},{\bm w} } | , |\check{u}_{t,{\bm x},{\bm w} }  | \}    ]    \equiv 
{\rm ucb}_t ({\bm x} ). 
\end{align*}
Moreover, noting that 
\begin{align*}
& \max \{   |\check{l}_{t,{\bm x},{\bm w} } | , |\check{u}_{t,{\bm x},{\bm w} }  | \} - (  \min \{   |\check{l}_{t,{\bm x},{\bm w} } | , |\check{u}_{t,{\bm x},{\bm w} }  | \}    -        {\rm STR} (\check{l}_{t,{\bm x},{\bm w} },\check{u}_{t,{\bm x},{\bm w} }   )  )   \\
&\leq      \check{u}_{t,{\bm x},{\bm w} } -  \check{l}_{t,{\bm x},{\bm w} } 
=(u_{t,{\bm x},{\bm w} }   -  l_{t,{\bm x},{\bm w} } ) + \mathbb{E}[   u_{t,{\bm x},{\bm w} }   -  l_{t,{\bm x},{\bm w} }    ] \\
&= 2 \beta^{1/2}_{t+1}  \sigma_t ({\bm x},{\bm w} )   +  \mathbb{E}[   2 \beta^{1/2}_{t+1}  \sigma_t ({\bm x},{\bm w} )  ] 
\leq 2  \max_{{\bm w} \in \Omega_{{\bm x}}} 2 \beta^{1/2}_{t+1} \sigma_t ({\bm x},{\bm w} ) ,
\end{align*}
we obtain 
$$
0 \leq {\rm ucb}_t ({\bm x} ) -{\rm lcb}_t ({\bm x} )  \leq \mathbb{E}[   2  \max_{{\bm w} \in \Omega_{{\bm x}}} 2 \beta^{1/2}_{t+1} \sigma_t ({\bm x},{\bm w} ) ]   =  2  \max_{{\bm w} \in \Omega_{{\bm x}}} 2 \beta^{1/2}_{t+1} \sigma_t ({\bm x},{\bm w} ) .
$$

Next, we prove the case of the standard deviation. 
By using the same argument as in the mean absolute deviation, we get 
$$
  \min \{    |\check{l}_{t,{\bm x},{\bm w} } |^2,  |\check{u}_{t,{\bm x},{\bm w} } |^2 \}  - {\rm STR} ^2 ( \check{l}_{t,{\bm x},{\bm w} } , \check{u}_{t,{\bm x},{\bm w} } ) 
\leq | f({\bm x},{\bm w})  -\mathbb{E}[ f({\bm x},{\bm w} ) ]  |^2 
\leq   \max \{    |\check{l}_{t,{\bm x},{\bm w} } |^2,  |\check{u}_{t,{\bm x},{\bm w} } |^2 \} .
$$
Therefore, we have 
\begin{align*}
{\rm lcb }_t ({\bm x} ) &\equiv 
\sqrt{\mathbb{E}[    \min \{    |\check{l}_{t,{\bm x},{\bm w} } |^2,  |\check{u}_{t,{\bm x},{\bm w} } |^2 \}  - {\rm STR} ^2 ( \check{l}_{t,{\bm x},{\bm w} } , \check{u}_{t,{\bm x},{\bm w} } )      ]} \\
&\leq 
\sqrt{\mathbb{E}[     | f({\bm x},{\bm w})  -\mathbb{E}[ f({\bm x},{\bm w} ) ]  |^2   ]  } 
\leq 
\sqrt{\mathbb{E}[    \max \{    |\check{l}_{t,{\bm x},{\bm w} } |^2,  |\check{u}_{t,{\bm x},{\bm w} } |^2 \}       ]} 
\equiv {\rm ucb}_t ({\bm x} ).
\end{align*}
 In addition, noting that $\sqrt{u} - \sqrt{v} \leq \sqrt{u-v} $ for any $u \geq v \geq 0$, we obtain 
$$
0 \leq  {\rm ucb}_t ({\bm x} ) - {\rm lcb}_t ({\bm x} ) 
\leq 
\sqrt{
\mathbb{E}[    \max \{    |\check{l}_{t,{\bm x},{\bm w} } |^2,  |\check{u}_{t,{\bm x},{\bm w} } |^2 \}       ] 
- \mathbb{E}[    \min \{    |\check{l}_{t,{\bm x},{\bm w} } |^2,  |\check{u}_{t,{\bm x},{\bm w} } |^2 \}  - {\rm STR} ^2 ( \check{l}_{t,{\bm x},{\bm w} } , \check{u}_{t,{\bm x},{\bm w} } )      ]
}.
$$
From Equation (17) of Appendix A.2 in \cite{iwazaki2020mean},  we have 
\begin{align*}
&\mathbb{E}[    \max \{    |\check{l}_{t,{\bm x},{\bm w} } |^2,  |\check{u}_{t,{\bm x},{\bm w} } |^2 \}       ] 
- \mathbb{E}[    \min \{    |\check{l}_{t,{\bm x},{\bm w} } |^2,  |\check{u}_{t,{\bm x},{\bm w} } |^2 \}  - {\rm STR} ^2 ( \check{l}_{t,{\bm x},{\bm w} } , \check{u}_{t,{\bm x},{\bm w} } )      ] \\
&
\leq 16 B \beta^{1/2} _{t+1} \mathbb{E}[  \sigma_t ({\bm x},{\bm w} ) ]  + 20 \beta_{t+1} \mathbb{E}[ \sigma^2_t ({\bm x},{\bm w} ) ] \\
&\leq 16 B \beta^{1/2}_{t+1}  \max_{{\bm w} \in \Omega_{{\bm x}}}  \sigma_t ({\bm x},{\bm w} ) + 20 \beta_{t+1}   \max_{{\bm w} \in \Omega_{{\bm x}}} \sigma^2_t ({\bm x},{\bm w} ) 
=
8B  \max_{{\bm w} \in \Omega_{{\bm x}}}  2 \beta^{1/2}_{t+1}  \sigma_t ({\bm x},{\bm w} ) + 5 \left (  \max_{{\bm w} \in \Omega_{{\bm x}}}  2 \beta^{1/2}_{t+1}  \sigma_t ({\bm x},{\bm w} )   \right )^2 .
\end{align*}
Hence, we get 
$$
0 \leq  {\rm ucb}_t ({\bm x} ) - {\rm lcb}_t ({\bm x} ) 
\leq  
\sqrt{
8B  \max_{{\bm w} \in \Omega_{{\bm x}}}  2 \beta^{1/2}_{t+1}  \sigma_t ({\bm x},{\bm w} ) + 5 \left (  \max_{{\bm w} \in \Omega_{{\bm x}}}  2 \beta^{1/2}_{t+1}  \sigma_t ({\bm x},{\bm w} )   \right )^2
}.
$$

Finally, we prove the case of the variance. 
By using the same argument as in the standard deviation, we get 
\begin{align*}
{\rm lcb }_t ({\bm x} ) &\equiv 
\mathbb{E}[    \min \{    |\check{l}_{t,{\bm x},{\bm w} } |^2,  |\check{u}_{t,{\bm x},{\bm w} } |^2 \}  - {\rm STR} ^2 ( \check{l}_{t,{\bm x},{\bm w} } , \check{u}_{t,{\bm x},{\bm w} } )      ] \\
&\leq 
\mathbb{E}[     | f({\bm x},{\bm w})  -\mathbb{E}[ f({\bm x},{\bm w} ) ]  |^2   ]  
\leq 
\mathbb{E}[    \max \{    |\check{l}_{t,{\bm x},{\bm w} } |^2,  |\check{u}_{t,{\bm x},{\bm w} } |^2 \}       ]
\equiv {\rm ucb}_t ({\bm x} ). 
\end{align*}
Furthermore, we obtain 
\begin{align*}
0 \leq {\rm ucb}_t ({\bm x}) -  {\rm lcb}_t ({\bm x} ) &= \mathbb{E}[    \max \{    |\check{l}_{t,{\bm x},{\bm w} } |^2,  |\check{u}_{t,{\bm x},{\bm w} } |^2 \}       ] 
- \mathbb{E}[    \min \{    |\check{l}_{t,{\bm x},{\bm w} } |^2,  |\check{u}_{t,{\bm x},{\bm w} } |^2 \}  - {\rm STR} ^2 ( \check{l}_{t,{\bm x},{\bm w} } , \check{u}_{t,{\bm x},{\bm w} } )      ] \\
&
\leq 16 B \beta^{1/2} _{t+1} \mathbb{E}[  \sigma_t ({\bm x},{\bm w} ) ]  + 20 \beta_{t+1} \mathbb{E}[ \sigma^2_t ({\bm x},{\bm w} ) ] \\
&\leq 16 B \beta^{1/2}_{t+1}  \max_{{\bm w} \in \Omega_{{\bm x}}}  \sigma_t ({\bm x},{\bm w} ) + 20 \beta_{t+1}   \max_{{\bm w} \in \Omega_{{\bm x}}} \sigma^2_t ({\bm x},{\bm w} ) \\
&=
8B  \max_{{\bm w} \in \Omega_{{\bm x}}}  2 \beta^{1/2}_{t+1}  \sigma_t ({\bm x},{\bm w} ) + 5 \left (  \max_{{\bm w} \in \Omega_{{\bm x}}}  2 \beta^{1/2}_{t+1}  \sigma_t ({\bm x},{\bm w} )   \right )^2 .
\end{align*}

\paragraph{Distributionally Robust}
Let $P $ be a candidate distribution of $P_{\bm w} ({\bm x} ) $, and let $\mathcal{A}$ be a family of candidate distributions. 
Also let $F ({\bm x} ;P) $, ${\rm lcb}_t ({\bm x};P) $ and ${\rm ucb}_t ({\bm x};P) $ be respectively risk measure, and its lower and upper with respect to $P$. 
Define 
$$
F({\bm x} )  \equiv \inf_{ P \in \mathcal{A} } F ({\bm x};P), \ 
{\rm lcb}_t ({\bm x} ) \equiv  \inf_{ P \in \mathcal{A} } {\rm lcb}_t ({\bm x};P) , \ 
{\rm ucb}_t ({\bm x} ) \equiv  \inf_{ P \in \mathcal{A} } {\rm ucb}_t ({\bm x};P) .
$$
From the property of infimum, for any $\epsilon >0$, there exists a distribution $\hat{P}$ such that 
$$
{\rm ucb}_t ({\bm x}; \hat{P} ) \leq {\rm ucb}_t ({\bm x} ) + \epsilon. 
$$
Hence, we get 
$$
 F({\bm x} ) \leq F ({\bm x}; \hat{P} ) \leq {\rm ucb}_t ({\bm x}; \hat{P} )   \leq {\rm ucb}_t ({\bm x} ) + \epsilon. 
$$
Since $\epsilon $ is an arbitrary positive number, we obtain 
$$
 F({\bm x} ) \leq  {\rm ucb}_t ({\bm x} ) .
$$
Similarly,  we also get 
$$
{\rm lcb}_t ({\bm x} ) \leq 
 F({\bm x} ) .  
$$ 
Furthermore, for any $\eta >0$, there exists a distribution $\tilde{P}$ such that 
$$
{\rm lcb}_t ({\bm x}; \tilde{P} ) \leq {\rm lcb}_t ({\bm x} ) + \eta. 
$$
Thus, we get 
\begin{align*}
{\rm ucb}_t ({\bm x} ) - {\rm lcb}_t ({\bm x} ) \leq {\rm ucb}_t ({\bm x};\tilde{P} ) - {\rm lcb}_t ({\bm x};\tilde{P} ) + \eta 
\leq q \left (  \max_{{\bm w} \in \Omega_{{\bm x}}}  2 \beta^{1/2}_{t+1}  \sigma_t ({\bm x},{\bm w} ) ; F\right ) +\eta.
\end{align*}
Since $\eta$ is an arbitrary positive number, we have 
$$
{\rm ucb}_t ({\bm x} ) - {\rm lcb}_t ({\bm x} )  
\leq q \left (  \max_{{\bm w} \in \Omega_{{\bm x}}}  2 \beta^{1/2}_{t+1}  \sigma_t ({\bm x},{\bm w} ) ; F\right ) . 
$$

\paragraph{Monotone Lipschitz Map} 
Let $\mathcal{M}$ be a K-Lipschitz map, and let  $F ({\bm x} ) $, ${\rm lcb}_t ({\bm x}) $ and ${\rm ucb}_t ({\bm x}) $ be respectively risk measure, and its lower and upper. 
Then, from the monotonicity of $\mathcal{M}$, we have 
$$
\min \{ \mathcal{M}  ({\rm lcb}_t ({\bm x})  ) , \mathcal{M}  ({\rm ucb}_t ({\bm x})  ) \} \leq \mathcal{M} (F ({\bm x} )  ) 
\leq 
\max \{ \mathcal{M}  ({\rm lcb}_t ({\bm x})  ) , \mathcal{M}  ({\rm ucb}_t ({\bm x})  ) \}  .
$$
In addition, using the Lipschitz continuity  of $\mathcal{M}$ we get 
\begin{align*}
0 &\leq \max \{ \mathcal{M}  ({\rm lcb}_t ({\bm x})  ) , \mathcal{M}  ({\rm ucb}_t ({\bm x})  ) \}  - 
\min \{ \mathcal{M}  ({\rm lcb}_t ({\bm x})  ) , \mathcal{M}  ({\rm ucb}_t ({\bm x})  ) \} \\
&\leq |  \mathcal{M}  ({\rm lcb}_t ({\bm x})  ) - \mathcal{M}  ({\rm ucb}_t ({\bm x})  )   |
\leq K | {\rm ucb}_t ({\bm x}) -{\rm lcb}_t ({\bm x})     |  \leq K  q \left (  \max_{{\bm w} \in \Omega_{{\bm x}}}  2 \beta^{1/2}_{t+1}  \sigma_t ({\bm x},{\bm w} )  \right ) . 
\end{align*}

\paragraph{Weighted Sum} 
Let $\alpha_1,\alpha_2 \geq 0$, and let  $F_i ({\bm x} ) $, ${\rm lcb}_{t,i} ({\bm x}) $ and ${\rm ucb}_{t,i} ({\bm x}) $ be respectively risk measure, and its lower and upper with $i=1,2$. 
Then, noting that $\alpha_1, \alpha_2 \geq 0$, we obtain
$$
{\rm lcb}_t ({\bm x} ) \equiv \alpha_1 {\rm lcb}_{t,1} ({\bm x}) +  \alpha_2 {\rm lcb}_{t,2} ({\bm x})  
\leq 
\alpha_1 F_1 ({\bm x} ) +  \alpha_2 F_2 ({\bm x} )
\leq 
\alpha_1 {\rm ucb}_{t,1} ({\bm x}) +  \alpha_2 {\rm ucb}_{t,2} ({\bm x}) \equiv {\rm ucb}_t ({\bm x} ).
$$
Moreover, we get 
\begin{align*}
0 &\leq {\rm ucb}_t ({\bm x} ) - {\rm lcb}_t ({\bm x} ) 
= \alpha_1 (  {\rm ucb}_{t,1} ({\bm x}) - {\rm lcb}_{t,1} ({\bm x}) ) + 
\alpha_2 (  {\rm ucb}_{t,2} ({\bm x}) - {\rm lcb}_{t,2} ({\bm x}) ) \\
&\leq \alpha_1 q_1 \left (  \max_{{\bm w} \in \Omega_{{\bm x}}}  2 \beta^{1/2}_{t+1}  \sigma_t ({\bm x},{\bm w} ) \right ) +
\alpha_2 q_2 \left (  \max_{{\bm w} \in \Omega_{{\bm x}}}  2 \beta^{1/2}_{t+1}  \sigma_t ({\bm x},{\bm w} ) \right ).
\end{align*}

\paragraph{Probabilistic Threshold}
Let $\theta \in \mathbb{R}$ be a threshold. 
Then, $l_{t,{\bm x},{\bm w} } \leq f({\bm x},{\bm w} ) \leq u_{t,{\bm x},{\bm w} } $ implies that 
$$
{\rm lcb}_t ({\bm x} ) \equiv 
\mathbb{P}   (  l_{t,{\bm x},{\bm w} }  \geq \theta   ) \leq \mathbb{P}   ( f({\bm x},{\bm w} )  \geq \theta   ) \leq 
\mathbb{P}   (  u_{t,{\bm x},{\bm w} }  \geq \theta   ) \equiv {\rm ucb}_t ({\bm x} ).
$$

\subsection{Extension of Theorem E.4 in \cite{kusakawa2022bayesian}}
We show the extension of Theorem E.4 in \cite{kusakawa2022bayesian}. 
In this subsection, we use ${\bm x}$ and $\mathcal{X}$ as the input variable and set of all input variables, respectively. 
In Theorem E.4 in \cite{kusakawa2022bayesian}, they proved that if linear, Gaussian or  Mat\'{e}rn (with parameter $\nu >1$) kernel 
 is used, then the posterior standard deviation satisfies the $L_{\sigma}$-data-independent-Lipschitz continuity. 
They have assumed that the variance of an error distribution for GP models is  $\sigma^2 >0$ for any input ${\bm x} $. 
We show that this assumption can be relaxed to the assumption that the noise variance is positive and depends on ${\bm x}$. 
Since the relaxation of noise variance to the heteroscedastic setting does not affect any essential part of their proof, only the sketch  of the proof is given here.
Let ${\bm X}_t $ be a $t \times d$ matrix. 
Then, in their proof, $\sigma^2$ appears only within the formula given below:
$$
{\bm I}_d - {\bm X}^\top _t ( {\bm X}_t {\bm X}^\top_t + a^{-2} \sigma^2 {\bm I}_t )^{-1} {\bm X}_t,
$$
 where $a$ is some positive constant. 
They considered the singular value decomposition ${\bm X}_t = {\bm H}^\prime {\bm \Lambda} {\bm H}$ and calculated 
$$
{\bm I}_d - {\bm X}^\top _t ( {\bm X}_t {\bm X}^\top_t + a^{-2} \sigma^2 {\bm I}_t )^{-1} {\bm X}_t = {\bm H} {\bm \Theta} {\bm H}^\top,
$$
where ${\bm \Theta}$ is the diagonal matrix whose $(j,j)$-th element $\theta_j$ satisfies $0 \leq \theta _j \leq 1$. 
In their proof, only the fact that ${\bm H}$ is an orthogonal matrix and   $0 \leq \theta _j \leq 1$. 
On the other hand, when the noise variance is heteroscedastic, that is, the variance is expressed as $s^2_t$ at iteration $t$, we have to consider the following:
$$
{\bm I}_d - {\bm X}^\top _t ( {\bm X}_t {\bm X}^\top_t + a^{-2} {\bm S}_t )^{-1} {\bm X}_t,
$$
where ${\bm S}_t $ is the diagonal matrix whose $(j,j)$-th element is $s^2_j$. 
Also in this case, noting that 
\begin{align*}
{\bm I}_d - {\bm X}^\top _t ( {\bm X}_t {\bm X}^\top_t + a^{-2} {\bm S}_t )^{-1} {\bm X}_t &=
{\bm I}_d - {\bm X}^\top _t ( {\bm S}^{1/2}_t  \{  {\bm S}^{-1/2}_t    {\bm X}_t {\bm X}^\top_t  {\bm S}^{-1/2}_t+ a^{-2} {\bm I}_t \} {\bm S}^{1/2}_t  )^{-1} {\bm X}_t \\
&={\bm I}_d - {\bm X}^\top _t  {\bm S}^{-1/2}_t  (  {\bm S}^{-1/2}_t    {\bm X}_t {\bm X}^\top_t  {\bm S}^{-1/2}_t+ a^{-2} {\bm I}_t  )^{-1}    {\bm S}^{-1/2}_t   {\bm X}_t \\
&= 
{\bm I}_d - \tilde{\bm X}^\top _t ( \tilde{\bm X}_t \tilde{\bm X}^\top_t + a^{-2}  {\bm I}_t )^{-1} \tilde{\bm X}_t,
\end{align*}
using the singular value decomposition $\tilde{\bm X}_t = \tilde{\bm H}^\prime \tilde{\bm \Lambda} \tilde{\bm H}$ we have 
$$
{\bm I}_d - \tilde{\bm X}^\top _t ( \tilde{\bm X}_t \tilde{\bm X}^\top_t + a^{-2}  {\bm I}_t )^{-1} \tilde{\bm X}_t = \tilde{\bm H} \tilde{\bm \Theta} \tilde{\bm H}^\top,
$$
where $\tilde{\bm H}$ is an orthogonal matrix and the $(j,j)$-th element $\tilde{\theta}_j $ of the diagonal matrix $\tilde{\bm \Theta}$ satisfies $0 \leq \tilde{\theta}_j \leq 1$. 
Therefore, also in the heteroscedastic setting, $L_{\sigma}$-data-independent-Lipschitz continuity holds.

\subsection{Proof of Lemma \ref{lem:AF_cal_app}}
Let $ \text{\bf UCB}_t ({\bm x} ) = (u_1,\ldots,u_L) \equiv {\bm u}$ and $\text{\bf LCB}_t (\hat{\Pi}_t ) = \{ (l^{(i)}_1,\ldots,l^{(i)}_L ) \mid 1 \leq i \leq k \} \equiv \mathcal{L}$. 
Here, if ${\bm u} \in \text{Dom} (\mathcal{L})$, then the following holds from the definition of $\text{dist} ({\bm a},B)$:
$$
a^{(\mathcal{X})}_t ({\bm x} ) = \text{dist} ({\bm u},\text{Dom}(\mathcal{L}) ) = \inf_ { {\bm b} \in  \text{Dom}(\mathcal{L})   } d_\infty ({\bm u}, {\bm b} ) = d_\infty ({\bm u},{\bm u} ) =0. 
$$
In addition, since ${\bm u} \in \text{Dom} (\mathcal{L})$, there exists $(l^{(i)}_1,\ldots,l^{(i)}_L ) $ such that 
$u_j \leq l^{(i)}_j$ for any $j \in [L]$. 
Thus, we have $\max \{ u_1- l^{(i)}_1 , \ldots, u_L- l^{(i)}_L \} \leq 0$. 
This implies that 
$$
\tilde{a}_t ({\bm x} ) = \min_ {1 \leq i \leq k }   \max \{ u_1- l^{(i)}_1 ,\ldots, u_L- l^{(i)}_L \}  \leq 0
$$
and $ \max \{  \tilde{a}_t ({\bm x} ) , 0 \} = 0$. 
Therefore, we get $a^{(\mathcal{X})}_t ({\bm x} ) = \max \{  \tilde{a}_t ({\bm x} ) , 0 \} $. 
Next, we consider the case where ${\bm u} \notin \text{Dom} (\mathcal{L})$. 
Let $a^{(\mathcal{X})}_t ({\bm x} ) =\eta$. 
Then, noting that ${\bm u} \notin \text{Dom} (\mathcal{L})$, for any $i \in \{1,\ldots, k \}$, there exists $j \in [L]$ such that $u_j > l^{(i)} _j $. 
This implies that 
$$
\tilde{a}_t ({\bm x} )=\min_ {1 \leq i \leq k }   \max \{ u_1- l^{(i)}_1 , \ldots, u_L- l^{(i)}_L \} \equiv \tilde{\eta } >0
$$
and  $ \max \{  \tilde{a}_t ({\bm x} ) , 0 \} = \tilde{a} _t ({\bm x} ) = \tilde{\eta}$. 
For this $\tilde{\eta}$, there exists $i$ such that 
$$
u_j- l^{(i)}_j  \leq \tilde{\eta}  \quad ^\forall j \in [L].
$$
Hence, we have $\tilde{\bm u} \equiv (u_1- \tilde{\eta} ,\ldots , u_L- \tilde{\eta} ) \in \text{Dom} (\mathcal{L})$ 
because $ u_j- \tilde{\eta}  \leq  l^{(i)}_j$ for any $j \in [L]$. 
Thus, from the definition of $a^{(\mathcal{X})}_t ({\bm x} )$, the following holds:
$$
\eta = a^{(\mathcal{X})}_t ({\bm x} ) = \text{dist} ({\bm u},\text{Dom}(\mathcal{L}) ) = \inf_ { {\bm b} \in  \text{Dom}(\mathcal{L})   } d_\infty ({\bm u}, {\bm b} )  \leq d_\infty ({\bm u},\tilde{\bm u}  ) = \tilde{\eta}.
$$
Here, we assume $\eta < \tilde{\eta}$. 
Then, noting that $\text{Dom}(\mathcal{L})$ is the closed set, there exists $\tilde{\bm l} =( \tilde{l}_1,\ldots, \tilde{l}_L ) \in \text{Dom}(\mathcal{L})$ such that $d_\infty ({\bm u}, \tilde{\bm l} ) = \eta$. 
Therefore, $ \tilde{\bm l}$ can be expressed as $ \tilde{\bm l} = (u_1 -s_1, \ldots,u_L -s_L )$, 
where $0 \leq |s_j| \leq \eta $ and at least one of $s_1,\ldots , s_L$ is $\eta$. 
Thus, since $(u_1 - \eta , \ldots , u_L -\eta ) \leq  \tilde{\bm l}$, noting that $(u_1 - \eta , \ldots , u_L -\eta ) \in \text{Dom}(\mathcal{L})$ there exists 
 $i $ such that 
$$
u_j - \eta \leq l^{(i)}_j \quad ^\forall j \in [L]. 
$$
This implies that $\max \{ u_1 - l^{(i)}_1,  \ldots , u_L - l^{(i)}_L \} \leq \eta $. 
Hence, it follows that 
$$
\tilde{\eta} = \min_ {1 \leq i \leq k }   \max \{ u_1- l^{(i)}_1 , \ldots , u_L- l^{(i)}_L \}  \leq \eta . 
$$
However, this is a contradiction with $\eta < \tilde{\eta}$. 
Consequently, we obtain $a^{(\mathcal{X})}_t ({\bm x} ) =\max \{  \tilde{a}_t ({\bm x} ) , 0 \} $.

\subsection{Proof of Theorem \ref{thm:inference discrepancy_app}}
From the theorem's assumption, the bounding box $\tilde{B}_{t} ({\bm x} )$ is HPBB. 
Therefore, with probability at least  $1-\delta$, the following holds for any $t \geq 0$:
$$
\text{Dom} (\text{\bf LCB}_{t} (\hat{\Pi}_{t}) ) \subset  {\rm Dom} ({\bm F} (\hat{\Pi}_{t})) \subset {\rm Dom} ( Z^\ast ) \subset \text{Dom} (\text{\bf UCB}_{t} (\mathcal{X})).
$$
Hence, using this, noting that the definition of $d_{\infty} (\cdot, \cdot )$, we get  
\begin{align*}
I^{(i)}_{t} = \max_{ {\bm y}  \in Z^\ast    } \min_{{\bm y}^\prime \in   {\rm Par} ({\bm F} (\hat{\Pi}_t)) }   d_\infty ({\bm y},{\bm y}^\prime ) &\leq 
\max_{ {\bm y}  \in \text{Par} (\text{\bf UCB}_{t} (\mathcal{X}) )   } \min_{{\bm y}^\prime \in  \text{Par} (\text{\bf LCB}_{t} (\hat{\Pi}_{t})) }   d_\infty ({\bm y},{\bm y}^\prime ) \\
&=\max_{ {\bm y}  \in \text{Par} (\text{\bf UCB}_{t} (\mathcal{X})  )  } \min_{{\bm y}^\prime \in  \text{Dom} (\text{\bf LCB}_{t} (\hat{\Pi}_{t}) )}   d_\infty ({\bm y},{\bm y}^\prime ) \\
&= \max_{ {\bm x}  \in \mathcal{X} } \min_{{\bm y}^\prime \in  \text{Dom} (\text{\bf LCB}_{t} (\hat{\Pi}_{t}) )}   d_\infty (\text{\bf UCB}_{t} ({\bm x}) ,{\bm y}^\prime ) = \max _{{\bm x} \in \mathcal{X}} a^{(\mathcal{X})}_{t} ({\bm x} ).
\end{align*}
Similarly, 
we get 
\begin{align*}
I^{(ii)}_{t} = \max_{ {\bm y}  \in  {\bm F} (\hat{\Pi}_t)    } \min_{{\bm y}^\prime \in Z^\ast  }   d_\infty ({\bm y},{\bm y}^\prime ) &\leq \max_{ {\bm y}^\prime  \in \text{Par} (\text{\bf UCB}_{t} (\mathcal{X})  )  } \min_{{\bm y} \in  \text{Dom} (\text{\bf LCB}_{t} (\hat{\Pi}_{t}) )}   d_\infty ({\bm y}^\prime,{\bm y} ) \\
&= \max_{ {\bm x}  \in \mathcal{X} } \min_{{\bm y} \in  \text{Dom} (\text{\bf LCB}_{t} (\hat{\Pi}_{t}) )}   d_\infty (\text{\bf UCB}_{t} ({\bm x}) ,{\bm y} ) = \max _{{\bm x} \in \mathcal{X}} a^{(\mathcal{X})}_{t} ({\bm x} ). 
\end{align*}
Thus, we have $I_{t} =\max \{ I^{(i)}_{t},I^{(ii)}_{t} \} \leq \max _{{\bm x} \in \mathcal{X}} a^{(\mathcal{X})}_{t} ({\bm x} )  =  a^{(\mathcal{X})}_{t} ({\bm x}_{t+1} ) $. 
Hence, if $a^{(\mathcal{X})}_{T} ({\bm x}_{T+1} )  \leq \epsilon$, then $ I_T \leq \epsilon$.

\subsection{Proof of Theorem \ref{thm:termination_app}}
From the definition of $a^{(\mathcal{X})}_{t} ({\bm x} )$, ${\bm x} _{t+1}$ and $ {\bm w}_{t+1}$, noting that  $\text{\bf LCB}_{t} ({\bm x}_{t+1}) \in \text{Dom} (\text{\bf LCB}_{t} (\hat{\Pi}_{t}) )$ we get 
\begin{align*}
a^{(\mathcal{X})}_{t} ({\bm x}_{t+1} ) \leq \|  \text{\bf UCB}_t ({\bm x}_{t+1} ) -  \text{\bf LCB}_t ({\bm x}_{t+1} ) \|_\infty & \leq q \left ( \max_{{\bm w} \in \Omega_{{\bm x}_{t+1}}} \sum_{m=1}^{M_f}  2 \tilde{\beta}^{1/2}_{m,t+1} \tilde{\sigma}^{(m)}_{t} ({\bm x}_{t+1},{\bm w}  ) \right )  \\
&= q \left (  \sum_{m=1}^{M_f}  2 \tilde{\beta}^{1/2}_{m,t+1} \tilde{\sigma}^{(m)}_{t} ({\bm x}_{t+1},{\bm w}_{t+1}  ) \right ).
\end{align*}
Let $\hat{t} = \argmin_{ 0 \leq i \leq t}    \sum_{m=1}^{M_f}  2 \tilde{\beta}^{1/2}_{m,t+1} \tilde{\sigma}^{(m)}_{t} ({\bm x}_{t+1},{\bm w}_{t+1}  )$. 
Then, the following inequality holds: 
\begin{align*}
\sum_{m=1}^{M_f}  2 \tilde{\beta}^{1/2}_{m,\hat{t}+1} \tilde{\sigma}^{(m)}_{\hat{t}} ({\bm x}_{\hat{t}+1},{\bm w}_{\hat{t}+1}  ) & \leq \frac{1}{t+1} 
\sum_{i=1}^{t+1}  \sum_{m=1}^{M_f}  2 \tilde{\beta}^{1/2}_{m,i} \tilde{\sigma}^{(m)}_{i-1} ({\bm x}_{i},{\bm w}_i  ) \\
&\leq \frac{1}{t+1}  \sqrt{(t+1)  \sum_{i=1 }^{t+1} \sum_{m=1}^{M_f} 4M_f \tilde{\beta}_{m,i} \tilde{\sigma}^{(m)2}_{i-1}  ({\bm x}_i , {\bm w} _i ) } \\
&\leq \frac{1}{t+1}  \sqrt{(t+1)  \sum_{m=1 }^{M_f} 4M_f \tilde{\beta}_{m,t+1}   \sum_{i=1}^{t+1} \tilde{\sigma}^{(m)2}_{i-1}  ({\bm x}_i , {\bm w} _i ) } \\
&\leq \frac{1}{t+1}  \sqrt{(t+1)  \sum_{m=1 }^{M_f} 4M_f \tilde{\beta}_{m,t+1}  \frac{2}{\log (1+ \lambda^{-1}_m \underline{\tau}^{-2}) } \tilde{\kappa}^{(m)}_{t+1} } =
\sqrt{
\frac{ \sum_{m=1} ^{M_f}   \tilde{ C}_m \tilde{\beta}_{m,t+1} \tilde{\kappa}^{(m)}_{t+1} }{t+1}
},
\end{align*}
where the second inequality is derived by Cauchy-Schwarz inequality and $(a_1 + \cdots + a_{M_f})^2 \leq M_f (a^2_1 + \cdots + a^2_{M_f})$, 
the third inequality is derived by monotonicity of $\tilde{\beta}_{m,t}$, 
and the fourth inequality is derived by the definition of the maximum information gain, $s^2 \leq (\varsigma^{-2}/\log(1+\varsigma^{-2})) 
\log (1 + s^2) $ for $s^2 \in [0,\varsigma^{-2} ] $, and $\tau^{-2}_{ {\bm x}_i,{\bm w}_i ,m} \tilde{\sigma}^{(m)2}_{i-1} ({\bm x}_i,{\bm w}_i) \leq \tau^{-2}_{ {\bm x}_i,{\bm w}_i ,m} \lambda^{-1}_m $. 
Therefore, we obtain 
$$
\max_{ {\bm x} \in \mathcal{X} }  a^{(\mathcal{X})}_{ \hat{t} } ({\bm x}) =
a^{(\mathcal{X})}_{ \hat{t} }  ( {\bm x}_{\hat{t}+1}  ) \leq q \left ( \sum_{m=1}^{M_f}  2 \tilde{\beta}^{1/2}_{m,\hat{t}+1} \tilde{\sigma}^{(m)}_{\hat{t}} ({\bm x}_{\hat{t}+1},{\bm w}_{\hat{t}+1}  )  \right ) \leq q \left ( \sqrt{
\frac{ \sum_{m=1} ^{M_f}    \tilde{C}_m \tilde{\beta}_{m,t+1} \tilde{\kappa}^{(m)}_{t+1} }{t+1}
}  \right ) = q(s_t).
$$
Thus, for some $T \geq 0$ satisfying $q(s_T ) \leq \epsilon$, there exists $\hat{T} \leq T$ such that 
$a^{(\mathcal{X})}_{ \hat{T} }  ( {\bm x}_{\hat{T}+1}  ) \leq q(s_T) \leq \epsilon$. 
Noting that $ 0 \leq \hat{T} \leq T$, the algorithm terminates after at most $T$ iterations.

\subsection{Proof of Theorem \ref{thm:q_app}}
From the definition of $q(a)$, since $q^{(m,l)} (a) $ is a strictly increasing function satisfying  $q^{(m,l)} (0) =0$, 
$q(a)$ is a strictly increasing function and satisfies $q(0) =0$.
Furthermore, noting that 
 $$
\max_{ {\bm w} \in \Omega_{{\bm x}_{t+1}}} 2 \tilde{\beta}^{1/2}_{m,t+1} \tilde{\sigma}^{(m)}_{t} ({\bm x}_{t+1} , {\bm w}  ) \leq 
\max_{ {\bm w} \in \Omega_{{\bm x}_{t+1}}} \sum_{m=1}^{M_f} 2 \tilde{\beta}^{1/2}_{m,t+1} \tilde{\sigma}^{(m)}_{t} ({\bm x}_{t+1} , {\bm w}  ),
$$  since $q^{(m,l)} (a) $ is a strictly increasing function, we get 
\begin{align*}
\|  \text{\bf UCB}_t ({\bm x}_{t+1} ) -  \text{\bf LCB}_t ({\bm x}_{t+1} ) \|_\infty &=
\max_{ m \in [M_f], l \in [L_m] }    | {\rm ucb}^{(m,l)}_{t} ({\bm x}_{t+1} ) -  {\rm lcb}^{(m,l)}_{t} ({\bm x}_{t+1} )  | \\
& \leq 
\max_{ m \in [M_f], l \in [L_m] }  q^{(m,l)}  \left (  \max_{ {\bm w} \in \Omega_{{\bm x}_{t+1}}} 2 \tilde{\beta}^{1/2}_{m,t+1} \tilde{\sigma}^{(m)}_{t} ({\bm x}_{t+1} , {\bm w}  ) \right ) \\
& \leq 
\max_{ m \in [M_f] , l \in [L_m] }  q^{(m,l)}  \left (  \max_{ {\bm w} \in \Omega_{{\bm x}_{t+1}}} \sum_{m=1}^{M_f} 2 \tilde{\beta}^{1/2}_{m,t+1} \tilde{\sigma}^{(m)}_{t} ({\bm x}_{t+1} , {\bm w}  ) \right ) \\
& 
= q \left (  \max_{ {\bm w} \in \Omega_{{\bm x}_{t+1}}} \sum_{m=1}^{M_f} 2 \tilde{\beta}^{1/2}_{m,t+1} \tilde{\sigma}^{(m)}_{t} ({\bm x}_{t+1} , {\bm w}  ) \right ) .
\end{align*}

\subsection{Proof of Theorem \ref{thm:inference discrepancy with errors_app}}
Let $r $ be a number. 
For any vector ${\bm a} = (a_1 , \ldots , a_s) $ and subset $B \subset \mathbb{R}^s$, 
we define $r+ {\bm a}  \equiv (r+a_1, \ldots , r+ a_s )$ and $r+B \equiv \{ r+  {\bm b} \mid {\bm b} \in B  \}$. 
Then, from the theorem's assumption, with probability at least $1-\delta$, the following holds for any $t \geq 0$:
$$
\text{Dom} (\text{\bf LCB}_{t} (\hat{\Pi}_{t}) -\epsilon_{{\rm lcb}}) \subset  {\rm Dom} ({\bm F} (\hat{\Pi}_{t})) \subset {\rm Dom} (Z^\ast) \subset \text{Dom} (\text{\bf UCB}_{t} (\mathcal{X})+\epsilon_{{\rm ucb}}).
$$
Hence, using this, noting that the definition of $d_{\infty} (\cdot, \cdot )$, we get  
\begin{align*}
I^{(i)}_{t} = \max_{ {\bm y}  \in Z^\ast    } \min_{{\bm y}^\prime \in   {\rm Par} ({\bm F} (\hat{\Pi}_t)) }   d_\infty ({\bm y},{\bm y}^\prime ) &\leq 
\max_{ {\bm y}  \in \text{Par} (\text{\bf UCB}_{t} (\mathcal{X})  +\epsilon_{{\rm ucb}}  )   } \min_{{\bm y}^\prime \in  \text{Par} (\text{\bf LCB}_{t} (\hat{\Pi}_{t})-\epsilon_{{\rm lcb}}) }   d_\infty ({\bm y},{\bm y}^\prime ) \\
&=\max_{ {\bm y}  \in \text{Par} (\text{\bf UCB}_{t} (\mathcal{X}) +\epsilon_{{\rm ucb}} )  } \min_{{\bm y}^\prime \in  \text{Dom} (\text{\bf LCB}_{t} (\hat{\Pi}_{t}) -\epsilon_{{\rm lcb}})}   d_\infty ({\bm y},{\bm y}^\prime ) \\
&= \max_{ {\bm x}  \in \mathcal{X} } \min_{{\bm y}^\prime \in  \text{Dom} (\text{\bf LCB}_{t} (\hat{\Pi}_{t}) -\epsilon_{{\rm lcb}})}   d_\infty (\text{\bf UCB}_{t} ({\bm x})+\epsilon_{{\rm ucb}} ,{\bm y}^\prime ) \\
&\leq 
\epsilon_{{\rm ucb}} + 
\max_{ {\bm x}  \in \mathcal{X} } \min_{{\bm y}^\prime \in  \text{Dom} (\text{\bf LCB}_{t} (\hat{\Pi}_{t}) -\epsilon_{{\rm lcb}})}   d_\infty (\text{\bf UCB}_{t} ({\bm x}) ,{\bm y}^\prime ) \\ 
&\leq 
\epsilon_{{\rm ucb}} + \epsilon_{{\rm lcb}}+
\max_{ {\bm x}  \in \mathcal{X} } \min_{{\bm y}^\prime \in  \text{Dom} (\text{\bf LCB}_{t} (\hat{\Pi}_{t}) )}   d_\infty (\text{\bf UCB}_{t} ({\bm x}) ,{\bm y}^\prime ) \\
&=\epsilon_{{\rm ucb}} + \epsilon_{{\rm lcb}} + \max _{{\bm x} \in \mathcal{X}} a^{(\mathcal{X})}_{t} ({\bm x} ) \\
&= \epsilon_{{\rm ucb}} + \epsilon_{{\rm lcb}} +  a^{(\mathcal{X})}_{t} ({\bm x}_{t+1} ) +    \max _{{\bm x} \in \mathcal{X}} a^{(\mathcal{X})}_{t} ({\bm x} ) -a^{(\mathcal{X})}_{t} ({\bm x}_{t+1} )  \\
&\leq \epsilon_{{\rm ucb}} + \epsilon_{{\rm lcb}} +  \epsilon_{\mathcal{X}}+a^{(\mathcal{X})}_{t} ({\bm x}_{t+1} ).
\end{align*}
Similarly, 
we get 
\begin{align*}
I^{(ii)}_{t} = \max_{ {\bm y}  \in  {\bm F} (\hat{\Pi}_t)    } \min_{{\bm y}^\prime \in Z^\ast  }   d_\infty ({\bm y},{\bm y}^\prime ) &\leq \max_{ {\bm y}^\prime  \in \text{Par} (\text{\bf UCB}_{t} (\mathcal{X}) +\epsilon_{{\rm ucb}} )  } \min_{{\bm y} \in  \text{Dom} (\text{\bf LCB}_{t} (\hat{\Pi}_{t})-\epsilon_{{\rm lcb}} )}   d_\infty ({\bm y}^\prime,{\bm y} ) \\
&\leq \epsilon_{{\rm ucb}} + \epsilon_{{\rm lcb}} +  \epsilon_{\mathcal{X}}+a^{(\mathcal{X})}_{t} ({\bm x}_{t+1} ).
\end{align*}
Thus, we have $I_{t} =\max \{ I^{(i)}_{t},I^{(ii)}_{t} \} \leq \epsilon_{{\rm ucb}} + \epsilon_{{\rm lcb}} +  \epsilon_{\mathcal{X}}+a^{(\mathcal{X})}_{t} ({\bm x}_{t+1} ) $. 
Hence, if $a^{(\mathcal{X})}_{T} ({\bm x}_{T+1} )  \leq \epsilon$, then $ I_T \leq \epsilon +  \epsilon_{{\rm ucb}} + \epsilon_{{\rm lcb}} +  \epsilon_{\mathcal{X}}$.

\subsection{Proof of Theorem \ref{thm:termination with errors_app}}
From the definition of $a^{(\mathcal{X})}_{t} ({\bm x} )$, ${\bm x} _{t+1}$ and $ {\bm w}_{t+1}$, noting that  $-\epsilon_{{\rm PF}}+\text{\bf LCB}_{t} ({\bm x}_{t+1}) \in \text{Dom} (\text{\bf LCB}_{t} (\hat{\Pi}_{t}) )$ we get 
\begin{align*}
a^{(\mathcal{X})}_{t} ({\bm x}_{t+1} ) \leq \|  \text{\bf UCB}_t ({\bm x}_{t+1} ) -  (\text{\bf LCB}_t ({\bm x}_{t+1} ) -\epsilon_{{\rm PF}}) \|_\infty &\leq \epsilon_{{\rm PF}}+ \|  \text{\bf UCB}_t ({\bm x}_{t+1} ) -  \text{\bf LCB}_t ({\bm x}_{t+1} ) \|_\infty \\
& \leq \epsilon_{{\rm PF}}+ q \left ( \max_{{\bm w} \in \Omega_{{\bm x}_{t+1}}} \sum_{m=1}^{M_f}  2 \tilde{\beta}^{1/2}_{m,t+1} \tilde{\sigma}^{(m)}_{t} ({\bm x}_{t+1},{\bm w}  ) \right )  \\
&\leq \epsilon_{{\rm PF}}+ q \left ( \epsilon_{\Omega}+ \sum_{m=1}^{M_f}  2 \tilde{\beta}^{1/2}_{m,t+1} \tilde{\sigma}^{(m)}_{t} ({\bm x}_{t+1},{\bm w}_{t+1}  ) \right ).
\end{align*}
Thus, by letting $\hat{t} = \argmin_{ 0 \leq i \leq t}    \sum_{m=1}^{M_f}  2 \tilde{\beta}^{1/2}_{m,t+1} \tilde{\sigma}^{(m)}_{t} ({\bm x}_{t+1},{\bm w}_{t+1}  )$, using the same argument as in the proof of Theorem \ref{thm:termination_app}, we have the desired result.

\subsection{Proof of Theorem \ref{thm:termination_app_uncontrollable}}
From the definition of $a^{(\mathcal{X})}_{t} ({\bm x} )$, ${\bm x} _{t+1}$ and $ {\bm w}_{t+1}$, noting that  $\text{\bf LCB}_{t} ({\bm x}_{t+1}) \in \text{Dom} (\text{\bf LCB}_{t} (\hat{\Pi}_{t}) )$ we get 
\begin{align*}
a^{(\mathcal{X})}_{t} ({\bm x}_{t+1} ) \leq \|  \text{\bf UCB}_t ({\bm x}_{t+1} ) -  \text{\bf LCB}_t ({\bm x}_{t+1} ) \|_\infty & \leq q \left ( \max_{{\bm w} \in \Omega_{{\bm x}_{t+1}}} \sum_{m=1}^{M_f}  2 \tilde{\beta}^{1/2}_{m,t+1} \tilde{\sigma}^{(m)}_{t} ({\bm x}_{t+1},{\bm w}  ) \right )  \\
&= q \left (  \sum_{m=1}^{M_f}  2 \tilde{\beta}^{1/2}_{m,t+1} \tilde{\sigma}^{(m)}_{t} ({\bm x}_{t+1},{\bm w}^\ast_{t+1}  ) \right ).
\end{align*}
Let $\mathcal{S} (\Omega_{{\bm x}_{t+1} };\zeta_{t+1} ) $ be a maximal $\zeta_{t+1}$-separated subset of $\Omega_{{\bm x}_{t+1} }$. 
Then, from the definition of $\mathcal{S} (\Omega_{{\bm x}_{t+1} };\zeta_{t+1} ) $, there exists a point $\check{\bm w} \in \mathcal{S} (\Omega_{{\bm x}_{t+1} };\zeta_{t+1} ) $ such that $ \|  {\bm w}^\ast_{t+1} -\check{\bm w} \|_1 \leq \zeta_{t+1}$. 
Hence, from the $L_{\sigma}$-data-independent Lipschitz continuity, we obtain 
\begin{align*}
 \sum_{m=1}^{M_f}  2 \tilde{\beta}^{1/2}_{m,t+1} \tilde{\sigma}^{(m)}_{t} ({\bm x}_{t+1},{\bm w}^\ast_{t+1}  )  
&=
 \sum_{m=1}^{M_f}  2 \tilde{\beta}^{1/2}_{m,t+1} \tilde{\sigma}^{(m)}_{t} ({\bm x}_{t+1},\check{\bm w}  ) +
 \sum_{m=1}^{M_f}  2 \tilde{\beta}^{1/2}_{m,t+1} \{ \tilde{\sigma}^{(m)}_{t} ({\bm x}_{t+1},{\bm w}^\ast_{t+1}  ) - \tilde{\sigma}^{(m)}_{t} ({\bm x}_{t+1},\check{\bm w}  )   \} \\
&\leq \sum_{m=1}^{M_f}  2 \tilde{\beta}^{1/2}_{m,t+1} L_{\sigma} \zeta_{t+1} + \sum_{m=1}^{M_f}  2 \tilde{\beta}^{1/2}_{m,t+1} \tilde{\sigma}^{(m)}_{t} ({\bm x}_{t+1}, \check{\bm w})  \\
&\leq  2M_f \tilde{\beta}^{1/2}_{t+1} L_{\sigma} \zeta_{t+1} + \sum_{m=1}^{M_f}  2 \tilde{\beta}^{1/2}_{m,t+1} \tilde{\sigma}^{(m)}_{t} ({\bm x}_{t+1},\check{\bm w}  )  .
\end{align*}
In addition, we get 
\begin{align*}
\sum_{m=1}^{M_f}  2 \tilde{\beta}^{1/2}_{m,t+1} \tilde{\sigma}^{(m)}_{t} ({\bm x}_{t+1},\check{\bm w}  )
&\leq \sum_{ \check{\bm w}  \in \mathcal{S} (\Omega_{{\bm x}_{t+1} };\zeta_{t+1} )   }  \sum_{m=1}^{M_f}  2 \tilde{\beta}^{1/2}_{m,t+1} \tilde{\sigma}^{(m)}_{t} ({\bm x}_{t+1},\check{\bm w}  ) 
\\
&\leq 
\underline{p_{\zeta_{t+1}}} ^{-1} \sum_{ \check{\bm w}  \in \mathcal{S} (\Omega_{{\bm x}_{t+1} };\zeta_{t+1} )   }  \sum_{m=1}^{M_f}  2 \tilde{\beta}^{1/2}_{m,t+1} \tilde{\sigma}^{(m)}_{t} ({\bm x}_{t+1},\check{\bm w}  ) \mathbb{P}_{ P_{\bm w} ({\bm x}_{t+1})  } [ {\bm w} \in {\rm Nei} ( \check{\bm w};\zeta_{t+1}/2 )  ]  \\
&\leq 2 \tilde{\beta}^{1/2}_{t+1} \underline{p_{\zeta_{t+1}}} ^{-1} \mathbb{E}_{ P_{\bm w} ({\bm x}_{t+1})  } \left [
\sum_{ \check{\bm w}  \in \mathcal{S} (\Omega_{{\bm x}_{t+1} };\zeta_{t+1} )   }  \sum_{m=1}^{M_f}  \tilde{\sigma}^{(m)}_{t} ({\bm x}_{t+1},\check{\bm w}  ) \1 [ {\bm w} \in {\rm Nei} ( \check{\bm w};\zeta_{t+1}/2 )  ] 
\right ] \\
&\equiv 2 \tilde{\beta}^{1/2}_{t+1} \underline{p_{\zeta_{t+1}}} ^{-1} \mathbb{E}_{ P_{\bm w} ({\bm x}_{t+1})  }  [S ({\bm x}_{t+1},{\bm w}) ] ,
\end{align*}
where $\1 [ \cdot ]$ represents the indicator function. 
Thus, we have 
$$
\sum_{m=1}^{M_f}  2 \tilde{\beta}^{1/2}_{m,t+1} \tilde{\sigma}^{(m)}_{t} ({\bm x}_{t+1},{\bm w}^\ast_{t+1}  )  
\leq 
2M_f \tilde{\beta}^{1/2}_{t+1} L_{\sigma} \zeta_{t+1} + 2 \tilde{\beta}^{1/2}_{t+1} \underline{p_{\zeta_{t+1}}} ^{-1} \mathbb{E}_{ P_{\bm w} ({\bm x}_{t+1})  }  [S ({\bm x}_{t+1},{\bm w}) ].
$$
Therefore, we get 
\begin{align*}
\sum_{i=0}^t \sum_{m=1}^{M_f}  2 \tilde{\beta}^{1/2}_{m,i+1} \tilde{\sigma}^{(m)}_{i} ({\bm x}_{i+1},{\bm w}^\ast_{i+1}  )  
&\leq 
\sum_{i=0}^t 2M_f \tilde{\beta}^{1/2}_{i+1} L_{\sigma} \zeta_{i+1} + \sum_{i=0}^t 2 \tilde{\beta}^{1/2}_{i+1} \underline{p_{\zeta_{i+1}}} ^{-1} \mathbb{E}_{ P_{\bm w} ({\bm x}_{i+1})  }  [S ({\bm x}_{i+1},{\bm w}) ] \\
&\leq 
2M_f L_{\sigma}  \tilde{\beta}^{1/2}_{t+1} 
\sum_{i=0}^t  \zeta_{i+1} + 2 \tilde{\beta}^{1/2}_{t+1} \underline{\tilde{p}_{\zeta_{t+1}}} ^{-1} \sum_{i=0}^t  \mathbb{E}_{ P_{\bm w} ({\bm x}_{i+1})  }  [S ({\bm x}_{i+1},{\bm w}) ].
\end{align*}
Here, $S ({\bm x}_{i+1},{\bm w})$ is the non-negative random variable satisfying $S ({\bm x}_{i+1},{\bm w}) \leq M_f \max \{ 1, \lambda^{-1}_1 , \ldots , \lambda^{-1}_m \} =J$. 
Hence, from Lemma 3 in \cite{pmlr-v75-kirschner18a}, with probability at least $1-\delta$, the following holds for any $i \geq 0$: 
$$
\sum_{i=0}^t  \mathbb{E}_{ P_{\bm w} ({\bm x}_{i+1})  }  [S ({\bm x}_{i+1},{\bm w}) ] 
\leq 
4J \log \frac{1}{\delta} + 8J \log (4J) +1 +
2 \sum_{i=0}^t S ({\bm x}_{i+1},{\bm w}_{i+1}) 
\leq 8J \log \frac{8J}{\delta} + 2 \sum_{i=0}^t S ({\bm x}_{i+1},{\bm w}_{i+1}) .
$$
Furthermore, from the definition of $S({\bm x}_{i+1},{\bm w}_{i+1}  )$, we have 
\begin{align*}
S ({\bm x}_{i+1},{\bm w}_{i+1}) 
=
 \sum_{m=1}^{M_f} \sum_{ \check{\bm w}  \in \mathcal{S} (\Omega_{{\bm x}_{i+1} };\zeta_{i+1} )   }  \tilde{\sigma}^{(m)}_{i} ({\bm x}_{i+1},\check{\bm w}  ) \1 [ {\bm w}_{i+1} \in {\rm Nei} ( \check{\bm w};\zeta_{i+1}/2 )  ] .
\end{align*}
Noting that $\check{\bm w}_1 \neq \check{\bm w}_2 \Rightarrow {\rm Nei} ( \check{\bm w}_1;\zeta_{i+1}/2 )  \cap  {\rm Nei} ( \check{\bm w}_2;\zeta_{i+1}/2 ) = \emptyset$, if there exists $\check{\bm w} \in  \mathcal{S} (\Omega_{{\bm x}_{i+1} };\zeta_{i+1} ) $ 
such that $ {\bm w}_{i+1} \in {\rm Nei} ( \check{\bm w};\zeta_{i+1}/2 ) $, then we obtain 
$$
\tilde{\sigma}^{(m)}_{i} ({\bm x}_{i+1},\check{\bm w}  ) =
\tilde{\sigma}^{(m)}_{i} ({\bm x}_{i+1},{\bm w}_{i+1}  )
+
\tilde{\sigma}^{(m)}_{i} ({\bm x}_{i+1},\check{\bm w}  )
-
\tilde{\sigma}^{(m)}_{i} ({\bm x}_{i+1},{\bm w}_{i+1}  ) \leq \tilde{\sigma}^{(m)}_{i} ({\bm x}_{i+1},{\bm w}_{i+1}  ) + L_{\sigma} \zeta_{i+1}/2. 
$$
Similarly, if $ {\bm w}_{i+1} \notin {\rm Nei} ( \check{\bm w};\zeta_{i+1}/2 ) $ for any $\check{\bm w} \in  \mathcal{S} (\Omega_{{\bm x}_{i+1} };\zeta_{i+1} ) $, the we get 
$$
\tilde{\sigma}^{(m)}_{i} ({\bm x}_{i+1},\check{\bm w}  ) =0 \leq 
\tilde{\sigma}^{(m)}_{i} ({\bm x}_{i+1},{\bm w}_{i+1}  ) + L_{\sigma} \zeta_{i+1}/2. 
$$
Therefore, we have 
$$
 2 \sum_{i=0}^t S ({\bm x}_{i+1},{\bm w}_{i+1}) 
\leq M_f L_{\sigma} \sum_{i=0}^t \zeta_{i+1}   +   2 \sum_{i=0}^t  \sum_{m=1}^{M_f} \tilde{\sigma}^{(m)}_{i} ({\bm x}_{i+1},{\bm w}_{i+1}  ).
$$
By combining previous results, we obtain 
\begin{align*}
&\sum_{i=0}^t \sum_{m=1}^{M_f}  2 \tilde{\beta}^{1/2}_{m,i+1} \tilde{\sigma}^{(m)}_{i} ({\bm x}_{i+1},{\bm w}^\ast_{i+1}  )   \\
&\leq 
2M_f L_{\sigma}  \tilde{\beta}^{1/2}_{t+1} 
\sum_{i=0}^t  \zeta_{i+1} + 2 \tilde{\beta}^{1/2}_{t+1} \underline{\tilde{p}_{\zeta_{t+1}}} ^{-1} \left ( 8J \log \frac{8J}{\delta} + 2 \sum_{i=0}^t S ({\bm x}_{i+1},{\bm w}_{i+1})    \right ) \\
&\leq 2M_f L_{\sigma}  \tilde{\beta}^{1/2}_{t+1} 
\sum_{i=0}^t  \zeta_{i+1} + 2 \tilde{\beta}^{1/2}_{t+1} \underline{\tilde{p}_{\zeta_{t+1}}} ^{-1} \left ( 8J \log \frac{8J}{\delta} + M_f L_{\sigma} \sum_{i=0}^t \zeta_{i+1}   +   2 \sum_{i=0}^t  \sum_{m=1}^{M_f} \tilde{\sigma}^{(m)}_{i} ({\bm x}_{i+1},{\bm w}_{i+1}  )  \right )  \\
&=
2 M_f L_{\sigma} \tilde{\beta}^{1/2}_{t+1} (1 + \underline{\tilde{p}_{\zeta_{t+1}}} ^{-1} ) \sum_{i=0}^t \zeta_{i+1} 
+      16J \log \frac{8J}{\delta}  \tilde{\beta}^{1/2}_{t+1} \underline{\tilde{p}_{\zeta_{t+1}}} ^{-1}  
+  2 \underline{\tilde{p}_{\zeta_{t+1}}} ^{-1} \sum_{i=0}^t  \sum_{m=1}^{M_f} 2 \beta^{1/2}_{t+1} \tilde{\sigma}^{(m)}_{i} ({\bm x}_{i+1},{\bm w}_{i+1}  ). 
\end{align*}
Finally, 
let $\hat{t} = \argmin_{ 0 \leq i \leq t}    \sum_{m=1}^{M_f}  2 \tilde{\beta}^{1/2}_{m,i+1} \tilde{\sigma}^{(m)}_{i} ({\bm x}_{i+1},{\bm w}^\ast_{i+1}  )$. 
Then, the following inequality holds: 
\begin{align*}
&\sum_{m=1}^{M_f}  2 \tilde{\beta}^{1/2}_{m,\hat{t}+1} \tilde{\sigma}^{(m)}_{\hat{t}} ({\bm x}_{\hat{t}+1},{\bm w}^\ast_{\hat{t}+1}  ) \\
& \leq 
\frac{2 M_f L_{\sigma} \tilde{\beta}^{1/2}_{t+1} (1 + \underline{\tilde{p}_{\zeta_{t+1}}} ^{-1} )}{t+1} \sum_{i=0}^t \zeta_{i+1} 
+\frac{ 16J \log \frac{8J}{\delta}  \tilde{\beta}^{1/2}_{t+1} \underline{\tilde{p}_{\zeta_{t+1}}} ^{-1}  }{t+1} +
\frac{ 2 \underline{\tilde{p}_{\zeta_{t+1}}} ^{-1}}{t+1} 
\sum_{i=0}^{t}  \sum_{m=1}^{M_f}  2 \tilde{\beta}^{1/2}_{i+1} \tilde{\sigma}^{(m)}_{i} ({\bm x}_{i+1},{\bm w}_{i+1}  ) .
\end{align*}
In addition, by using the same argument as in the proof of Theorem \ref{thm:termination_app}, we get
\begin{align*}
&\sum_{m=1}^{M_f}  2 \tilde{\beta}^{1/2}_{m,\hat{t}+1} \tilde{\sigma}^{(m)}_{\hat{t}} ({\bm x}_{\hat{t}+1},{\bm w}^\ast_{\hat{t}+1}  ) \\
& \leq 
\frac{2 M_f L_{\sigma} \tilde{\beta}^{1/2}_{t+1} (1 + \underline{\tilde{p}_{\zeta_{t+1}}} ^{-1} )}{t+1} \sum_{i=0}^t \zeta_{i+1} 
+\frac{ 16J \log \frac{8J}{\delta}  \tilde{\beta}^{1/2}_{t+1} \underline{\tilde{p}_{\zeta_{t+1}}} ^{-1}  }{t+1} +
\sqrt{
4 \underline{\tilde{p}_{\zeta_{t+1}}} ^{-2}
\frac{ \sum_{m=1} ^{M_f}   \tilde{ C}_m \tilde{\beta}_{t+1} \tilde{\kappa}_{t+1} }{t+1}
} \\
& = 
\frac{2 M_f L_{\sigma} \tilde{\beta}^{1/2}_{t+1} (1 + \underline{\tilde{p}_{\zeta_{t+1}}} ^{-1} )}{t+1} \sum_{i=0}^t \zeta_{i+1} 
+\frac{ 16J \log \frac{8J}{\delta}  \tilde{\beta}^{1/2}_{t+1} \underline{\tilde{p}_{\zeta_{t+1}}} ^{-1}  }{t+1} +
\sqrt{
 \underline{\tilde{p}_{\zeta_{t+1}}} ^{-2} \tilde{\beta}_{t+1} \tilde{\kappa}_{t+1}
\frac{ \sum_{m=1} ^{M_f}   \hat{ C}_m  }{t+1}
} \\ 
& \leq 
\frac{2 M_f L_{\sigma} \tilde{\beta}^{1/2}_{t+1} (1 + \underline{\tilde{p}_{\zeta_{t+1}}} ^{-1} )}{t+1} \sum_{i=0}^t \zeta_{i+1} 
+\frac{ 16J \log \frac{8J}{\delta}  \tilde{\beta}^{1/2}_{t+1} \underline{\tilde{p}_{\zeta_{t+1}}} ^{-1}  }{t+1} +
\sqrt{
 \underline{\tilde{p}_{\zeta_{t+1}}} ^{-2} \tilde{\beta}_{t+1} \tilde{\kappa}_{t+1}
\frac{ M_f \max \{  \hat{ C}_1, \ldots, \hat{C}_{M_f}  \}  }{t+1}
} \\
& = 
\frac{2 M_f L_{\sigma} \tilde{\beta}^{1/2}_{t+1} (1 + \underline{\tilde{p}_{\zeta_{t+1}}} ^{-1} )}{t+1} \sum_{i=0}^t \zeta_{i+1} 
+\frac{ 16J \log \frac{8J}{\delta}  \tilde{\beta}^{1/2}_{t+1} \underline{\tilde{p}_{\zeta_{t+1}}} ^{-1}  }{t+1} +
\sqrt{
\frac{  \hat{C}  \underline{\tilde{p}_{\zeta_{t+1}}} ^{-2} \tilde{\beta}_{t+1} \tilde{\kappa}_{t+1}  }{t+1}
} \\
&\equiv \hat{s}_t.
\end{align*}
Therefore, we obtain 
$$
a^{(\mathcal{X})}_{ \hat{t} }  ( {\bm x}_{\hat{t}+1}  ) \leq  q(\hat{s}_t).
$$
Thus, for some $T \geq 0$ satisfying $q(\hat{s}_T ) \leq \epsilon$, there exists $\hat{T} \leq T$ such that 
$a^{(\mathcal{X})}_{ \hat{T} }  ( {\bm x}_{\hat{T}+1}  ) \leq q(\hat{s}_T) \leq \epsilon$. 
Noting that $ 0 \leq \hat{T} \leq T$, the algorithm terminates after at most $T$ iterations.

\subsection{Proof of Theorem \ref{thm:inference discrepancy with errors_app_uncontrollable}}
The proof of Theorem \ref{thm:inference discrepancy with errors_app_uncontrollable} is same as in the proof of Theorem 
\ref{thm:inference discrepancy with errors_app}.

\subsection{Proof of Theorem \ref{thm:termination with errors_app_uncontrollable}}
From the definition of $a^{(\mathcal{X})}_{t} ({\bm x} )$ and ${\bm x} _{t+1}$, noting that  $-\epsilon_{{\rm PF}}+\text{\bf LCB}_{t} ({\bm x}_{t+1}) \in \text{Dom} (\text{\bf LCB}_{t} (\hat{\Pi}_{t}) )$ we get 
\begin{align*}
a^{(\mathcal{X})}_{t} ({\bm x}_{t+1} ) \leq \|  \text{\bf UCB}_t ({\bm x}_{t+1} ) -  (\text{\bf LCB}_t ({\bm x}_{t+1} ) -\epsilon_{{\rm PF}}) \|_\infty &\leq \epsilon_{{\rm PF}}+ \|  \text{\bf UCB}_t ({\bm x}_{t+1} ) -  \text{\bf LCB}_t ({\bm x}_{t+1} ) \|_\infty \\
& \leq \epsilon_{{\rm PF}}+ q \left ( \max_{{\bm w} \in \Omega_{{\bm x}_{t+1}}} \sum_{m=1}^{M_f}  2 \tilde{\beta}^{1/2}_{m,t+1} \tilde{\sigma}^{(m)}_{t} ({\bm x}_{t+1},{\bm w}  ) \right )  \\ 
&=
\epsilon_{{\rm PF}}+ q \left (  \sum_{m=1}^{M_f}  2 \tilde{\beta}^{1/2}_{m,t+1} \tilde{\sigma}^{(m)}_{t} ({\bm x}_{t+1},{\bm w}^\ast_{t+1}  ) \right ) .
\end{align*}
Thus, by letting $\hat{t} = \argmin_{ 0 \leq i \leq t}    \sum_{m=1}^{M_f}  2 \tilde{\beta}^{1/2}_{m,i+1} \tilde{\sigma}^{(m)}_{i} ({\bm x}_{i+1},{\bm w}^\ast_{i+1}  )$, using the same argument as in the proof of Theorem \ref{thm:termination_app_uncontrollable}, we have the desired result.

\section{Experimental Details and Additional Experiments}\label{app:exp_details}
In this section, we give experimental details and additional experiments. 
All experiments were performed using {\tt R} software version 3.6.3. 
For all experiments except for additional experiments, we set the tradeoff parameter $\beta^{1/2}_{m,t} $ to 3. 

\subsection{Details of Synthetic Function Experiments without Input Uncertainty}
In the synthetic function experiments without IU, the input space $\mathcal{X} $ was a set of grid points divided into $[-5,5] \times [-5,5]$ equally spaced at $50 \times 50$. 
For black-box functions, we used Booth, Matyas, Himmelblau's and McCormic benchmark functions. 
We standardized these functions and further multiplied by minus one. 
The functional forms we actually used in our experiments are given as follows:
\begin{itemize}
\item Booth function:
$$
f(x_1,x_2)=\frac{  -(x_1+2x_2-7)^2-(2x_1+x_2-5)^2+157.35               }{\sqrt{28896.11}}.
$$
\item Matyas function:
$$
f(x_1,x_2)=\frac{  -0.26(x_1^2+x_2^2)+0.48x_1x_2+4.3342              }{\sqrt{23.52052}}.
$$
\item Himmelblau's function:
$$
f(x_1,x_2)=\frac{  -(x_1^2+x_2-11)^2+(x_1+x_2^2-7)^2+136.71              }{\sqrt{12503.63}}.
$$
\item McCormic function:
$$
f(x_1,x_2)=\frac{  - \sin (x_1+x_2) -(x_1-x_2)^2 +1.5x_1 -2.5x_2-1           17.67              }{\sqrt{460.573}}.
$$
\end{itemize}
We performed the following two cases: (i) Two-objective Pareto optimization problem using first two benchmark functions, (ii) four-objective Pareto optimization problem using all benchmark functions.    
For each black-box function, we used the independent GP model $\mathcal{G} \mathcal{P} (0,k({\bm x},{\bm x}^\prime ))$, where the kernel function $k({\bm x},{\bm x}^\prime )$ is given by 
$$
k({\bm x},{\bm x}^\prime ) =  2 \exp \left (  
-\frac{  \|  {\bm x}  -{\bm x}^\prime \|^2_2   }{2}
\right ).
$$
We used the zero-mean independent Gaussian noise with variance $10^{-6} $ for all black-box functions. 
As evaluation indicators, we used the simple Pareto hypervolume (PHV) regret, which is a commonly used indicator in the context of MOBOs, and inference discrepancy. 
Let $\mathcal{X}_t = \{ {\bm x}_1 , \ldots , {\bm x}_t \} $  and $\mathcal{Y}_t = \{ {\bm y}_1 , \ldots, {\bm y}_t \}$
 be the set of input variables and observed values, respectively. 
Also let ${\bm r} $ be a reference point of a multi-objective black-box function ${\bm f} ({\bm x} ) =(f^{(1)} ({\bm x} ), \ldots , f^{(m)} ({\bm x} ) )$. 
Then, the simple PHV that we used in the experiments is given by 
$$
{\rm Vol} ({\bm f} (\mathcal{X} ) ; {\bm r} ) - {\rm Vol} ({\bm f} (\mathcal{X}_t ) ; {\bm r} ) ,
$$
where 
${\bm f} (A) \equiv \{  {\bm f} ({\bm a} ) \mid {\bm a} \in A \}$ and ${\rm Vol} ({\bm f} (A) ; {\bm r} )$ is the Lebesgue measure for 
$\{  {\bm b}  \mid   {\bm r}  \leq {\bm b}  \ {\tt and} \ {\bm b}  \leq {\bm f}  ({\bm a}  )  ,   {\bm a} \in A  \} $.
For a multi-objective black-box function ${\bm f} ({\bm x} )$, we used $r_j =   \min_{{\bm x} \in \mathcal{X}  }  f^{(j)} ({\bm x} )$ as the $j$-th reference point.  
As AFs, we  considered the random sampling (Random), uncertainty sampling (US), EHVI \citep{emmerich2008computation}, EMmI \citep{svenson2010multiobjective}, ePAL \citep{zuluaga2016varepsilon}, ParEGO \citep{knowles2006parego}, PFES \citep{suzuki2020multi} and proposed AF (Proposed). 
The next evaluation point was selected at random in Random. 
 We used  the AF $a_t ({\bm x} ) = \sigma^{(1)2}_t ({\bm x} ) + \cdots + \sigma^{(m)2}_t ({\bm x} )$ for US. 
In EHVI, we calculated sampling-based expected hypervolume  improvement given by 
$$
\frac{1}{S} \sum_{s=1}^S  \left \{ {\rm Vol}  ( \mathcal{Y}_t \cup \{ {\bm y}_s ({\bm x} ) \}   ; {\bm r} ) -  {\rm Vol}  ( \mathcal{Y}_t    ; {\bm r} )  \right \},
$$
where $ {\bm y}_s ({\bm x} )$ is generated from the posterior distribution of ${\bm f} ({\bm x} )$ and we set $S=20$. 
In EMmI, we calculated sampling-based expected maximin distance improvement given by 
$$
\frac{1}{S} \sum_{s=1}^S {\rm dist} ( {\bm y}_s ({\bm x} )  ,  {\rm Dom} ( \mathcal{Y}_t)),
$$
where $S$ and $ {\bm y}_s ({\bm x} )$ are the same definition in EHVI.
In ePAL, we performed the $\epsilon$-PAL algorithm with parameter ${\bm \epsilon } = (\epsilon_1, \ldots , \epsilon_m ) = (0,\ldots , 0)$. 
In ParEGO, for each iteration $t$, we uniformly generated the vector of coefficients ${\bm\lambda}_t =( \lambda^{(1)}_t,\ldots, \lambda^{(m)}_t)^\top $ with $0 \leq \lambda^{(i)}_t \leq 1$ and $\sum_{i=1}^m \lambda^{(i)}_t =1$, and calculated the scalarization $\tilde{y}_{t,\tilde{t}} =   0.05 {\bm \lambda}^\top_t  {\bm y}_t + \max_{1 \leq i \leq m } \lambda^{(i)}_t y^{(i)}_t $ for all $\tilde{t} \leq t$. 
We constructed the GP model for $\tilde{y}_{t,1}, \ldots , \tilde{y}_{t,t} $ using $\mathcal{X}_t$, where the kernel function was used the same kernel for $f({\bm x})$ but the noise variance was set to $10^{-8}$. 
We calculated the expected improvement (EI) \citep{movckus1975bayesian}  and the next point was selected by maximizing EI. 
In PFES, we used the random feature map \citep{rahimi2007random} to obtain posterior sample path. 
We generated a 500-dimensional random feature vector before BO, and used it for all iterations. 
The posterior sample path was generated 10 times for each iteration, and we calculated the PFES AF. 
In the four-objective Pareto optimization setting,  the maximum number of Pareto optimal input points defined based on the sample path  was restricted to 50 due to the computational cost. 
We also compared the commonly used evolutionary computation-based method NSGA-II \citep{deb2002fast}. 
The NSGA-II method was performed using {\tt nsga2R} version 1.1 in {\tt R}. 
In {\tt nsga2R}, we set the tournament size, crossover probability, crossover distribution index, mutation probability and mutation distribution index to  2, 0.9, 20, 0.1 and 3, respectively. 
We considered the population size $p$ to 5, 10, 15, 20, 30, 50, 100, 150 and 300. 
For each $p$, we set the number of generations to $300/p$. 
In NSGA-II, we used $\hat{\Pi}_t $ as the set of input variables reported by the algorithm.
Under this setup, one initial point was taken at random and the algorithm was run until the number of iterations reached 300. 
This simulation repeated 100 times and the average simple PHV regret and inference discrepancy at each iteration were calculated.
In NSGA-II, only results with the highest average performance at the end of the 300 iterations are shown ($p=30, 150$ in the two and four-objective settings, respectively).

\subsection{Details of Synthetic Function Experiments with Input Uncertainty} 
Here, the input space $\mathcal{X} \times \Omega $ was a compact subset.  
For $\mathcal{X} \times \Omega$, we considered infinite and finite set settings. 
We set $\mathcal{X} \times \Omega =[0.25,0.75]^2 \times [-0.25,0.25]^2$ in the infinite set setting. 
In the finite setting,  $\mathcal{X} \times \Omega$ was a set of grid points divided into $[-1,1]^3 \times [-1,1]^3$ equally spaced at $7^3 \times 7^3=117649$. 
\paragraph{ZDT1 Function} 
The black-box function in the infinite setting was used the ZDT1 benchmark function $\text{\bf ZDT1} ({\bm a} ) \in \mathbb{R}^2$ 
 with a two-dimensional input ${\bm a} $, and the environmental variable ${\bm w}$ was used as the input noise for ${\bm x}$. 
We standardized the ZDT1 function and further multiplied by minus one. 
The functional form we actually used is given as follows: 
\begin{align*}
g^{(1)} (a_1,a_2) &= a_1, \\ 
h (a_1,a_2) &= 1+ 9a_2 , \\ 
g^{(2)} (a_1,a_2) &=  h(a_1,a_2) -\sqrt{g^{(1)} (a_1,a_2) h (a_1,a_2) },\\
\text{\bf ZDT1} ({\bm a} ) &=  (f^{(1)} (a_1,a_2),f^{(2)} (a_1,a_2)) =   \left (
-\frac{g^{(1)} (a_1,a_2)-0.5}{\sqrt{0.042}},  -\frac{g^{(2)} (a_1,a_2)-3.9085}{\sqrt{2.5615}}
\right ).
\end{align*}
Thus, our considered black-box function was defined by $\text{\bf ZDT1} ({\bm x} +{\bm w}) $. 
We assumed ${\bm w}$ was the uniform distribution on $\Omega$ and used the Bayes risk $\mathbb{E} [\text{\bf ZDT1}  ({\bm x} +{\bm w})] $. 
We used the independent zero-mean Gaussian noise distribution with variance $10^{-6}$ for $f^{(i)} (a_1,a_2 )$. 
We constructed the independent GP model $\mathcal{G} \mathcal{P} (0, k({\bm  \theta},{\bm \theta}^\prime))$ for $f^{(i)}$, where 
${\bm \theta} = (x_1,x_2,w_1,w_2 )$ and 
$$
k({\bm  \theta},{\bm \theta}^\prime) = \exp \left (
-\frac{  \| {\bm \theta}-{\bm\theta}^\prime \|^2_2  }{0.2}
\right ).
$$
In order to calculate the true PF $Z^\ast$, we performed {\tt nsga2R} with population size 500 and the number of generations is 100. 
As comparison methods, we considered the Bayesian quadrature-based method (BQ) \citep{qing2023robust} and surrogate-assisted bounding box approach (SABBa) \citep{rivier2022surrogate}. 
Furthermore, four naive methods, Naive-random, Naive-US, Naive-EMmI and Naive-ePAL, were used for comparison.
In BQ, Bayes risk $\mathbb{E} [\text{\bf ZDT1}  ({\bm x} +{\bm w})] $ was modeled by the Bayesian quadrature, and its posterior distribution is again a GP. 
In this experiment, we can compute the exact posterior mean and variance, and we used them. 
Let ${\bm \mu}^{({\rm BQ})} _t ({\bm x} ) $ be a posterior mean for  $\mathbb{E} [\text{\bf ZDT1}  ({\bm x} +{\bm w})] $.
Then, we used $\hat{\Pi}_t $ to the set of Pareto optimal inputs calculated by ${\bm \mu}^{({\rm BQ})} _t ({\bm x} _1), \ldots , 
{\bm \mu}^{({\rm BQ})} _t ({\bm x}_t )$. 
The AF for ${\bm x}$, we used sampling-based approximation with sample size 20. 
In Proposed, we computed the sample average for  $\mu^{(m)}_{t} ({\bm x},{\bm w} ) - 3 \sigma^{(m)}_t ({\bm x},{\bm w} )$ and $\mu^{(m)}_{t} ({\bm x},{\bm w} ) + 3 \sigma^{(m)}_t ({\bm x},{\bm w} )$ by generating only two sample ${\bm w}_1$ and ${\bm w}_2$, 
and used them to ${\rm lcb}^{(m)}_t ({\bm x})$ and ${\rm ucb}^{(m)}_t ({\bm x})$. 
In order to calculate $\hat{\Pi}_t$, we used {\tt nsga2R} with population size 50 and the number of generations is 50. 
In SABBa, we set the number of new design set $\mathcal{X}_{new}$ to be read to 10. 
The elements in $\mathcal{X}_{new}$ were selected uniformly at random. 
We omitted the first approximation and then set $N_{first} =0$. 
The number of initial design set was set to 1, and for each iteration.
Similarly, the number of function evaluation was also set to 1. 
In the GP model for  $\mathbb{E} [\text{\bf ZDT1}  ({\bm x} +{\bm w})] $, we used  
$$
k({\bm x},{\bm x}^\prime) = \exp \left (
-\frac{  \| {\bm x}-{\bm x}^\prime \|^2_2  }{0.1}
\right ).
$$
In the AF for ${\bm x}$, we calculated the sampling-based AF calculation with sample size 20. 
For the threshold ${\bm s}_1$ and ${\bm s}_2$, we set ${\bm s}_1 ={\bm s}_2 = (  h_1 r_t, h_2 r_t )$, where 
$h_1 = \max_{ {\bm x} \in \mathcal{X}}  F^{(1)} ({\bm x} )  - \min_{ {\bm x} \in \mathcal{X}}  F^{(1)} ({\bm x} )$ and 
$h_2 = \max_{ {\bm x} \in \mathcal{X}}  F^{(2)} ({\bm x} )  - \min_{ {\bm x} \in \mathcal{X}}  F^{(2)} ({\bm x} )$. 
Here, $F^{(i)} ({\bm x} ) $ is the $i$-th element of $\mathbb{E} [\text{\bf ZDT1}  ({\bm x} +{\bm w})] $. 
Furthermore, the initial value of $r_t$ was set to $0.5$ and multiplied by $0.9$ each time a new $\mathcal{X}_{new}$ was read, and if $r_t <0.001$, then we fixed $r_t =0.001$.  
The $\hat{\Pi}_t$ was set to the Pareto-optimal points defined based on ${\bm \rho}_{SA} ({\bm x})$ and $\tilde{\bm \rho} ({\bm x} )$  (see, \cite{rivier2022surrogate} for details) with respect to the inputs read so far. 
In the naive methods, ${\bm w}$ was generated five times from the same ${\bm x}$ in one iteration $t$, and the sample mean  of the black-box function values were calculated. 
By using ${\bm x}$ and these values, the experiments in naive four methods were  performed as a usual MOBO. 
We used the following kernel function:
$$
k({\bm x},{\bm x}^\prime) = \exp \left (
-\frac{  \| {\bm x}-{\bm x}^\prime \|^2_2  }{0.1}
\right ).
$$
The same calculation (approximation) method as in the without IU setting was used for calculating AFs. 
For all methods, the maximization of AFs was performed using \textit{optim} function with the L-BFGS-B method in {\tt R}. 

\paragraph{6D-Rosenbrock Function}
The black-box function in the finite setting was used the six-dimensional Rosenbrock function $f(w_1,w_2,x_1,x_2,x_3,w_3) \in \mathbb{R}$. 
The functional form that we used is given as follows:
$$
f(a_1,a_2,a_3,a_4,a_5,a_6) = \frac{273.45-\sum_{i=1}^5 \{  100 (a_{i+1} -a_i^2) + (1-a_i)^2  \}}{\sqrt{28153.22}}.
$$
We assume that  ${\bm w}$ was a discretized normal distribution on $\Omega = \Omega_1 \times \Omega_2 \times \Omega_3$. 
For each ${\bm w} \in \Omega_i$, the probability math function of ${\bm w} $ is given by 
$$
p({\bm w} ) =  \frac{\phi({\bm w})}{  \sum_{  \hat{\bm w} \in \Omega_i  }   \phi(\hat{\bm w}) } ,
$$
where $\phi (x) $ is the probability density function of standard normal distribution. 
As risk measures, we used the expectation and negative standard deviation: 
$$
\mathbb{E}[f(w_1,w_2,x_1,x_2,x_3,w_3) ], \quad -\sqrt{\mathbb{V}[f(w_1,w_2,x_1,x_2,x_3,w_3) ]} .
$$
As comparison methods, we considered the Mean-variance-based method (MVA) \citep{iwazaki2020mean}, SABBa, 
 Naive-random, Naive-US, Naive-EMmI and Naive-ePAL. 
We used the independent zero-mean Gaussian noise distribution with variance $10^{-6}$ for $ f(w1,w2,x1,x2,x3,w3)$. 
We constructed the GP model $\mathcal{G} \mathcal{P} (0, k({\bm  \theta},{\bm \theta}^\prime))$ for $f$, where 
${\bm \theta} = (x_1,x_2,x_3,w_1,w_2,w_3 )$ and 
$$
k({\bm  \theta},{\bm \theta}^\prime) = \exp \left (
-\frac{  \| {\bm \theta}-{\bm\theta}^\prime \|^2_2  }{4}
\right ).
$$
This experiment is the setting that the number of black-box functions and risk measures are different. 
Thus, in Proposed, the algorithm was performed using Algorithm \ref{alg:2}. 
In SABBa, we set the number of new design set $\mathcal{X}_{new}$ to be read to 10. 
The elements in $\mathcal{X}_{new}$ were selected uniformly at random. 
We omitted the first approximation and then set $N_{first} =0$. 
The number of initial design set was set to 1, and for each iteration.
Similarly, the number of function evaluation was also set to 1. 
In the GP model for Bayes risk and negative standard deviation, we used the following kernel:
$$
k({\bm x},{\bm x}^\prime) = \exp \left (
-\frac{  \| {\bm x}-{\bm x}^\prime \|^2_2  }{4}
\right ).
$$
In the AF for ${\bm x}$, we calculated the sampling-based AF calculation with sample size 20. 
For the threshold ${\bm s}_1$ and ${\bm s}_2$, we set ${\bm s}_1 ={\bm s}_2 = (  h_1 r_t, h_2 r_t )$, where 
$h_1 = \max_{ {\bm x} \in \mathcal{X}}  F^{(1)} ({\bm x} )  - \min_{ {\bm x} \in \mathcal{X}}  F^{(1)} ({\bm x} )$ and 
$h_2 = \max_{ {\bm x} \in \mathcal{X}}  F^{(2)} ({\bm x} )  - \min_{ {\bm x} \in \mathcal{X}}  F^{(2)} ({\bm x} )$. 
Here, $F^{(1)} ({\bm x} ) $ and $F^{(2)} ({\bm x} ) $ are Bayes risk and negative standard deviation, respectively. 
Furthermore, the initial value of $r_t$ was set to $2$ and multiplied by $0.9$ each time a new $\mathcal{X}_{new}$ was read, and if $r_t <0.001$, then we fixed $r_t =0.001$.  
The $\hat{\Pi}_t$ was set to the Pareto-optimal points defined based on ${\bm \rho}_{SA} ({\bm x})$ and $\tilde{\bm \rho} ({\bm x} )$  with respect to the inputs read so far. 
In the naive methods, ${\bm w}$ was generated five times from the same ${\bm x}$ in one iteration $t$, and the sample mean and the negative square root of the sample variance of the black-box function values were calculated. 
By using ${\bm x}$ and these values, the experiments in naive four methods were  performed as a usual MOBO. 
We used the following kernel function:
$$
k({\bm x},{\bm x}^\prime) = \exp \left (
-\frac{  \| {\bm x}-{\bm x}^\prime \|^2_2  }{2}
\right ).
$$
The same calculation (approximation) method as in the without IU setting was used for calculating AFs.

\subsection{Details of Real-world Simulation Model}
We applied the proposed method to docking simulations for real-world chemical compounds. 
As a dataset for compounds, we used the software suite {\em Schr\"{o}dinger} \citep{schrodinger2021}. 
Given a set of compounds, we applied the software ``QikProp'' in the suite, a software to compute various chemical properties, for explanatory variables. 
We also applied the software ``Glide'' in the suite, a software for calculating docking scores. 
We took the black-box function as the original docking score plus 5 and then multiplied by -1. 
When performing docking simulations, it is necessary to specify both the protein of interest and the specific site on the protein where compounds are expected to dock. 
We used the protein ``KAT1'', whose structure is available at \url{https://pdbj.org/mine/summary/6v1x}, and the 16th and 18th sites  computed by the software ``SiteMap'' in the suite. 
Each chemical compound $C_i$ may have an  isomer $S_{ij}$, and in this experiment the maximum number of  isomers  was limited to 10.
 For each $i$, we computed a 51-dimensional isomer-independent vector of explanatory variables ${\bm x}_i $ and a 51-dimensional environment variable ${\bm w}_{ij} $ that can vary with isomers, using physicochemical features of $(C_i,S_{ij})$ computed using QikProp. 
Specifically,  the 51-dimensional physicochemical features of $(C_i,S_{ij})$ calculated by QikProp were used as ${\bm w}_{ij} $. 
In addition, we defined ${\bm x}_i $ as ${\bm x}_i  =\frac{1}{N_i} \sum_{j=1}^{N_i} {\bm w}_{ij}$. 
Thus, the black-box functions, the docking scores in the 16th and 18th sites, can be expressed as  $f^{(1)} ({\bm x}_i,{\bm w}_{ij} )$ and  $f^{(2)}  ({\bm x}_i,{\bm w}_{ij} )$, respectively. 
As risk measures for $C_i$, we considered the following  measures: 
\begin{description}
\item [Worst-case (WC):] $F^{(m)} ({\bm x}_i) = \min_{ j \in [N_i]   }   f^{(m)}  ({\bm x}_i,{\bm w}_{ij} ) $. 
\item [Worst-case Bayes risk (WCBR):] Define the Bayes risk under the worst-case candidate distribution, that is, 
\begin{align*}
F^{(m)} ({\bm x}_i) &=   \min_{ {\bm \alpha}_i \in \mathcal{A}_i  }  \sum _{j=1}^{N_i} \alpha_{ij} f^{(m)}  ({\bm x}_i,{\bm w}_{ij} ).
\end{align*}
\end{description}
The  $\mathcal{A}_i $ is the set of  $ {\bm \alpha}_i \in \mathbb{R}^{N_i}$ satisfying 
\begin{align*}
   0 \leq \alpha_{ij} \leq 1, \sum_{j=1}^{N_i} \alpha_{ij} =1, \| {\bm\alpha}_i - \tilde{\bm\alpha}_i \|_1  \leq 0.25 , 
\end{align*}
where $\tilde{\alpha}_{ij}  =  N^{-1}_i $ .
The total number of compounds was 429, and the total number of data including isomers was 920. 
We compared  Proposed, SABba and naive four methods. 
In this experiment, we used the independent GP model for $f^{(m)}$, where the kernel function is given by 
$$
k({\bm \theta},{\bm \theta}^\prime) = 25 \exp \left (
-\frac{  \| {\bm \theta}-{\bm \theta}^\prime \|^2_2  }{l}
\right ).
$$
The length scale parameter was computed using the median heuristic $l = 0.5 {\rm Median}  \{  \| {\bm \theta}_i - {\bm \theta}_j \|^2 \mid 1 \leq i < j \leq 920 \} $.
In this experiment, there was no observation noise. 
Nevertheless, we added $10^{-3} {\bm I}_t $ to the kernel matrix ${\bm K}_t $ to stabilize the inverse matrix calculation. 
In SABBa,  we set the number of new design set $\mathcal{X}_{new}$ to be read to 10. 
The elements in $\mathcal{X}_{new}$ were selected uniformly at random. 
We omitted the first approximation and then set $N_{first} =0$. 
The number of initial design set was set to 1, and for each iteration.
Similarly, the number of function evaluation was also set to 1. 
In the GP model for risk measures, we used the following kernel:
$$
k({\bm x},{\bm x}^\prime) = 25 \exp \left (
-\frac{  \| {\bm x}-{\bm x}^\prime \|^2_2  }{l}
\right ),
$$
where the length scale parameter was computed using the median heuristic $l = 0.5 {\rm Median}  \{  \| {\bm  x}_i - {\bm x}_j \|^2 \mid 1 \leq i < j \leq 429 \} $.
In the AF for ${\bm x}$, we calculated the sampling-based AF calculation with sample size 20. 
For the threshold ${\bm s}_1$ and ${\bm s}_2$, we set ${\bm s}_1 ={\bm s}_2 = (  h_1 r_t, h_2 r_t )$, where 
$h_1 = \max_{ {\bm x} \in \mathcal{X}}  F^{(1)} ({\bm x} )  - \min_{ {\bm x} \in \mathcal{X}}  F^{(1)} ({\bm x} )$ and 
$h_2 = \max_{ {\bm x} \in \mathcal{X}}  F^{(2)} ({\bm x} )  - \min_{ {\bm x} \in \mathcal{X}}  F^{(2)} ({\bm x} )$. 
Furthermore, the initial value of $r_t$ was set to $0.5$ and multiplied by $0.9$ each time a new $\mathcal{X}_{new}$ was read, and if $r_t <0.01$, then we fixed $r_t =0.01$.  
We regarded this as a high accurate setting. 
As a low accurate setting, we considered that   the initial value of $r_t$ was set to $2$ and multiplied by $0.99$ each time a new $\mathcal{X}_{new}$ was read, and if $r_t <0.01$, then we fixed $r_t =0.01$.  
In the original SABBa method, \cite{rivier2022surrogate} does not provide the worst-case Bayes risk setting. 
Hence, we modified the first formula of Equation (6) in \cite{rivier2022surrogate} to $\inf_{{\bm \xi} \in \mathcal{A}  } \mathbb{E}_{{\bm \xi}}  [  \bar{\varepsilon}_{q_{{\bm x}}} ({\bm \xi}) ]$.
The $\hat{\Pi}_t$ was set to the Pareto-optimal points defined based on ${\bm \rho}_{SA} ({\bm x})$ and $\tilde{\bm \rho} ({\bm x} )$  with respect to the inputs read so far. 
In the naive four  methods, we calculated docking scores for all isomers in the compound $C_i$ at iteration $t$ and determined the exact risk values.  
By using ${\bm x}$ and these values, the experiments in naive four methods were  performed as a usual MOBO. 
We used the following kernel function:
$$
k({\bm x},{\bm x}^\prime) = 25 \exp \left (
-\frac{  \| {\bm x}-{\bm x}^\prime \|^2_2  }{l}
\right ),
$$
where the length scale parameter was computed using the median heuristic $l = 0.5 {\rm Median}  \{  \| {\bm  x}_i - {\bm x}_j \|^2 \mid 1 \leq i < j \leq 429 \} $.
The same calculation (approximation) method as in the without IU setting was used for calculating AFs.

\subsection{Additional Experiments} 
\begin{figure*}[tb]
\begin{center}
 \begin{tabular}{cc}
 \includegraphics[width=0.45\textwidth]{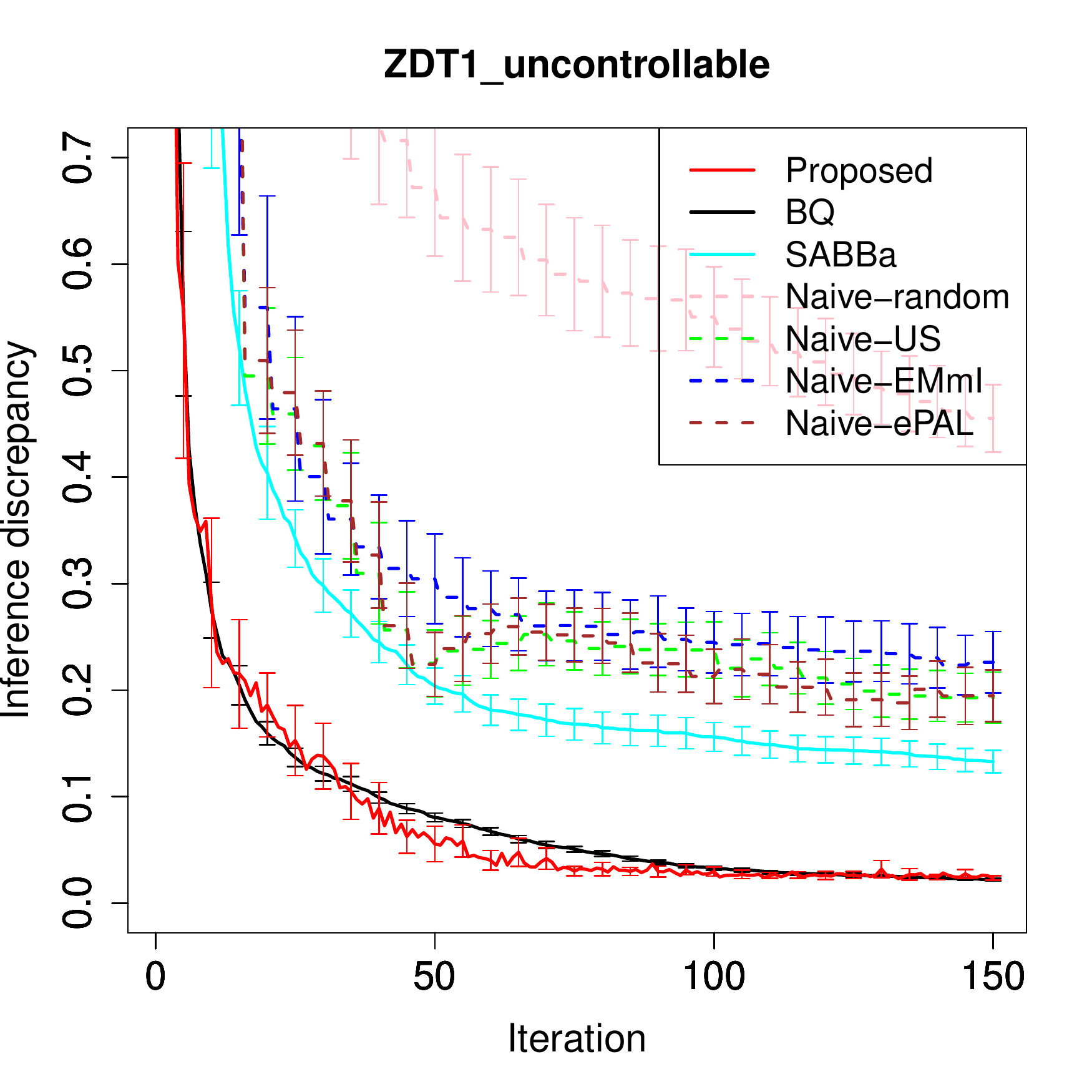} &
 \includegraphics[width=0.45\textwidth]{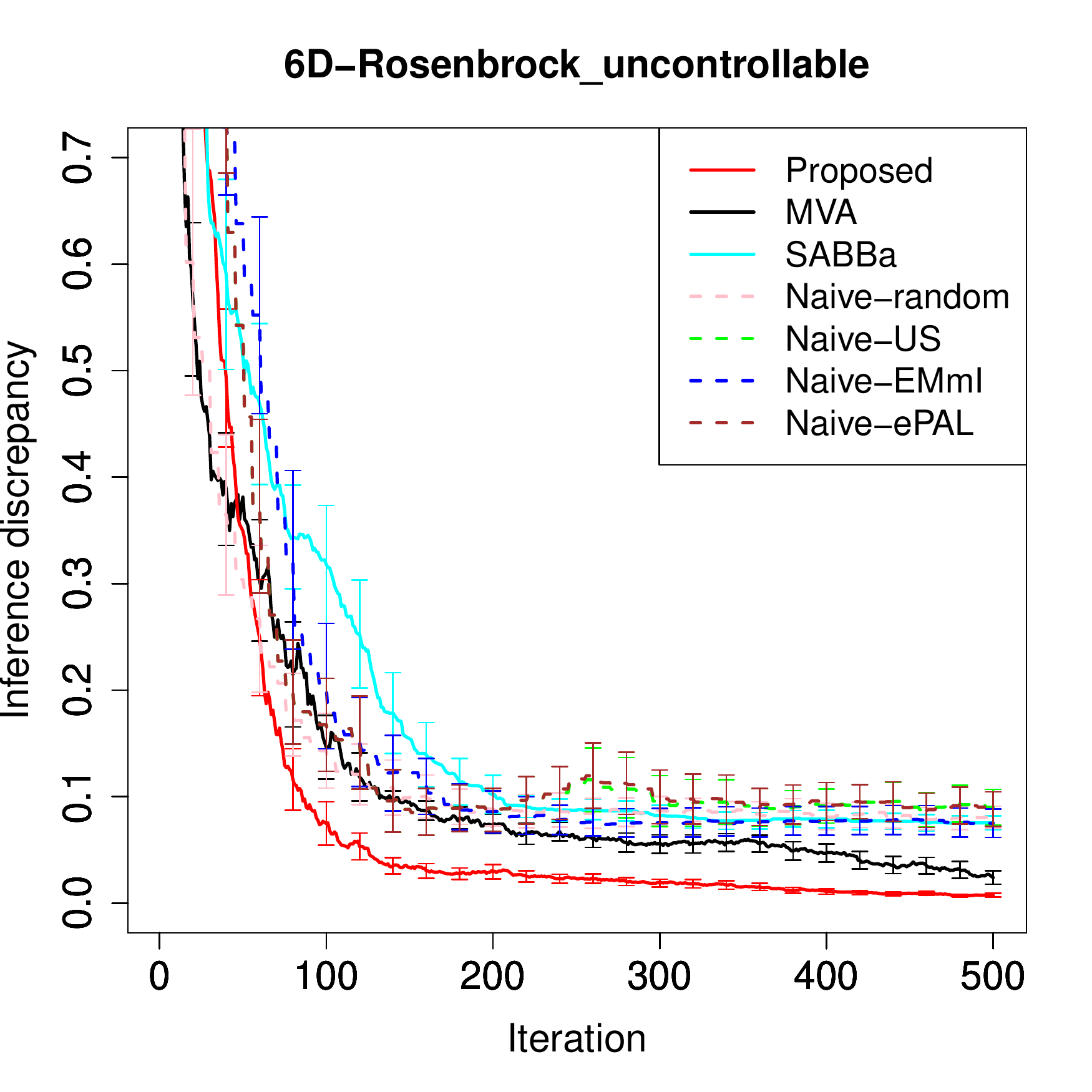} 
 \end{tabular}
\end{center}
 \caption{Comparison with MOBO methods. 
Solid (and dashed) lines are averages of the inference discrepancy for each iteration in  100 trials. 
Each error bar length represents the six times the standard error. 
The left  and right columns respectively represent the ZDT1 and six-dimensional Rosenbrock setups under the uncontrollable setting.
}
\label{fig:exp_additional1}
\end{figure*}

\paragraph{Uncontrollable Setting for Synthetic Experiments} 
Here, we give the results of synthetic experiments (ZDT1 and 6D-Rosenbrock) under the uncontrollable setting. 
We performed the same experiments except for the selection of ${\bm w}$. 
Figure \ref{fig:exp_additional1} shows the similar results as in the simulator setting. 

\begin{figure}[tb]
\begin{center}
 \begin{tabular}{c}
 \includegraphics[width=0.9\textwidth]{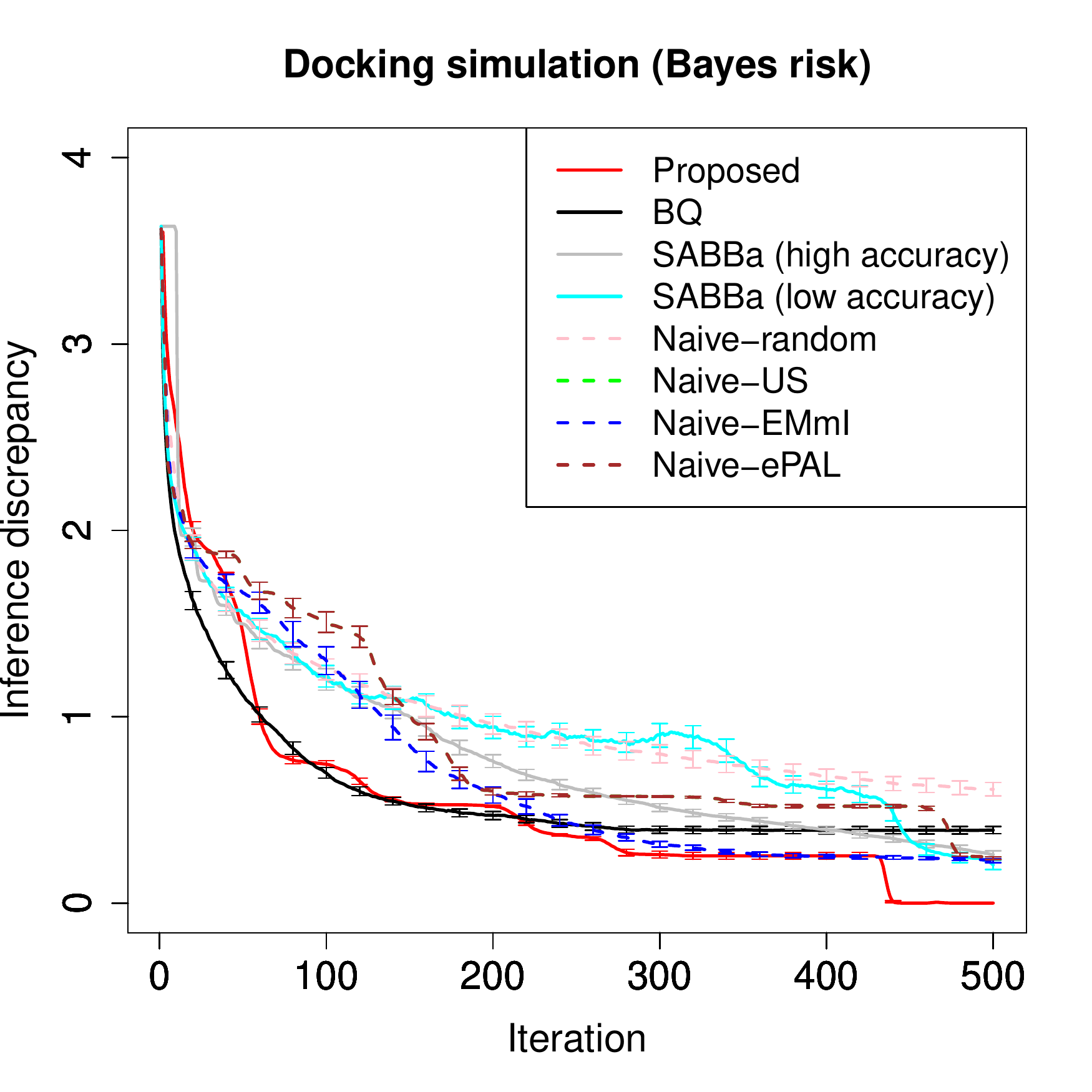} 
 \end{tabular}
\end{center}
 \caption{Comparison with MOBO methods. 
Solid (and dashed) lines are averages of the  inference discrepancy of Bayes risk setting for each iteration  in 920 or 429 trials. 
Each error bar length represents the six times the standard error. 
}
\label{fig:exp_additional2}
\end{figure}

\paragraph{Docking Simulation Experiments Using Bayes Risk}
In the docking simulation experiments, we also considered Bayes risk (BR) $F^{(m)} ({\bm x}_i) = \frac{1}{N_i} f^{(m)}  ({\bm x}_i,{\bm w}_{ij} )$. 
In this experiment, we also considered the BQ method. 
In BQ, $f^{(m)} ({\bm x}_i,{\bm w}_{ij} )$ was modeled in the same way as in Proposed. 
The AF for ${\bm x} $ was calculated using sampling-based approximation with sample size 20. 
Figure \ref{fig:exp_additional2} shows the similar results as in the WC and WCBR settings. 
Also in the BR setting, only the proposed method correctly identifies the true PF at the end of 500 iterations at all 920 different initial points.
Specifically, after 481 iterations, the true PF is identified for all 920 different initial points. 

\begin{figure*}[tb]
\begin{center}
 \begin{tabular}{ccc}
 \includegraphics[width=0.3\textwidth]{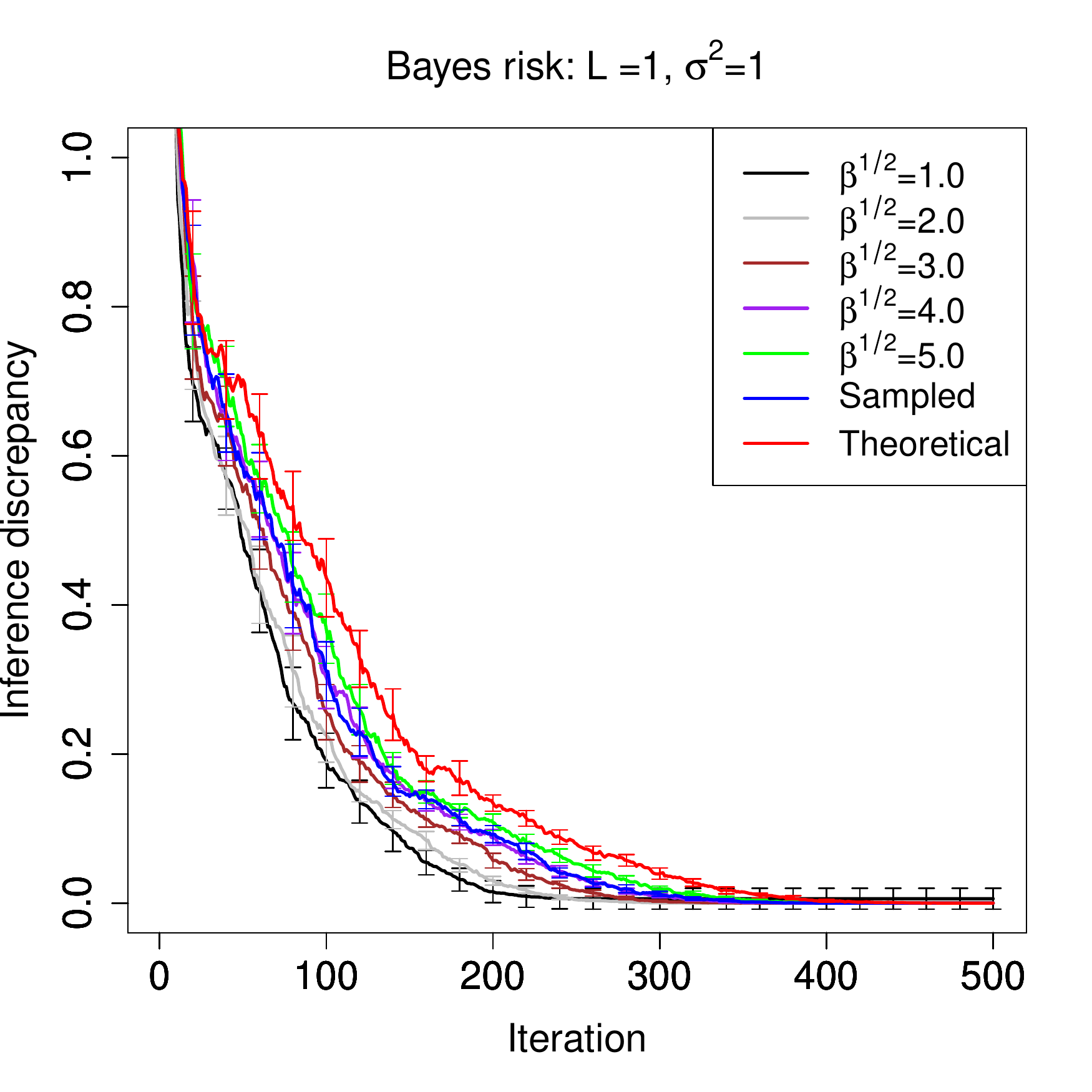} &
 \includegraphics[width=0.3\textwidth]{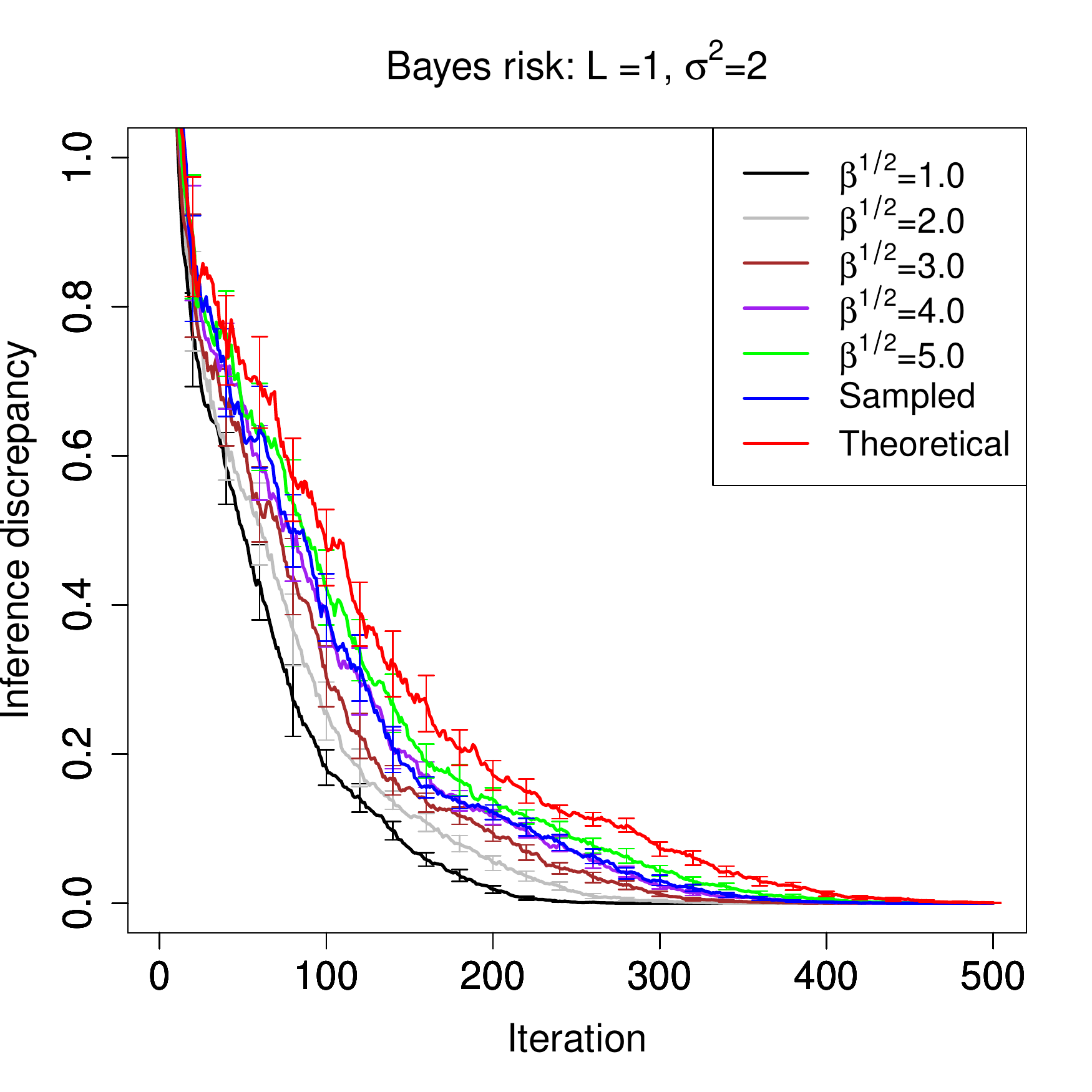} &
 \includegraphics[width=0.3\textwidth]{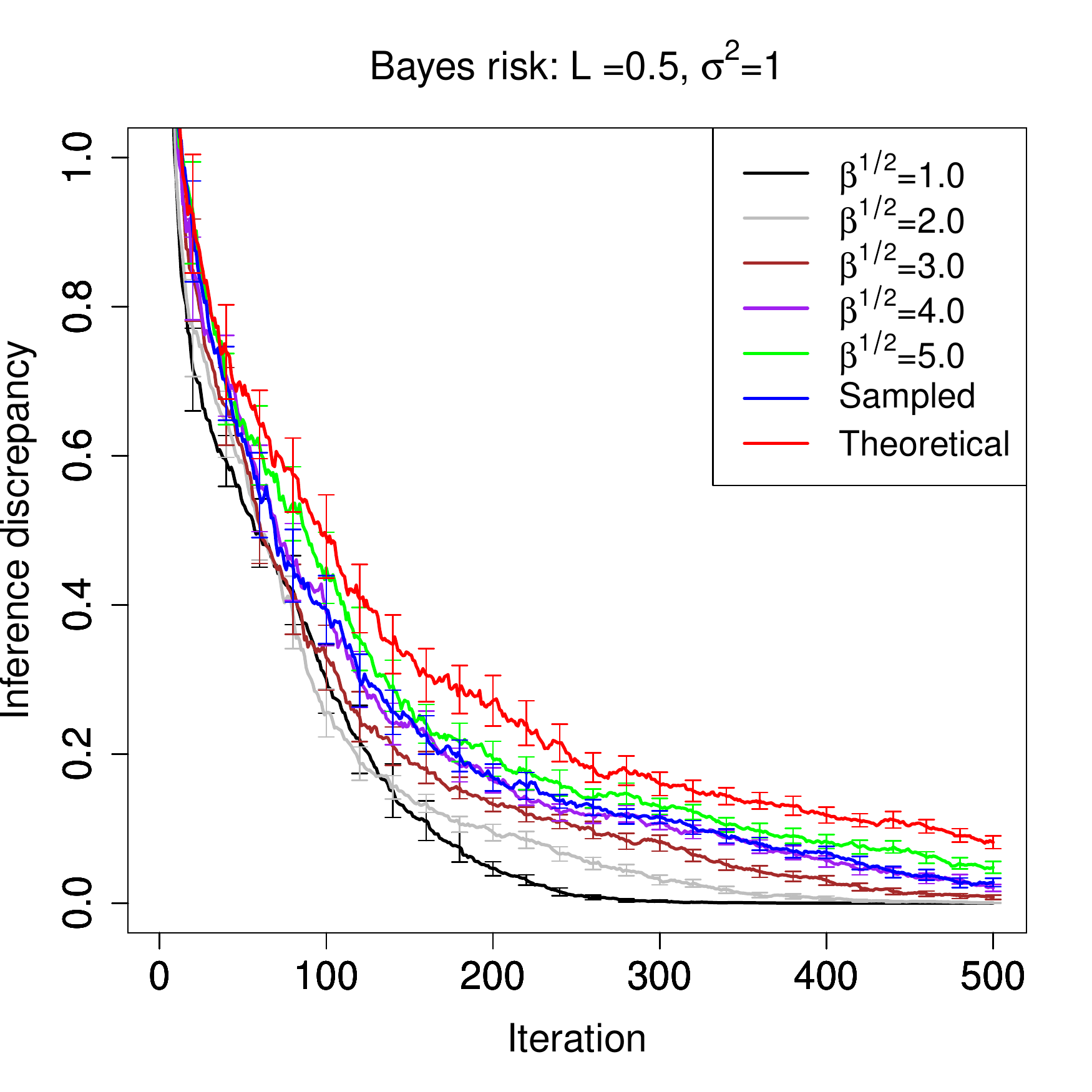} \\
 \includegraphics[width=0.3\textwidth]{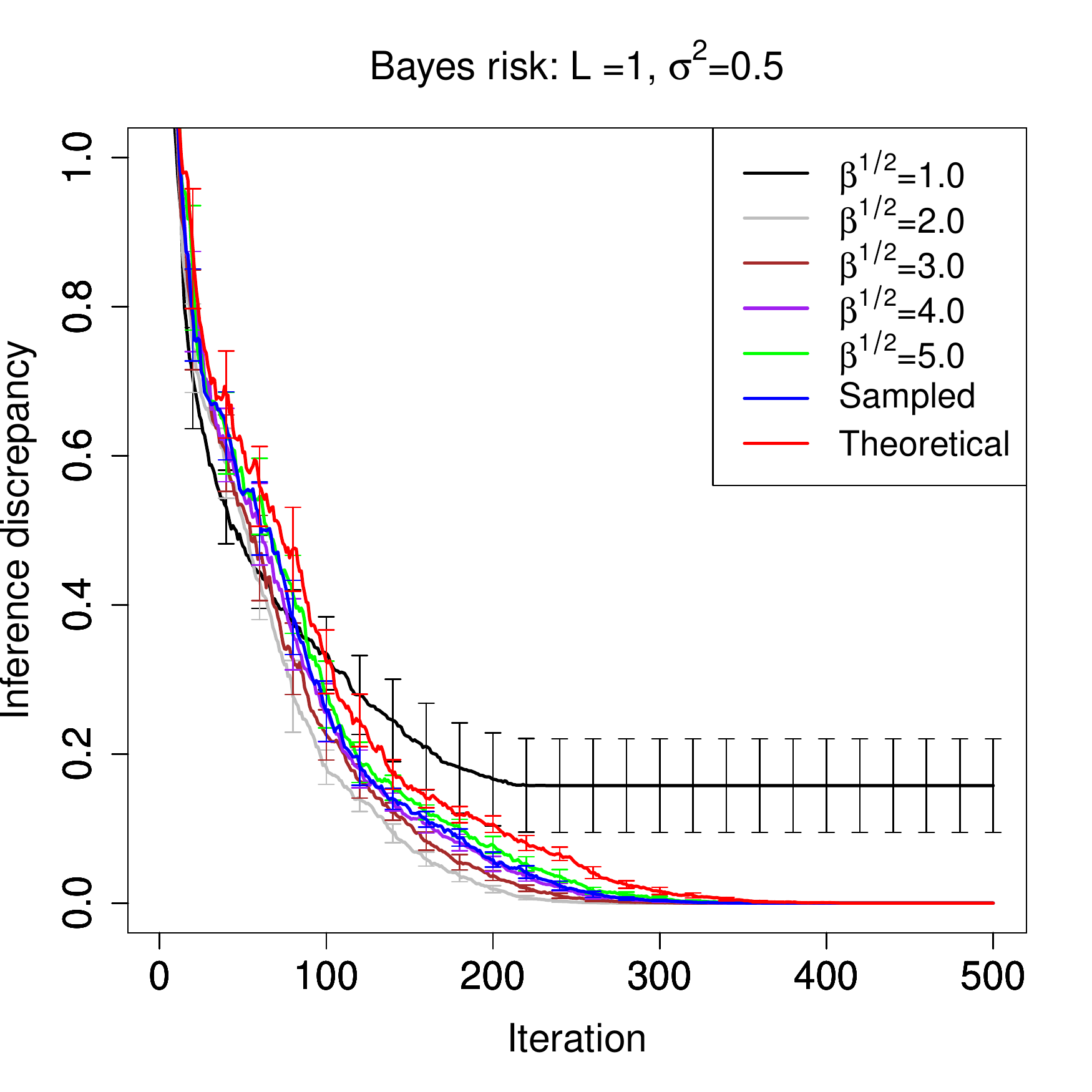} &
 \includegraphics[width=0.3\textwidth]{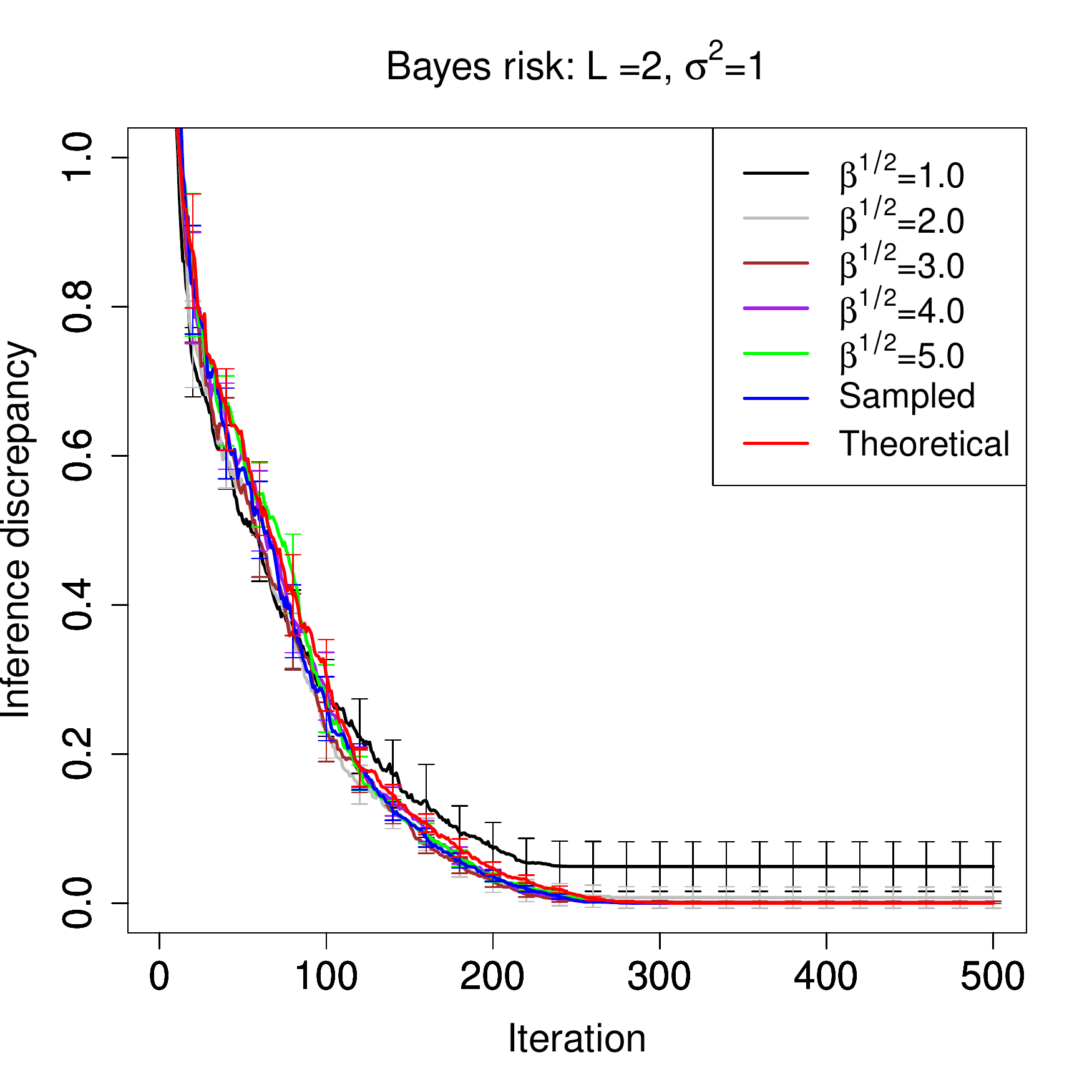} &
 \includegraphics[width=0.3\textwidth]{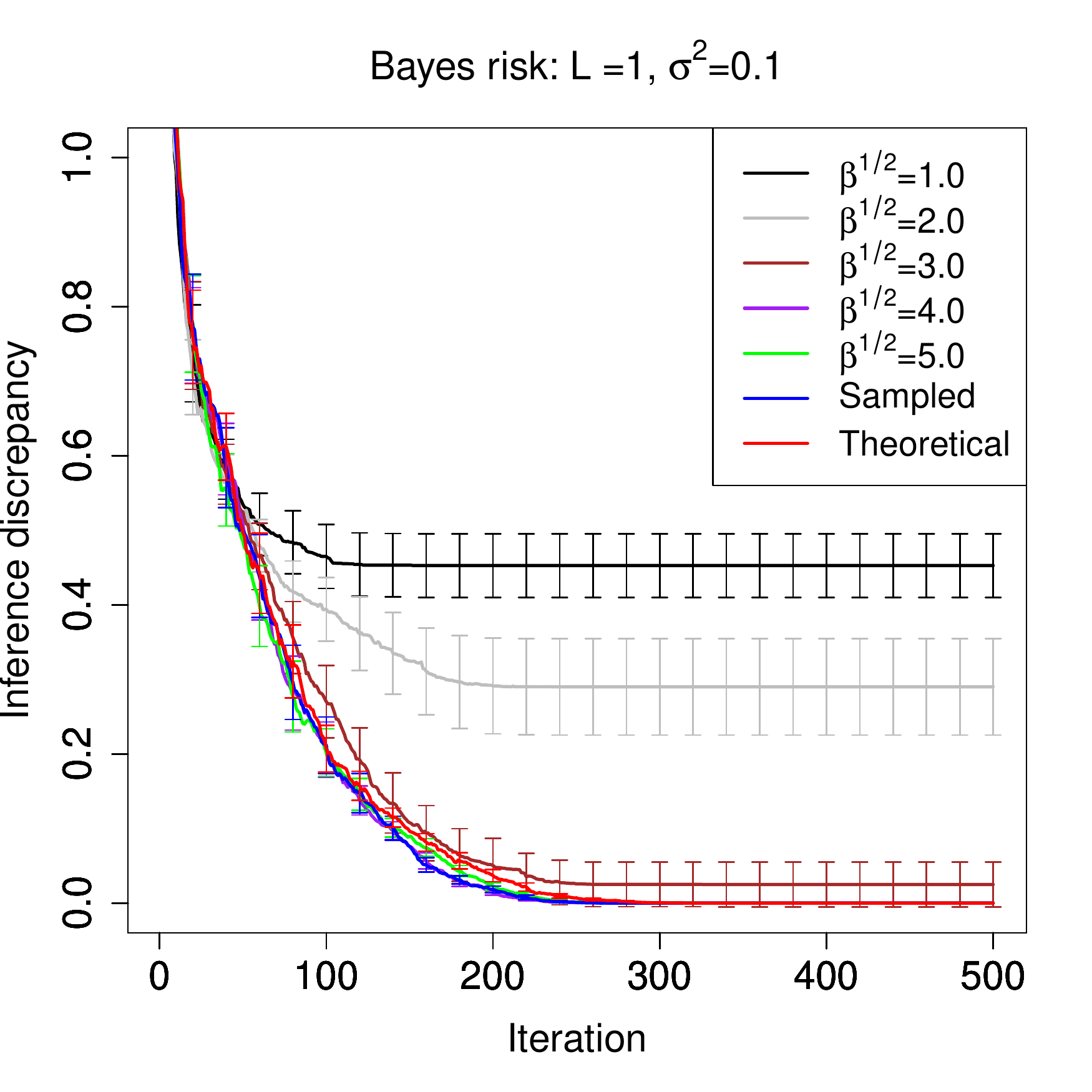} 
 \end{tabular}
\end{center}
 \caption{Comparison with different hyperparameters.
Solid lines are averages of the  inference discrepancy for each iteration in  100 trials. 
Each error bar length represents the six times the standard error. 
In the top row, the left column represents the case that the kernel of the surrogate model is equal to the true kernel. 
The right two columns represent the cases that the posterior variance is predicted larger. 
In the bottom row, the left, center and right  columns represent the cases that the posterior variance is predicted smaller. 
}
\label{fig:exp_HP}
\end{figure*}

\paragraph{Hyperparameter Sensitivity} 
In this section, we confirm the sensitivity for hyperparameters. 
In this experiment, the input space $\mathcal{X} \times \Omega \subset \mathcal{R}^2 \times \mathcal{R} $ was a set of grid points divided into $[-2,2]^3$ equally spaced at $16 \times 16 \times 16 =4096$. 
The true black-box functions $f^{(1)} (x_1,x_2,w_1) $ and $f^{(2)} (x_1,x_2,w_1) $ were defined as the independent sample path from the GP 
$\mathcal{G} \mathcal{P} (0, k^\ast (\cdot,\cdot, ) )$, where $k^\ast (\cdot,\cdot, ) $ is given by 
$$
k^\ast (      (x_1,x_2,w_1) , (x^\prime_1, x^\prime_2,w^\prime_1 )) 
=
\exp  \left (
-\frac{   (x_1-x^\prime_1)^2 +   (x_2-x^\prime_2)^2    + (w_1-w^\prime_1)^2  }{1}
\right ).
$$
We used the zero-mean independent Gaussian noise with variance $10^{-6} $. 
As the distribution of $ w \in \Omega$, we used the discretized normal distribution $p(w) $ given by 
$$
p(w) =   \frac{\phi(w)}{\sum_{w^\prime \in \Omega}  \phi (w^\prime)}.
$$
We considered Bayes risk in this experiment. 
As the GP surrogate model, we used independent GP model for $f^{(1)} $ and $f^{(2)}$, and the kernel function that we used is given by 
$$
k (      (x_1,x_2,w_1) , (x^\prime_1, x^\prime_2,w^\prime_1 )) 
=
\sigma^2  \exp  \left (
-\frac{   (x_1-x^\prime_1)^2 +   (x_2-x^\prime_2)^2    + (w_1-w^\prime_1)^2  }{L}
\right ).
$$ 
We considered  six cases for $(L,\sigma^2)$, 
$$
(L,\sigma^2) = (1,1) , \ 
(L,\sigma^2) = (1,2) , \ 
(L,\sigma^2) = (0.5,1) , \ 
(L,\sigma^2) = (1,0.5) , \ 
(L,\sigma^2) = (2,1) , \ 
(L,\sigma^2) = (1,0.1) .
$$
Similarly, we considered seven cases for $\beta^{1/2}_t$, 
\begin{align*}
\beta^{1/2}_t &=1, \ 
\beta^{1/2}_t =2, \ 
\beta^{1/2}_t =3, \ 
\beta^{1/2}_t =4, \ 
\beta^{1/2}_t =5, \\
\beta^{1/2}_t &=\sqrt{2 \log (2 \times 4096 /2) + r_t }, \ 
  \beta^{1/2}_t =\sqrt{2 \log (2 \times 4096  \pi ^2 t^2 /(6 \times 0.1))}, 
\end{align*}
where $r_t $ is a realized value from the exponential distribution with mean $0.5$. 
The last two definitions of $\beta^{1/2}_t $ are proposed by \cite{pmlr-v202-takeno23a} and \cite{GPUCB}, respectively. 
We regarded them as Sampled and Theoretical values, respectively. 
 Under this setup, one initial point was taken at random and the algorithm was run until the number of iterations reached 500. 
This simulation repeated 100 times and the average inference discrepancy at each iteration was calculated.
From the top of Fig. \ref{fig:exp_HP}, it can be confirmed that 
 $\beta^{1/2}_t =2$ is sufficient if the correct kernel is used, and  $\beta^{1/2}_t =1$ is sufficient for  the right two columns of the top row that  the posterior variance is predicted  larger. 
On the other hand, if the cases the posterior variance is predicted  smaller, $\beta^{1/2}_t =3 $ is still insufficient in the case of 
$L=1,\sigma^2=0.1$.

\paragraph{Computational Time Experiments} 
Here, we measured the computational time required to obtain $({\bm x}_{t+1},{\bm w}_{t+1}) $ for each iteration $t$ in  the proposed method, where the time required for  modeling GP is not included in the measurement time because 
all MOBO methods, including the comparison methods, perform GP modeling. 
We measured the computational time for each iteration in a single trial and calculated its average and standard deviation for the iterations in the experiments performed in the main body.
From Table \ref{tab:computational_time}, the computational time for AFs in the proposed method is acceptable even in a 6D-Rosenbrock setting with more than 100,000 candidate points. 
In contrast, the reason why the computational time in the ZDT1 setting is larger than the others is due to the finite approximation of  PF by the NSGA-II algorithm. 
Therefore, the computational time  can be reduced if the population size $n_p$ or number of generations $n_g$ is reduced. 
Nevertheless, the computational time is acceptable even for our experimental setup, $n_p=n_g =50$.

\begin{table}[tb]
  \centering
    \caption{Computational time (second) for obtaining $({\bm x}_{t+1},{\bm w}_{t+1} )$ in the proposed method}
  \begin{tabular}{c|c|c} \hline 
Experimental setup & Average  & Standard deviation \\ \hline \hline
Two-objective optimization without IU & 0.93 & 0.60 \\ \hline 
Four-objective optimization without IU & 2.03 & 1.26 \\ \hline 
ZDT1 with IU & 5.60 & 2.25 \\ \hline 
6D-Rosenbrock with IU & 0.73 & 0.16 \\ \hline 
Docking simulation (WC) & 0.0161 & 0.0030 \\ \hline 
Docking simulation (WCBR) & 0.0236 & 0.0089 \\ \hline \hline
  \end{tabular}
\label{tab:computational_time}
\end{table}

  \end{document}